\documentclass{article}

\usepackage[final]{neurips_data_2022}

\usepackage[utf8]{inputenc} % allow utf-8 input
\usepackage[T1]{fontenc}    % use 8-bit T1 fonts
\usepackage[colorlinks, linkcolor=red]{hyperref} % hyperlinks
\usepackage{url}            % simple URL typesetting
\usepackage{booktabs}       % professional-quality tables
\usepackage{amsfonts}       % blackboard math symbols
\usepackage{nicefrac}       % compact symbols for 1/2, etc.
\usepackage{microtype}      % microtypography
\usepackage{bm}
\usepackage{amsmath}
\usepackage{xcolor}         % colors
\usepackage{graphicx, wrapfig}
\usepackage{subfigure}
\usepackage[figuresright]{rotating}
\usepackage{appendix}
\usepackage{listings}
\usepackage{makecell}
\usepackage{longtable}
\usepackage{threeparttable}
\usepackage{multirow,array}
\usepackage{hyperref}
\usepackage{comment}
\hypersetup{
    colorlinks=true,
    linkcolor=red,
    citecolor=blue,
    filecolor=magenta,      
    urlcolor=blue,
linktocpage}

\title{Towards Human-Level Bimanual Dexterous Manipulation with Reinforcement Learning}

\begin{document}

\maketitle

\begin{center}
\vspace{-1.5cm}
{\textbf{Yuanpei Chen}$^1$, \textbf{Tianhao Wu}$^2$, \textbf{Shengjie Wang}$^1$, \textbf{Xidong Feng}$^3$, \textbf{Jiechuang Jiang}$^1$, \\ \textbf{Stephen Marcus McAleer}$^4$, \textbf{Yiran Geng}$^{1,5}$, \textbf{Hao Dong}$^2$, \textbf{Zongqing Lu}$^1$, \\ \textbf{Song-Chun Zhu}$^{1,5}$}, \textbf{Yaodong Yang}$^{1,5,\dag}$ \\
$^{1}$Institute for AI, Peking University \\
$^{2}$Center on Frontiers of Computing Studies, Peking University \\
$^{3}$University College London \\
$^{4}$Carnegie Mellon University \\
$^{5}$Beijing Institute for General Artificial Intelligence \\
\end{center}
% \footnotetext{$^{1}$Institute for AI, Peking University}
% \footnotetext{$^{2}$Center on Frontiers of Computing Studies, Peking University}
% \footnotetext{$^{3}$University College London}
% \footnotetext{$^{4}$Carnegie Mellon University}
% \footnotetext{$^{5}$Beijing Institute for General Artificial Intelligence}

\vspace{0.5cm}
\begin{abstract}
Achieving human-level dexterity is an important open problem in robotics. However, tasks of dexterous hand manipulation, even at the baby level, are challenging to solve through reinforcement learning (RL). 
The difficulty lies in the high degrees of freedom and the required cooperation among heterogeneous agents (e.g., joints of fingers).  
In this study, we propose the \textbf{Bi}manual \textbf{Dex}terous \textbf{Hands} Benchmark (Bi-DexHands), a simulator that involves two dexterous hands with tens of bimanual manipulation tasks and thousands of target objects. Specifically, tasks in Bi-DexHands are designed to match different levels of human motor skills according to cognitive science literature. We built Bi-DexHands in the Issac Gym; this enables highly efficient RL training,  reaching 30,000+ FPS by only one single NVIDIA RTX 3090. 
We provide a comprehensive benchmark for popular RL algorithms under different settings; this includes Single-agent/Multi-agent RL, Offline RL, Multi-task RL, and Meta RL. Our results show that the PPO type of on-policy algorithms can master simple manipulation tasks that are equivalent up to 48-month human babies (e.g., catching a flying object, opening a bottle), while multi-agent RL can further help to master manipulations that require skilled bimanual cooperation (e.g., lifting a pot, stacking blocks).  
Despite the success on each single task, when it comes to acquiring multiple manipulation skills, existing RL algorithms fail to work in most of the multi-task and the few-shot learning settings, which calls for more substantial development from the RL community. 
Our project is open sourced at \href{https://github.com/PKU-MARL/DexterousHands}{https://github.com/PKU-MARL/DexterousHands}. \footnote{$\dag$: Corresponding to <yaodong.yang@pku.edu.cn>}
\end{abstract}

\vspace{-0.3cm}
\section{Introduction}
\vspace{-0.2cm}

Humans have a skillful ability to manipulate objects of different shapes, sizes, and materials, which rely on the dexterity of our hands and fingers. Building a robot inspired by human hands that can autonomously manipulate various objects has always been an important component of the robotics field\,\cite{billard2019trends}. The development of human dexterity begins in infancy and is influenced by what the physical environment provides, including the objects available to the child\,\cite{nellis1994review}. As infants and children develop physical and intelligence, they are more likely to attempt complex movements, and often learn dexterity through attempting movements and the consequences of their actions\,\cite{gottlieb2014synthesizing, lockman2000perception, berger2007learning}. Similarly, robot dexterity can not be a constant program pre-set in the laboratory. To acquire the capability of object manipulations in the real world, robots must be able to learn dexterous manipulation skills as if we were infants. As a result, we expect robots to learn to master the ability of dexterous manipulation at the human level from daily tasks. 

Recently, reinforcement learning (RL) algorithms have outperformed human experts in many fields that require decision makings \cite{silver2016mastering, vinyals2019grandmaster}. In contrast to the traditional control methods, RL can complete some challenging tasks in learning dexterous in-hand manipulation\,\cite{zhu2019dexterous, akkaya2019solving, chen2022system} or grasping \cite{geng2022end,li2022gendexgrasp}. However, manipulation that generates changes on the object is still difficult\,\cite{gemici2014learning}. More difficult is generalization across tasks, although previous work can achieve simple level of tasks such as throwing\,\cite{ghadirzadeh2017deep}, sliding\,\cite{shi2017dynamic}, poking\,\cite{agrawal2016learning}, pivoting\,\cite{antonova2017reinforcement}, and pushing\,\cite{woodruff2017planning}, but is still difficult to perform well in unstructured or contact-rich environments, which require the ability to combine and generalize complex manipulation skills. In a nutshell,  reaching human-level sophistication of hand dexterity and bimanual coordination remains an open challenge for modern robotics researchers.

To help solve the problems mentioned above and let robots have the same dexterous manipulation ability as humans, we developed, in the Isaac Gym\,\cite{makoviychuk2021isaac} simulator, a novel benchmark on bimanual dexterous manipulation for RL algorithms along with a diverse set of tasks and objects named \textbf{Bi-DexHands}. We follow the principle of Fine Motor Subtest (FMS)\,\cite{2006Bayley} to design tens of tasks, which provides the opportunities to observe and evaluate specific skills that 
demonstrate a child’s ability to use their hands to play with toys, manipulate objects, and use tools. Next, we tested the baselines of various model-free RL algorithms to show the ability of the baseline algorithm in these tasks, not only the standard RL algorithms but also multi-agent RL (MARL), offline RL, multi-task RL, and Meta RL algorithms, each of them focuses on the bimanual collaboration, learning from demonstration, and task generalization, respectively. Our major goal is to facilitate researchers to master human-level bimanual dexterous manipulations with RL. Not limited to this, we also hope this study to provide a new platform for the community of RL, robotics, and cognitive science. Bi-DexHands are developed with the following key features:
\begin{itemize}
   \item \textbf{Isaac Gym Efficiency}: Building on the Isaac Gym simulator, Bi-DexHands supports running thousands of environments simultaneously.  On one NVIDIA RTX 3090 GPU, Bi-DexHands can reach 30,000+ mean FPS by running 2,048 environments in parallel.
   \item \textbf{Comprehensive RL Benchmark}: We provide the first bimanual manipulation task environment for common RL, MARL, offline RL, multi-task RL, and Meta RL practitioners, along with a comprehensive benchmark for SOTA continuous control model-free RL methods.
   \item \textbf{Heterogeneous-agent Cooperation}: Agents in Bi-DexHands (\textit{i.e.}, joints, fingers, hands,...) are genuinely heterogeneous; this is  different from common multi-agent environments such as SMAC\,\cite{samvelyan2019starcraft} where agents can simply share parameters to solve the task.
   \item \textbf{Task Generalization}: We introduced a variety of dexterous manipulation tasks (e.g., hand over, lift up, throw, place, put...) as well as enormous target objects from the YCB\,\cite{calli2017yale} and SAPIEN\,\cite{xiang2020sapien} dataset, thus allowing meta-RL and multi-task RL algorithms to be tested on the task generalization front.
   \item \textbf{Cognition}: We provided the underlying relationship between our dexterous tasks and the motor skills of humans at different ages. This  facilitates researchers on studying robot skill learning and development, in particular in comparison to humans.
   
\end{itemize}

% \begin{itemize}
%   \item We designed a large number of scenarios on bimanual dexterous manipulation. More importantly, these proposed tasks have been proved be solvable by model-free reinforcement learning algorithms.
%   \item We provided two types of interface, single-agent and multi-agent modes, and implemented the mainstream algorithms respectively. It is worthy of noting that the definition of finger agents makes it possible to evaluate the cooperative level between different fingers.
%   \item we introduced a variety of dexterous manipulation tasks (e.g., handover, lift up, throw, place, put...) as well as enormous target objects from the YCB and SAPIEN dataset (>2,000 objects), thus allowing meta-RL and multi-task RL algorithms to be tested on the task generalization front.
%   \item Thanks to the GPU accelerating simulation on Isaac Gym, our benchmark support running thousands of environments simultaneously, and results illustrate the characteristic benefits for the learning of on-policy algorithms. 
% \end{itemize}

\vspace{-0.3cm}
\section{Related Work}
\vspace{-0.2cm}

% Using a more dexterous hand is one solution to this problem, but at the same time requires more the advanced methods to manipulate it, which remains a difficult challenge in today's robotics field.
% \subsection{Trends and challenges in robot manipulation}
Today, robots are skilled in some repetitive and familiar environments like assembled in the factory. Grasping is a milestone in robotics manipulation. For decades, researchers have been working to establish a stable grasping theory\,\cite{bicchi2000robotic}. However, most previous methods have relied on various assumptions, such as known object information or no uncertainty in the process. In recent years, data-driven approaches have been successful in this regard, being able to deal with uncertainty in perception and generate grasping methods for known, familiar, and even unknown objects in real-time\,\cite{bohg2013data}. Grasping is only a part of the manipulation. Today's robots can perform some simple behaviors like grasping, pushing, and throwing. But it is still difficult to manipulate in unstructured scenes and contact-rich situations. Moving objects while in-hand manipulation is also a complex challenge. One step to address this challenge is to use hands with intrinsic dexterity\,\cite{bircher2017two, rahman2016dexterous}, which often mimic human hands\,\cite{ozawa2017grasp}. Another undeveloped area is bimanual manipulation, a method of using a second hand to provide additional dexterity\,\cite{vahrenkamp2011bimanual, smith2012dual}. Learning for manipulation is important for robots to continuously learn and achieve intelligent control. It is especially suitable for modeling manipulation on complex non-rigid objects and reducing control dimensions\,\cite{nguyen2011model}, but it still suffers from problems such as lack of accurate models, reality gaps, and difficulty in collecting expert data. There are many other robotic manipulation benchmarks\,\cite{yu2020meta, rlbench, robosuite}, but none of them use dexterous hands. Therefore, our work proposes a bimanual dexterous manipulation benchmark, hoping to facilitate researchers to address the challenges of robotic manipulation we mentioned above.

% \subsection{Dexterous manipulation}

Dexterous five-finger hands provide an essential tool to perform a multitude of tasks in human-centric environments. However, such dexterous manipulation remains a challenging problem because of the high dimensional actuation space and contact-rich model. Before the emergence of RL-based controllers, a large variety of manipulation tasks highly relied on accurate dynamics models and trajectory optimization methods\,\cite{kim2021integrated, okamura2000overview, kumar2016optimal}. For example, Williams et al.\,\cite{williams2017information} used the model predictive path integral control (MPPI) method to perform the task successfully, dexterous manipulation of a cube. Charlesworth et al.\,\cite{charlesworth2021solving} improved the MPPI method to make the handing over task between two hands tractable. Since RL simplifies the design process of the controller, model-agnostic approaches have become more and more popular in the field of robotic control\,\cite{Generalization, MaskedV}. In terms of 
dexterous manipulation, many works achieve a significant improvement compared with traditional controllers. OpenAI et al.\,\cite{akkaya2019solving} developed an RL-based controller to reorient a block or a Rubik’s cube. Considering the poor generalization of current approaches, Chen et al.\,\cite{chen2022system} presented an efficient system for learning how to reorient a large number of objects without access to shape information. Qin et al.\,\cite{FromOneHand, DexMV} perform learning from demonstrations for dexterous manipulation collected from teleoperation or video. While their studies demonstrate that RL enables efficient and scalable learning on single-hand manipulation, bimanual manipulation remains a hardship for model-free reinforcement learning\,\cite{charlesworth2021solving}. In this paper, our benchmark provides a wide range of well-designed and challenging daily life scenarios for comprehensive RL algorithms, hoping to help the researcher toward master human-level bimanual dexterous manipulation.

\vspace{-0.3cm}
\section{Formulations $\&$ Algorithms}
 \vspace{-0.2cm}

\begin{wrapfigure}{r}{7cm}
 \centering  %居中
  \vspace{-1cm}
 \includegraphics[width=\hsize]{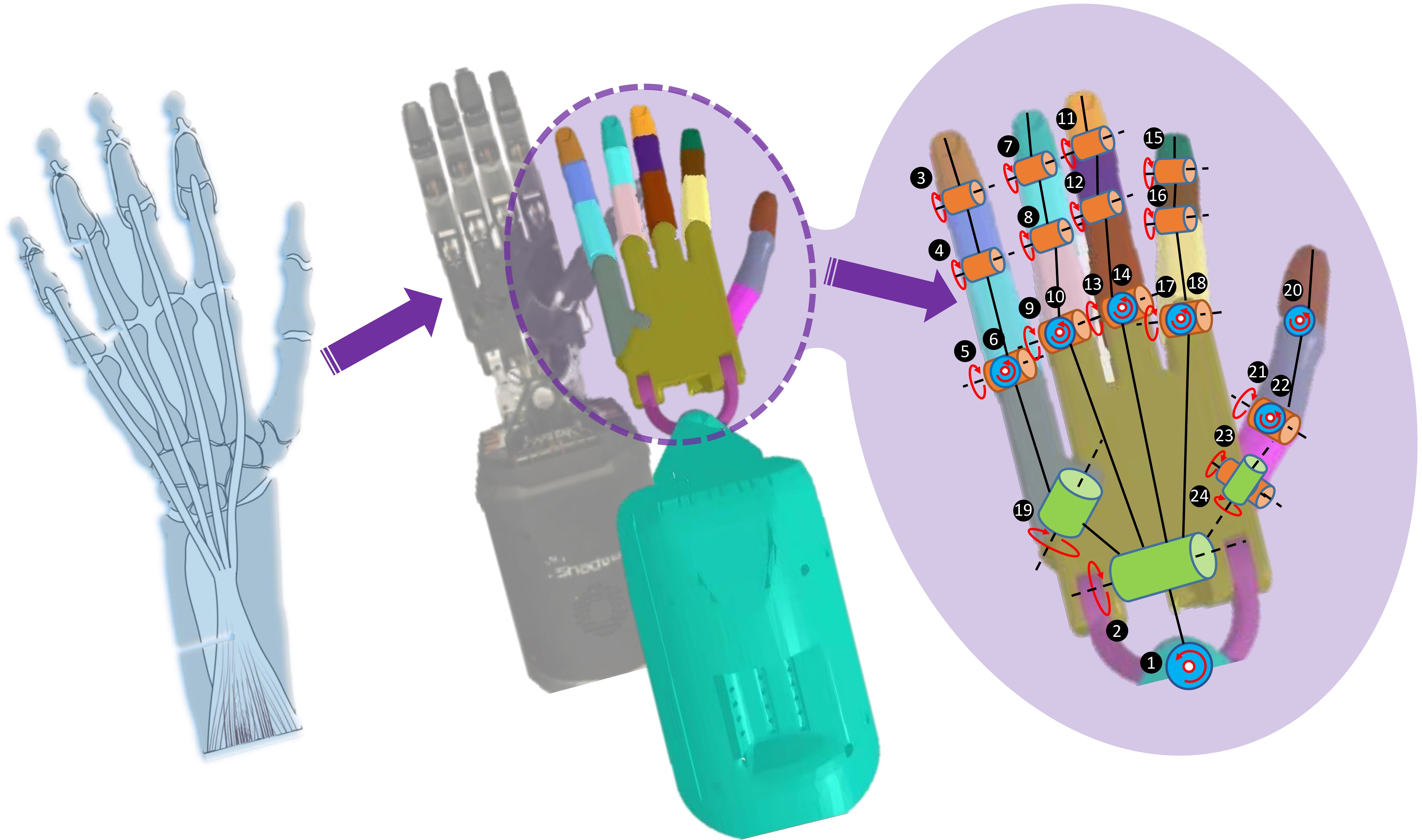} 
 \caption{Degree-Of-Freedom (DOF) configuration of the Shadow Hand similar to the skeleton of a human hand.} 
 \label{fig1}
 \vspace{-0.5cm}
\end{wrapfigure}

In order to create a platform toward mastering human-level dexterity, we use two Shadow Hands to manipulate in our environment. Shadow Hand\,\cite{shadowhand} is a popular robotic hand usually used in some dexterous manipulation tasks. It is designed to resemble the typical human male hand in shape and size, and capable of performing a variety of flexible and delicate operations. Shadow Hand's DoF is shown in Fig.\ref{fig1}, designed to mimic the human skeleton as much as possible. Concretely, the 24-DoF hand is actuated by 20 pairs of agonist-antagonist tendons, while the other four joints remain under-actuated. 

Furthermore, our low-level controller runs at 1k Hz, as well as the RL-based policy outputs the relative positions of actuated joints at 30 Hz. It is worth noting that compared with previous studies, the base of the hand is not fixed in some tasks. Instead, the policy can control the position and orientation of the base within a restricted space, which takes advantage of the function of the wrist, thus making the Shadow Hand more bio-mimetic. Meanwhile, we can efficiently perform the task in real-world applications by linking the base to a robotic arm. Refer to Appendix \ref{Appendix:A1} for more details about the physical parameters of the Shadow Hand.

% We simulated the physical system with the NVIDIA Isaac Gym engine in our released tasks. The core benefits of the Isaac Gym are two-fold. First, different from the other similar engines, such as MuJoCo and Pybullet physics engines, it supports the simulation runs on the GPU platform. In this case, the advantage makes it possible to run the whole program based on GPU tensors, dramatically accelerating the learning process. Secondly, Isaac Gym supports running thousands of environments simultaneously on a single GPU because of the fast operation speed on the GPU. It provides a practical tool for performing our challenging tasks using reinforcement learning algorithms. As we all know, one of the significant limitations of reinforcement learning is low sample efficiency. To our knowledge, the Isaac Gym is the first physics simulation system introducing the GPU accelerated tensor API for environments.

Our benchmark aims at providing solutions for bimanual dexterous manipulation in a comphensive field of RL. To achieve that, We consider five RL formulations including: Single-agent RL, Multi-agent RL (MARL), Offline RL, Multi-task RL, and Meta-RL in Bi-DexHands. 
\begin{comment}
In MARL, we can study the coordination and cooperation between dual dexterous hands, and the multi-task/Meta RL can study the generalization of dual dexterous hands between some well-designed tasks. Meanwhile, the offline RL can study policies from the collected dual dexterous hands data. 
\end{comment}
In the following part, we will introduce the detailed formulation and the corresponding implemented algorithms in our benchmark of these five RL formulations. 

\textbf{RL/MARL.} In order to evaluate the performance of RL/MARL \cite{yang2020overview,deng2021complexity}, we formulate our scenarios as a decentralized partially observable MDP (Dec-POMDP). The Dec-POMDP consists of 10 components, $Z=< \mathcal {N},\mathcal{M},S,\pmb{O},\pmb{A},\pmb{\Pi},P,R,\rho,\gamma>$. Initially, the robotic hands are manually separated as $\mathcal {N}$ agents, the set of which represents $\mathcal{M}$. When starting the simulation, the state of the environment (i.e., the information of robots and objects) is set at $s_0 \in S$ according to the initial distribution of states $\rho(s_0)$. Then at the time step $t$, $s_t$ represents the state, and the i-th agent receives an observation $o^i_t \in \pmb{O}$ relying on $s_t$. Hereafter, the policy of the i-th agent, $\pi_i \in \pmb \Pi$, takes the $o^i_t$ as input, and outputs an action $a^i_t \in A_i$. Additionally, we denote the joint action of all agents by $\pmb{a}_t \in \pmb A$, and the equation $\pmb{A} = [A_1,..A_{\mathcal N}]$ is naturally satisfied. After that, i-th agent can obtain a reward $r^i_t$  based on $R(s_t, \pmb{a_t})$, as well as all agents transitions to the next state $s_{t+1}$ with the possibility of the transition function $P(s_{t+1}|s_t,\pmb{a}_t)$. The goal is to find the optimal policy $\pmb \Pi$ to maximize the sum of rewards $\mathbb{E}_{\pmb \Pi}[\sum_{t=0}^{T-1}\gamma^{t}\sum_{i=1}^{\mathcal{N}}r^{i}_t]$ in an episode with $T$ time steps. It should be pointed out that when $\mathcal{N}=1$, it is the problem formulation of single-agent RL.

In this setting, We implemented state-of-the-art continuous single-agent RL algorithms, such as PPO\,\cite{schulman2017proximal}, SAC\,\cite{haarnoja2018soft}, TRPO\,\cite{schulman2015trust}, DDPG\,\cite{lillicrap2015continuous}, and TD3\,\cite{fujimoto2018addressing} algorithms. Taking our continuous control and fully cooperative environments into consideration, we introduced HAPPO/HATRPO \cite{kuba2021settling,kuba2021trust,kuba2022heterogeneous}, MAPPO\,\cite{yu2021surprising}, IPPO\,\cite{de2020independent}, and MADDPG\,\cite{lowe2017multi} algorithms.

\textbf{Offline RL.} Offline RL follows the formulation of standard MDP, where the goal is to maximize the expected return $\mathbb{E}_{\pi}[\sum_{t=0}^{T-1}\gamma^t{r_t}]$. However, in offline RL, the agent has to learn policy only using the transitions in previously collected dataset $\mathcal{D}=\{(s_t, a_t, s_{t+1}, r_t)\}$, without interacting with the environment. The fundamental challenge of offline RL is value errors of out-of-distribution actions \cite{fujimoto2021minimalist}. We implemented BCQ \cite{fujimoto2019off}, TD3+BC \cite{fujimoto2021minimalist}, and IQL \cite{kostrikov2021offline} algorithms for offline RL.

\textbf{Multi-task RL.} Multi-task reinforcement learning aims to train a single policy $\pi (a|s, z)$, which can achieve good results on different tasks. $z$ represents an encoding of the task ID. The goal of our policy is to maximize the reward given by $\mathbb{E}_{\mathcal{T} \sim p(\mathcal{T})}[\mathbb{E}_{\pi}[\sum_{t=0}^{T-1}\gamma^t{r_t}]$, where $p(\mathcal{T})$ is a task distribution in our benchmark. In practice, multi-task RL adds the context vector corresponding to the type of environment (e.g., one-hot task ID) into states to learn a general skill. We implemented multi-task PPO, multi-task TRPO, and multi-task SAC algorithms for multi-task RL.

\textbf{Meta RL.} Meta RL \cite{liu2021settling}, also known as "learning to learn", aims to gain the ability to train on tasks to extract the common features of these tasks, so as to quickly adapt to new and unseen tasks. In Meta-RL, both training and test environments are assumed to follow the same task distribution $p(\mathcal{T})$. In Bi-DexHands, we design some common structures between different tasks for meta-training to ensure that it can adapt efficiently to new tasks. Compared with Multi-task RL, Meta RL is not allowed to get direct task-level information such as task ID. It needs to solve entirely new tasks by task inference and adaptation purely based on interactions.
\begin{comment}
During meta-training, we select $M$ tasks in the task distribution for meta-policy training. During meta-test, we select $N$ task in the task distribution that has not appeared in meta-training for fine-turning. The meta-policy must quickly adapt to new tasks in a few steps and achieve high performance. 
\end{comment}
We implemented model-agnostic meta learning (MAML)\,\cite{duan2016rl} and proximal meta-policy search (ProMP)\,\cite{rothfuss2018promp} algorithms for Meta RL.

% and fast reinforcement learning via slow reinforcement learning RL${^2}$\,\cite{finn2017model} algorithms for meta reinforcement learning.
\vspace{-0.3cm}
\section{Bimanual dexterous manipulation benchmark}
\label{gen_inst}
\vspace{-0.2cm}

In this section, we will discuss the construction of Bi-DexHands, a benchmark for bimanual dexterous manipulation over diverse scenarios.

\begin{figure}[t]
 \centering  %居中
 \includegraphics[width=\hsize]{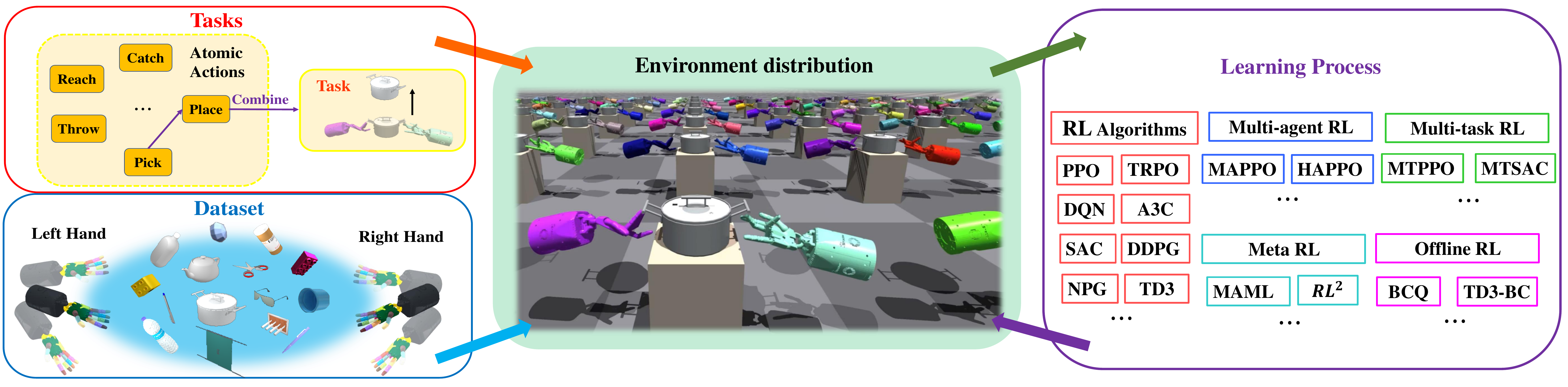} 
 \caption{Framework of Bi-DexHands, a bechmark for learning bimanual dexterous manipulation.}
 \label{fig2}
 \vspace{-0.2cm}
 \end{figure}
 
 \vspace{-0.3cm}
\subsection{System design}
\vspace{-0.2cm}

As we mentioned before, the core of Bi-DexHands is to build up a learning framework for two Shadow Hands capable of diverse skills as humans, such as reaching, throwing, catching, picking and placing. To be specific, Bi-DexHands consists of three components: datasets, tasks, and learning algorithms, as shown in Fig.\ref{fig2}. Varying worlds provide a large number of basic settings for robots, including the configuration of robotic hands and objects. Meanwhile, a variety of tasks corresponding to children's behaviors at different ages make it possible to learn dexterity like a human. Combining a dataset and task, we can generate a specific environment or scenario for the following learning. Eventually, our experiments demonstrate that reinforcement learning is able to facilitate the robots to achieve some remarkable performance on such challenging tasks, and there is still some room for improvement and more difficult tasks for future work.
\vspace{-0.3cm}
\subsection{Construction of datasets}
\vspace{-0.2cm}

% Interestingly, we noticed that the relative pose between two hands changes a lot, such that the robots make the different demands of skills for accomplishing the task. That means the initial configuration of two hands has a crucial impact on the following learning. Thus, to better increase the number of worlds, we designed different poses of hand, like abreast, underarm and overarm settings. Furthermore, the bases of hands should be chosen as a fixed or actuated mode

The construction of the datasets corresponds to the configuration of robots and objects. The core goal of datasets is to generate a large variety of scenes for robot learning. As we mentioned in the last part, the robots in our benchmark are two dexterous Shadow Hands. Other than the robots, the objects also play an essential role in constructing the datasets. For extending the types of tasks, we introduced a variety of objects from the YCB \cite{calli2017yale} and SAPIEN \cite{xiang2020sapien} datasets. Two datasets contain many everyday objects. Notably, the SAPIEN dataset provides many articulated objects with motion annotations and rendering material, which means these objects are close to the real ones significantly. Therefore, it provides a natural way to build a connection between the worlds of our benchmark and scenes of daily life. Concretely, Fig.\ref{fig2} shows the construction of datasets, and we can see that the object includes pots, pens, eggs, scissors, eyeglasses, doors, and other common tools. After defining the configuration of robots and the type of objects, we build the specific world based on the Isaac Gym simulator. Meanwhile, each world defines variable initial poses of robots and objects, providing a diverse set of environments.
\vspace{-0.3cm}
\subsection{Design of tasks}
\vspace{-0.2cm}

An infant's behavior experiences a multi-stage development, such as social, communication, and physical parts\,\cite{del2011development}. Particularly in bimanual dexterous manipulation, there are some relationships between some common behaviors of babies and the ages. To gain insights into the underlying relationships, we conducted an in-depth analysis and built a mapping between the baby's age and tasks according to the Fine Motor Subtest (FMS)\,\cite{2006Bayley}. As the baby's age increases, the difficulty of completing the designed tasks also increases, because the baby can complete more and more difficult behaviors as the body develops. So it is also of great importance to evaluate the performance of trained agents, because we can roughly point out agents' intelligence level by analogy with a baby's movement for bimanual dexterous manipulation. An overview of the correspondence of our tasks to the FMS is shown in Table.\ref{task_desci}. For more details on the tasks, please refer to Appendix~\ref{Appendix:A2}.

\begin{table}[htbp]
    \centering
    \setlength\tabcolsep{4pt}
    \caption{Task name and the description of the human skill in the corresponding age. References under the human age are the cognitive science literature referenced for the behavior designed, and the difficulty level of the tasks is under the task name. Easy level tasks are more basic skills, medium level tasks need more precise control and finger dexterity, and hard level tasks require handing dynamic interaction and tool use.}
    \begin{tabular}{ccccc}
    
    \toprule[2pt]  %添加表格头部粗线
\textbf{Human Task Name} & \makecell[c]{\textbf{Human’s Skill Description}} & \makecell[c]{\textbf{Age (months)}} & \textbf{Demo} \\
    \bottomrule[2pt]
    \makecell[c]{Push Block \\ Easy}  & \makecell[c]{ Child extends one or both arms\\ forward and touches the block with\\ any part of either hand}& \makecell[c]{5-6 \\ \cite[Chapter 3]{2006Bayley}} &\begin{minipage}[b]{0.2\columnwidth}
		\raisebox{-.5\height}{\includegraphics[width=\linewidth]{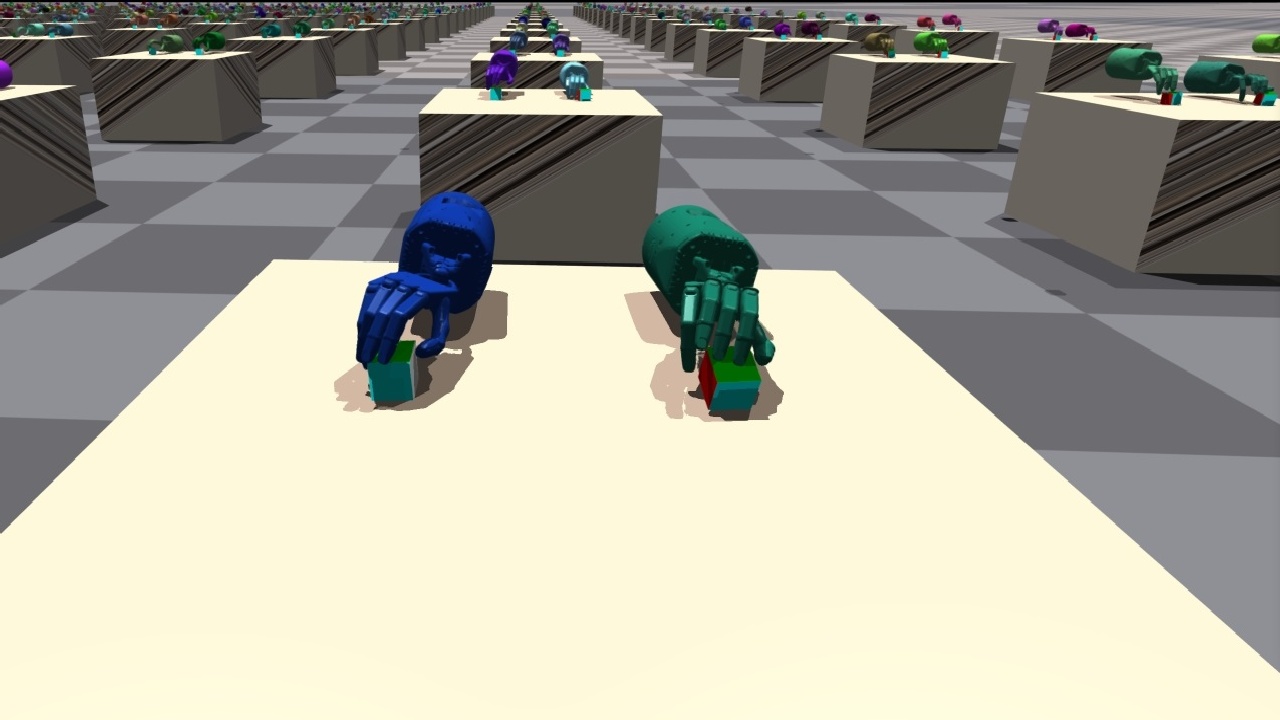}} 
	\end{minipage}\\\hline
    
    \makecell[c]{Open Scissor \& \\
Open Pen Cap \\ Easy} & \makecell[c]{They use one hand to hold \\ a toy and the other hand \\ manipulate it}& \makecell[c]{7 \\ \cite[Chapter 4]{weiss2010bayley}}&\begin{minipage}[b]{0.2\columnwidth}
		\raisebox{-.5\height}{\includegraphics[width=\linewidth]{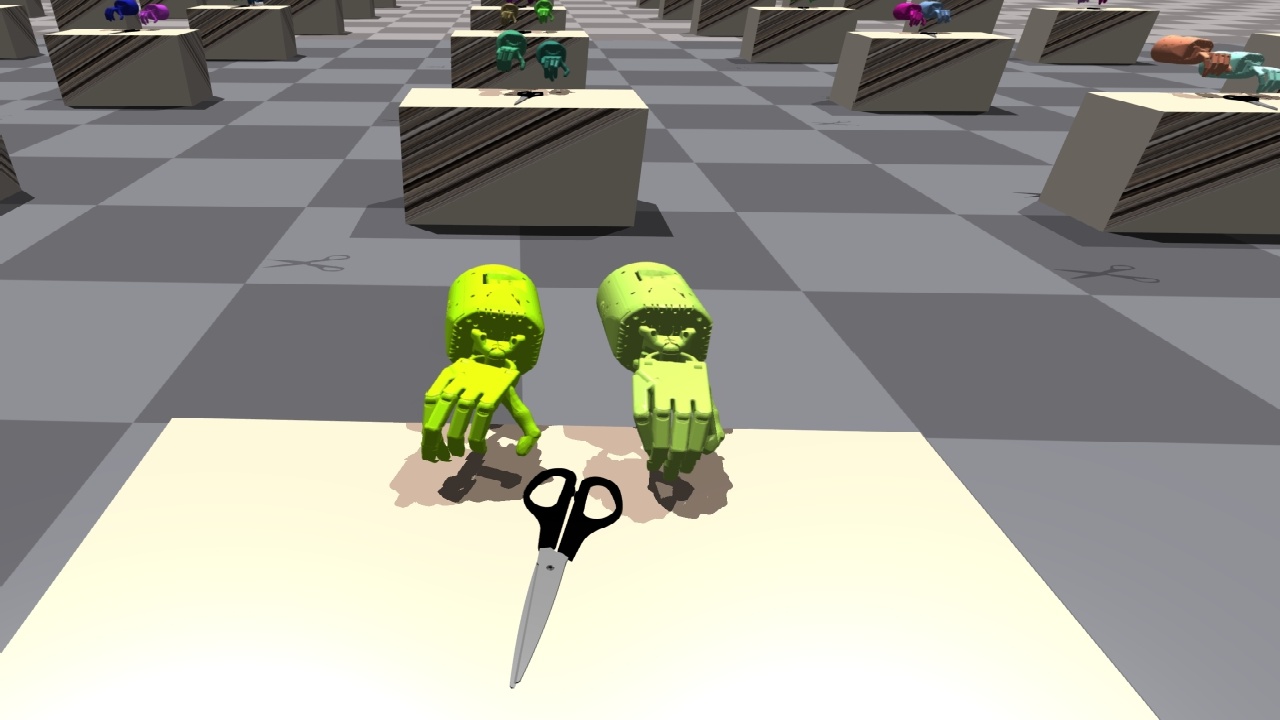}} 
	\end{minipage}\\\hline

%     \makecell[c]{Hand Over \\ (Underarm \& \\ Over2underarm \& \\ Abreast)} & \makecell[c]{ Child transfers the ring from hand  \\ to hand}& \makecell[c]{9-10 \\ \cite[Chapter 3]{2006Bayley} \\Easy} 
% &\begin{minipage}[b]{0.2\columnwidth}
% 		\raisebox{-.5\height}{\includegraphics[width=\linewidth]{results/static_image/hand_over.jpg}} 
% 	\end{minipage}\\\hline
     
     \makecell[c]{Turn Button \\ ON/OFF \\ Easy} & \makecell[c]{They can push and squish soft   \\ stuff or push hard things, like a\\ button on a toy phone or popup toy}& \makecell[c]{11 \\\cite[11 months]{FWP_2018}}&
     \begin{minipage}[b]{0.2\columnwidth}
		\raisebox{-.5\height}{\includegraphics[width=\linewidth]{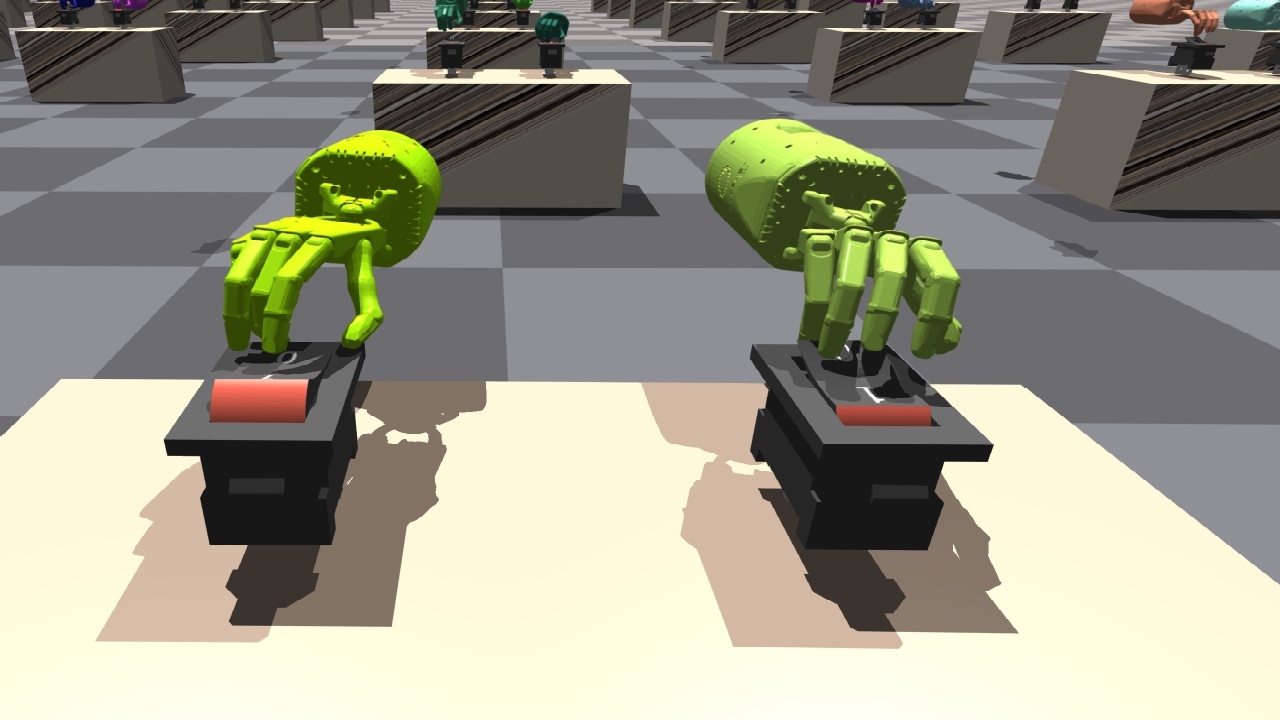}} 
	\end{minipage}\\\hline
     
     \makecell[c]{Swing Cup \\ Easy} & \makecell[c]{ They can turn a ball on their toy \\ mobile, a steering wheel on a toy  \\ car, or the faucet in the tub} & \makecell[c]{11 \\ \cite[11 months]{FWP_2018}}&
     \begin{minipage}[b]{0.2\columnwidth}
		\raisebox{-.5\height}{\includegraphics[width=\linewidth]{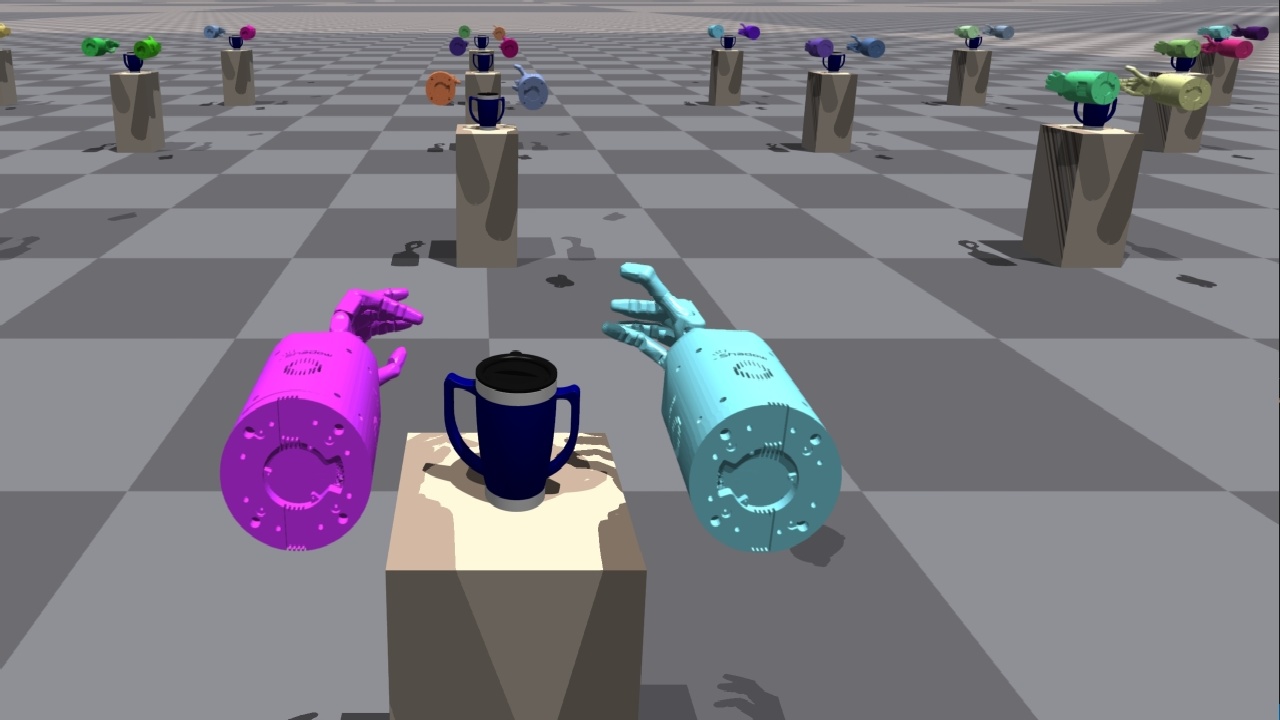}} 
	\end{minipage}\\\hline
     
     \makecell[c]{Lift Pot \&\\Lift Cup \\ Easy} & \makecell[c]{ They can put a sippy cup to their  \\ mouth to drink}& \makecell[c]{12 \\\cite[12 months]{FWP_2018}}&\begin{minipage}[b]{0.2\columnwidth}
		\raisebox{-.5\height}{\includegraphics[width=\linewidth]{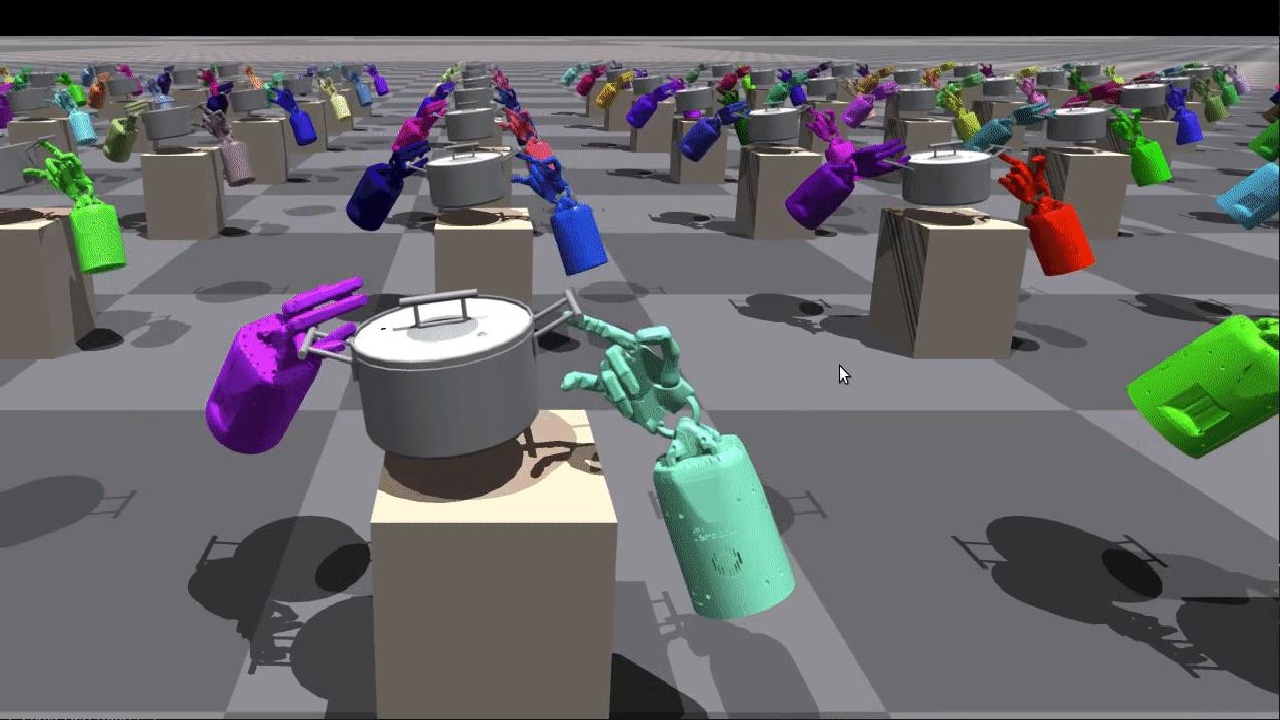}} 
	\end{minipage}\\\hline

     \makecell[c]{Door Open \&\\ Close \\ Easy} & \makecell[c]{ Toddlers can open and close\\ cupboards and oven doors}& \makecell[c]{13 \\ \cite[13 months]{FWP_2018}}&\begin{minipage}[b]{0.2\columnwidth}
		\raisebox{-.5\height}{\includegraphics[width=\linewidth]{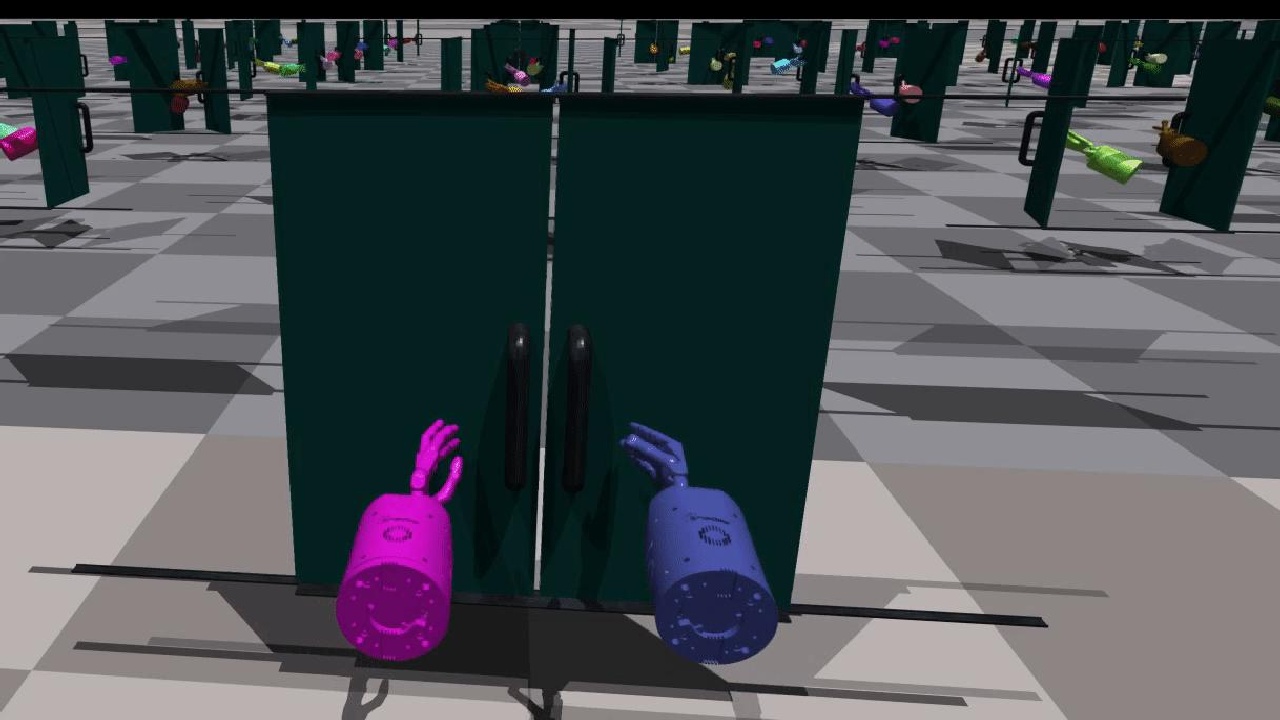}} 
	\end{minipage}\\\hline

     \makecell[c]{Re-Orientation \\ Medium} & \makecell[c]{Infant further refines this ability\\ to differentiate individual finger\\ movement and manipulate objects}& \makecell[c]{18 \\ \cite[Chapter 4]{weiss2010bayley}}&\begin{minipage}[b]{0.2\columnwidth}
		\raisebox{-.5\height}{\includegraphics[width=\linewidth]{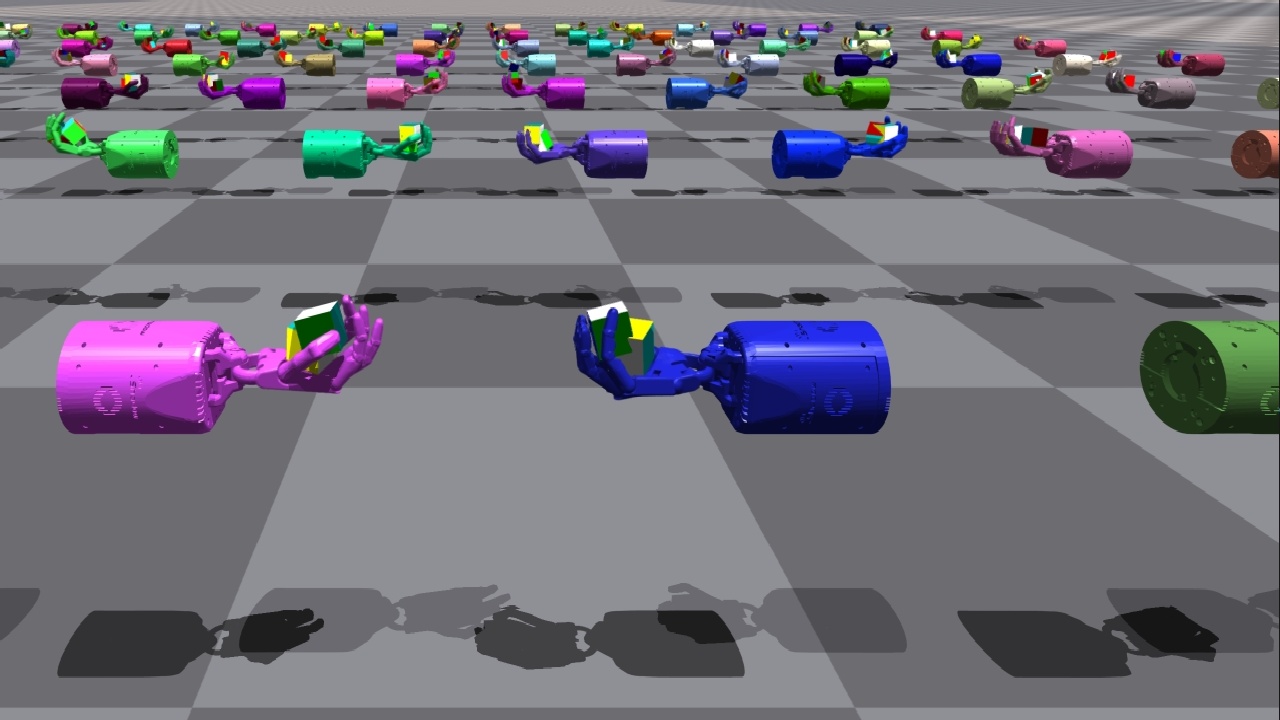}} 
	\end{minipage}\\\hline

     \makecell[c]{Stack Block(2,6,8) \\ Medium} & \makecell[c]{ Child stacks at least 2/6/8 blocks  \\in any trial.}& \makecell[c]{2:22-28\\ 6,8:33-42 \\ \cite[Chapter 3]{2006Bayley}}&\begin{minipage}[b]{0.2\columnwidth}
		\raisebox{-.5\height}{\includegraphics[width=\linewidth]{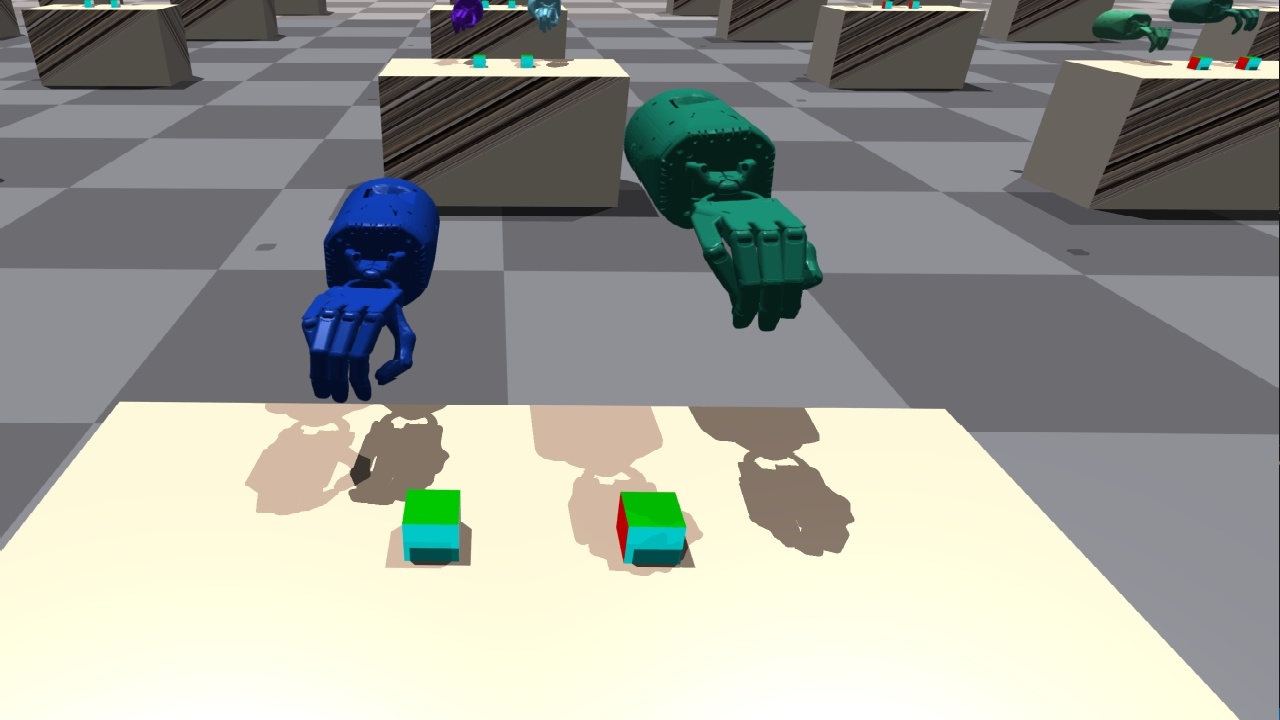}} 
	\end{minipage}\\\hline
     
     \makecell[c]{Pull a Ball into \\Bucket \\ Medium} &  \makecell[c]{ Child place s 10 pellets in the \\bottle in 60 seconds or less, one \\pellet at a time.}& \makecell[c]{22-28\\ \cite[Chapter 3]{2006Bayley}}&\begin{minipage}[b]{0.2\columnwidth}
		\raisebox{-.5\height}{\includegraphics[width=\linewidth]{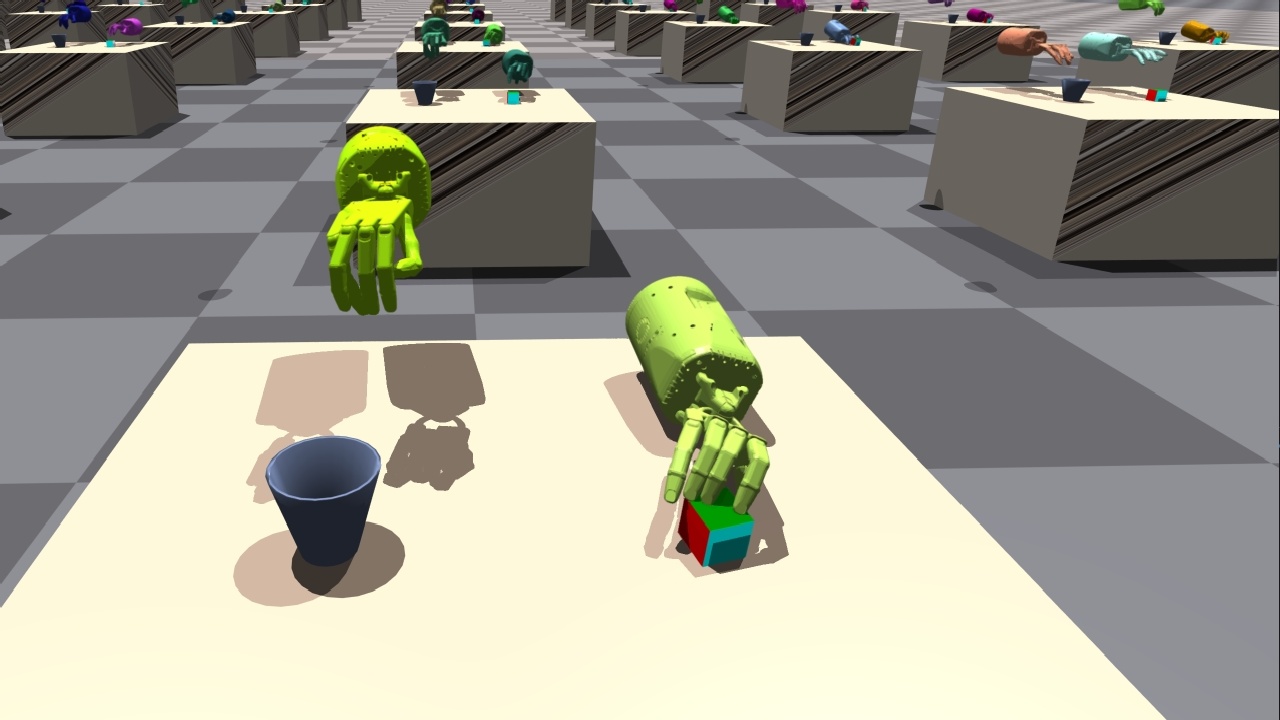}} 
	\end{minipage}\\\hline
     
     \makecell[c]{Open Bottle Cap \\(prismatic joint)\\ Medium} & \makecell[c]{\ Uses hands to twist things, like \\turning doorknobs or unscrewing\\lids }& \makecell[c]{30 \\ \cite[Table 6]{zubler2022evidence}}&\begin{minipage}[b]{0.2\columnwidth}
		\raisebox{-.5\height}{\includegraphics[width=\linewidth]{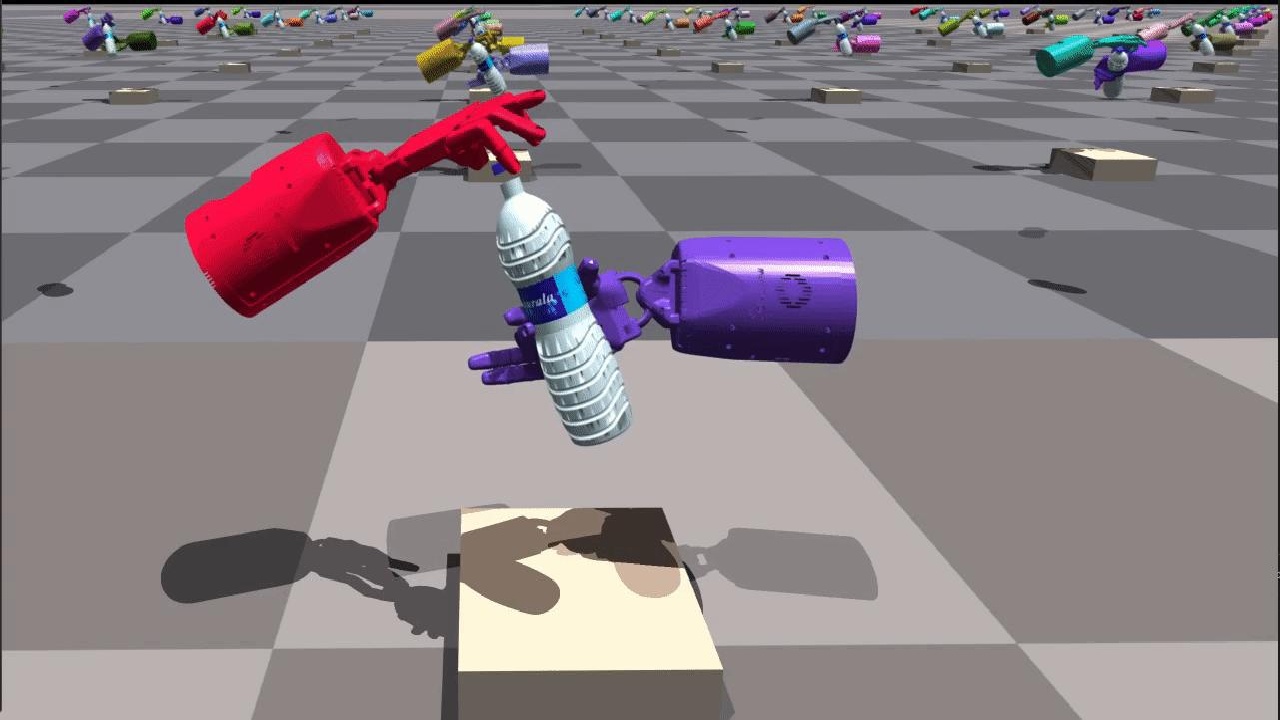}} 
	\end{minipage}\\\hline
     
     \makecell[c]{Catch Underarm\\Hard}  & \makecell[c]{\ Catches a large ball most of the \\time}& \makecell[c]{48\\ \cite[Table 6]{zubler2022evidence}} &\begin{minipage}[b]{0.2\columnwidth}
		\raisebox{-.5\height}{\includegraphics[width=\linewidth]{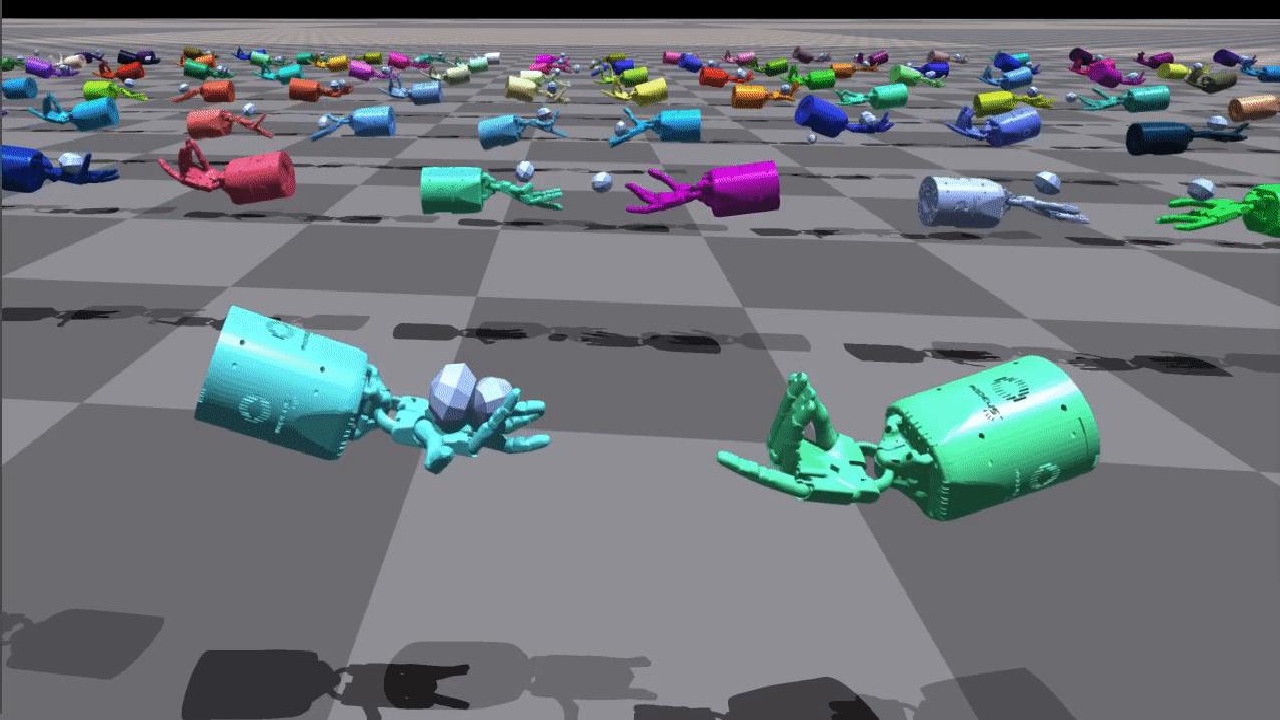}} 
	\end{minipage}\\\hline
	
     \makecell[c]{Pour Water \\Hard} & \makecell[c]{\ Serves himself food or pours \\water, with adult supervision}& \makecell[c]{48\\ \cite[Table 6]{zubler2022evidence}}&\begin{minipage}[b]{0.2\columnwidth}
		\raisebox{-.5\height}{\includegraphics[width=\linewidth]{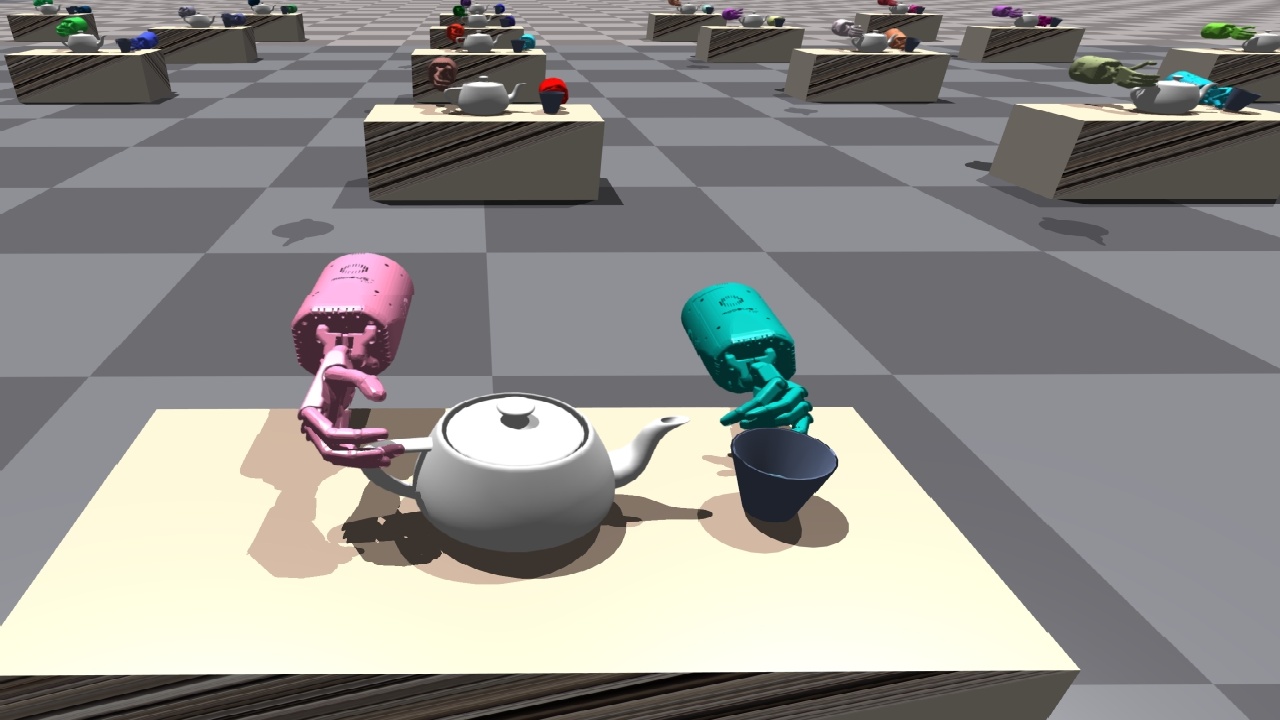}} 
	\end{minipage}\\\hline

     \makecell[c]{Two Catch \\Underarm \\ Hard} & \makecell[c]{Some adults can throw objects\\ between two hands like magic}& \makecell[c]{adult}&\begin{minipage}[b]{0.2\columnwidth}
		\raisebox{-.5\height}{\includegraphics[width=\linewidth]{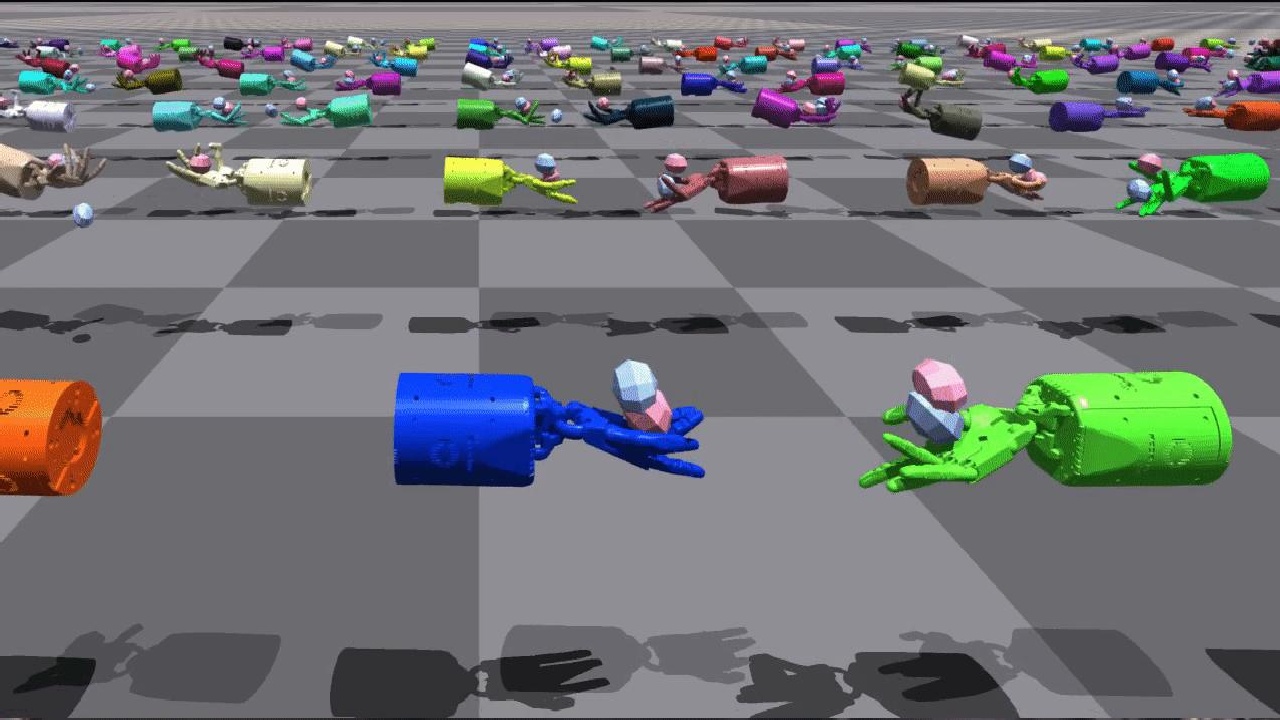}} 
	\end{minipage}\\
    \bottomrule[2pt] %添加表格底部粗线
    \end{tabular}
    \label{task_desci}
\end{table}

\vspace{-0.3cm}
\subsection{Design of Multi-task/Meta RL}
\vspace{-0.2cm}

The design of our Multi-task/Meta RL categories is generally similar to Meta-World \cite{yu2020meta}, divided into ML1, MT1, ML4, MT4, ML20, and MT20. Each of our tasks has object variation, which as we can interact with different kinds of objects in daily life scenes, providing a foundation for us to learn dexterous manipulation like humans. In the following, we will introduce 6 tasks categories for Multi-task/Meta-RL. More details can refer to Appendix~\ref{Appendix:D}.

\textbf{MT1\&ML1: Learning a multi-task policy \& Few-shot adaptation within one task:} Both ML1 and MT1 are categories for generalization ability within the same task, and their generalization ability is reflected in the ability to complete tasks under different goals. ML1 uses meta-reinforcement learning for few-shot adaptation, in which goal information will not be provided. MT1 uses the multi-task method for generalization, and the information on the goal will be provided in a fixed set.

\textbf{MT4\&MT20: Learning a multi-task policy belonging to 4\&20 training tasks:} MT4 and MT20 conduct policy training in 4\&20 tasks and hope to complete all tasks in only one policy. In MT4, we hope to learn policy with similar human skills, so we try to combine similar tasks as much as possible. MT20 uses all of our 20 tasks. In MT4 and MT20, we use a one-hot task ID to represent different tasks, and the information on the goal will be provided in a fixed set.

\textbf{ML4\&ML20: Learning a Few-shot adaptation for new 1\&5 test tasks from 3\&15 training tasks:} ML4 and ML20 are categories for learning meta-policies in 3\&15 tasks respectively and hoping to adapt to new 1\&5 testing tasks. There is no doubt that this is a  difficult challenge. We choose the tasks which using the catch behavior for design in ML4. The ML20 requires adaptation in all 15 tasks with large differences designed according to baby intelligence, which is the most difficult challenge in our benchmark. Similarly, we will variate the goal for each task, and will not provide task information, requiring Meta RL algorithms to identify the tasks.

\vspace{-0.2cm}
\section{Benchmarking reinforcement learning algorithms}
\vspace{-0.1cm}

In this section, we conduct a full benchmark of the RL algorithms in Bi-DexHands. We firstly quantify our environment speed to demonstrate the running efficiency of Bi-DexHands. Then We offer the benchmark results and corresponding discussion and analysis on those five RL formulations.
\begin{comment}
Next, we will show the results of RL, MARL and offline RL. Finally, we will summarize the generalization ability of multi-task RL and Meta RL in Bi-DexHands.
\end{comment}
All of our experiments are run with Intel i7-9700K CPU @ 3.60GHz and NVIDIA RTX 3090 GPU. For the hyperparameters of all algorithms, please refer to Appendix~\ref{Appendix:B}.

\vspace{-0.3cm}
\subsection{Environmental speed}
\vspace{-0.2cm}

Thanks to Isaac Gym's high-performance GPU parallel simulating capabilities, we can greatly improve the sampling efficiency of our RL algorithm while using fewer computing resources. We believe that the high sampling efficiency improves the exploration ability of the RL algorithm, allowing us to successfully learn the bimanual dexterous manipulation policy. To demonstrate the Isaac Gym's efficiency of Bi-DexHands, We provided some results of environmental speed in Table.\ref{speed} by running on-policy algorithms. Both PPO and HAPPO can achieve more than 20k FPS.
\vspace{-0.2cm}
\begin{table}[htbp]
    \centering
    \caption{Mean and standard deviation of FPS (frame per second) of the environments in Bi-DexHands.}
    \vspace{-0.10cm}
    \begin{tabular}{ccccc}
    \toprule[1.2px]  %添加表格头部粗线
    Algorithms & CatchUnderarm & CatchOver2Underarm & CatchAbreast & TwoCatchUnderarm\\\hline
    PPO  & $35554\pm613$ & $35607\pm344$ & $35164\pm450$ & $32285\pm898$\\
    HAPPO & $23929\pm98$ & $23827\pm135$ & $23456\pm255$ & $23205\pm168$\\
    \bottomrule[1.2px] %添加表格底部粗线
    \label{speed}\\
    \end{tabular}
    \vspace{-0.5cm}
\end{table}

\vspace{-0.2cm}
\subsection{RL/MARL results}
\vspace{-0.1cm}

\begin{figure}[t]
 \centering  %居中
 \includegraphics[width=1\hsize]{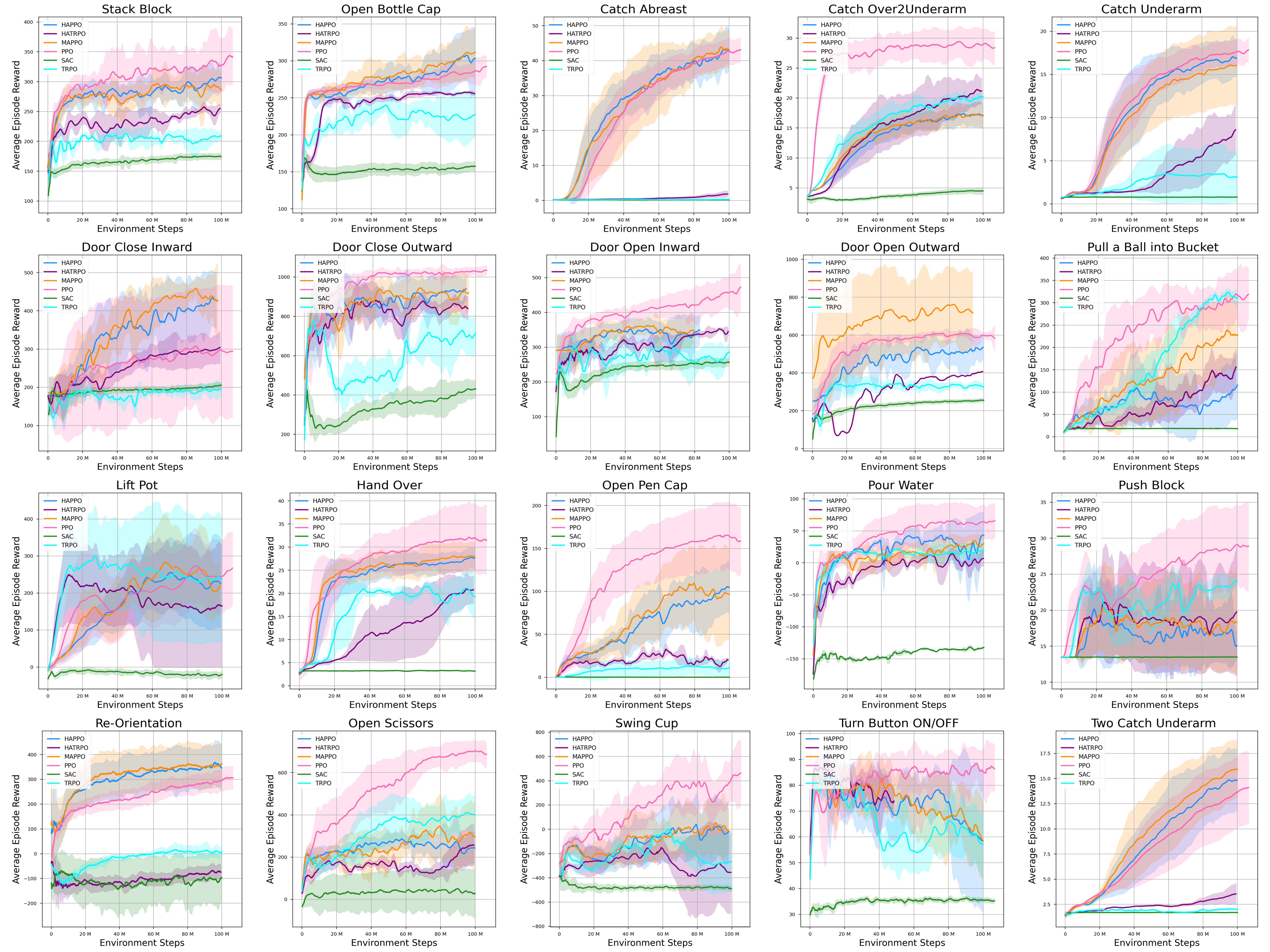} 
 \caption{Learning curves for all 20 tasks. The shaded region represents the standard deviation of the score over 10 trials. Curves are smoothed uniformly for visual clarity. All algorithms interact with environments in 100M steps and the number of parallel simulations is 2048.} 
 \label{fig_merge}
 \vspace{-0.5cm}
\end{figure}

Currently, we evaluate the performance of PPO, SAC, TRPO, MAPPO, HATRPO and HAPPO algorithms on these 20 tasks, and we implemented the rest of the RL/MARL algorithms in our Github repository. The performance of each algorithm are shown in Figure~\ref{fig_merge}. Note that the experiments of MARL algorithms run based on two agents, which means each hand represents an agent. It can be observed that the PPO algorithm performs well on most tasks. Although there are some tasks that require two-hand cooperation, PPO algorithm is still better than HAPPO, MAPPO algorithms in most cases. This may be because PPO algorithm is able to use all observations for training the policy, while MARL can only use partial observations. However, in most tasks, the more difficult and require the cooperation of both hands, the smaller the performance gap between PPO and HAPPO, MAPPO, indicating that the multi-agent algorithm can improve the performance of bimanual cooperative manipulation. Another finding is that the SAC algorithm does not work on almost all tasks. It may be due to 1) the off-policy algorithm has a lower improvement in high sampling efficiency than on-policy. 2) The policy entropy of SAC brings instability to policy learning under the high-dimension input. We discuss this finding in detail in Appendix~\ref{Appendix:C}.

\vspace{-0.2cm}
\subsection{Offline RL results}
\vspace{-0.2cm}

We build offline datasets with four datatypes, \textit{i.e.}, random, replay, medium, and medium-expert. The data collection follows that in D4RL-MuJoCo \cite{d4rl}, which is a standard offline benchmark, and the details are given in Appendix~\ref{app:collection}. We evaluate behavior cloning (BC), BCQ \cite{fujimoto2019off}, TD3+BC \cite{fujimoto2021minimalist}, and IQL \cite{kostrikov2021offline} on two tasks, Hand Over and Door Open Outward, and report normalized scores in Table~\ref{tab:offline}. BCQ and TD3+BC could obtain significant performance improvement compared with behavior policy (BC). However, the action space and state space in Bi-DexHands are much larger than that in MuJoCo, which means the problem of out-of-distribution action \cite{fujimoto2021minimalist} is more severe in Bi-DexHands datasets. That is the reason why IQL could only achieve performance improvement in several datasets. Due to the potential large distribution shift, we believe Bi-DexHands can be a more challenging and meaningful offline benchmark for offline RL research.

\vspace{-0.20cm}
\begin{table*}[h]
		\centering
		\caption{Normalized score in offline tasks.}
		\vspace{0.10cm}
		\begin{small}
			\begin{tabular}{ccccccc}
				\toprule
			Tasks	&       Datasets     & Online PPO     & BC     &      BCQ & TD3+BC & IQL \\\midrule
				\multirow{4}{*}{Hand Over} & random         & $100.0$  & $0.7 \pm 0.2$        & $1.0 \pm 0.1$ & $0.9 \pm 0.2$ & $0.7 \pm 0.4$\\
				& replay         & $100.0$  & $17.5 \pm 3.5$  & $61.6 \pm 4.9$&$\bm{70.1} \pm 2.1$ & $43.1 \pm 2.3$\\
				& medium & $100.0$    & $61.6 \pm 1.0$        & $\bm{66.1} \pm 1.9$ & $\bm{65.8} \pm 2.2$ & $57.4 \pm 1.5$\\
				& medium-expert & $100.0$    & $63.3 \pm 1.4$        & $\bm{81.7} \pm 4.9$ & $\bm{84.9} \pm 5.3$ & $67.2 \pm 3.6$\\ \midrule
				\multirow{4}{*}{\makecell[c]{Door\\Open Outward}} & random         & $100.0$    & $2.1 \pm 0.6$      & $23.8 \pm 2.9$ & $\bm{34.9} \pm 4.3$ & $3.8 \pm 1.0$\\
				& replay         & $100.0$  & $36.9\pm 4.3$       & $48.8 \pm 4.5$ & $\bm{60.5} \pm 2.6$ & $31.7 \pm 2.0$\\
				& medium & $100.0$   & $63.9 \pm 0.7$       & $60.1 \pm 2.3$ & $\bm{66.3} \pm 0.7$ & $56.6 \pm 1.2$\\
				& medium-expert & $100.0$  & $69.0 \pm 6.4$       & $\bm{73.7} \pm 4.5$ & $\bm{71.9} \pm 3.5$ & $53.8 \pm 1.8$\\
				\bottomrule
			\end{tabular}
		\end{small}
		\vspace{-0.2cm}
		\label{tab:offline}
	\end{table*}
	
\subsection{Generalization ability }
\vspace{-0.1cm}

% In order to achieve its full potential of a large variety of tasks and goals, we introduced some multi-task/Meta RL algorithms to enhance the generalization ability. Therefore, the trained agents have the ability of accomplishing multiple tasks at the same time. We implemented multi-task PPO, TRPO and SAC algorithms. Compared with traditional algorithms, multi-task RL methods take an one-hot encoding vector corresponding to the type of each environment as input in the observation. 

The goals of our generalization evaluation is 1) to find out the ability of current multi-task and meta reinforcement learning algorithms to generalize on the tasks we designed. 2) to find out whether the tasks that are harder for babies are also harder for RL. The previous RL/MARL results have proved that our individual task is solvable.
For goal 1), we evaluate the multi-task PPO \cite{schulman2017proximal} and ProMP \cite{rothfuss2018promp} algorithms on MT1, ML1, MT4, ML4, MT20, and ML20. We also provided the results of random policy and using the PPO algorithm in individual task as the ground truth for comparison. The average reward for each training is shown in Table~\ref{tab:meta}. We can observe that the multi-task PPO does not perform well, and the ProMP have  tiny performance improvement compared with random policy. It may because it's hard to learn policy from individually each task itself in Bi-DexHands. Therefore, we still have a lot of room to improve the generalization ability of bimanual dexterous hands under cross-task setting, which is a meaningful open challenge for the community.

\vspace{-0.30cm}
\begin{table*}[h]
		\centering
		\caption{The average reward of all tasks for MT1, ML1, MT4, ML4, MT20, and ML20 on 10 seeds.}
		\vspace{0.10cm}
		\begin{small}
			\begin{tabular}{ccccccccccc}
				\toprule
			\multirow{2}{*}{Method}	&       \multirow{2}{*}{MT1}     &  \multirow{2}{*}{MT4}  & \multirow{2}{*}{MT20}     &  \multirow{2}{*}{Method} & \multicolumn{2}{c}{ML1} & \multicolumn{2}{c}{ML4} & \multicolumn{2}{c}{ML20} \\
			
			\cline{6-7}
			\cline{8-9}
			\cline{10-11}
			
                &    &    &      &   & train&test & train&test & train&test \\\midrule
				Ground Truth & 15.2 & 24,3  & 32.5& Ground Truth & 15.0 & 15.8& 28.0& 13.1& 33.7& 26.1\\
				Multi-task PPO &  9.4 & 5.4  & 8.9& ProMP & 0.95 & 1.2& 2.5& 0.5& 0.02& 0.36\\
				Random  &  0.61 & 1.1  & -2.5& Random & 0.59 & 0.68& 1.5& 0.24& -2.9& 0.27\\
				\bottomrule
			\end{tabular}
		\end{small}
		\vspace{-0.3cm}
		\label{tab:meta}
	\end{table*}

\begin{figure}[ht]
 \centering  %居中
 \includegraphics[width=1\hsize]{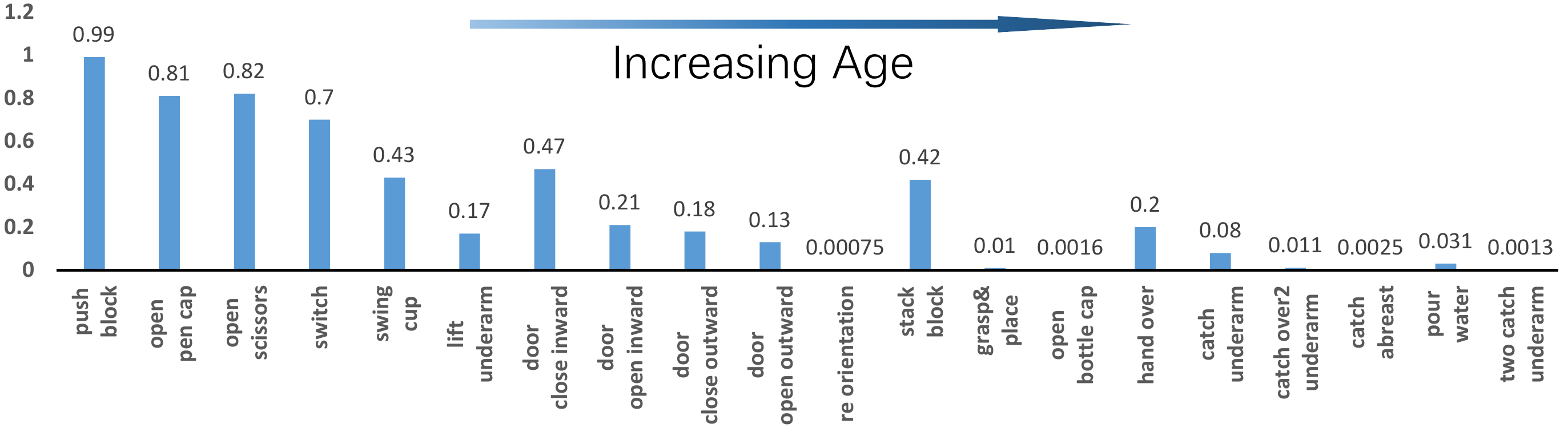} 
 \caption{The normalized reward run by the MTPPO algorithm under the MT20 setting. The tasks from left to right according to the increase of corresponding age. The normalized score is computed by $\text{score}=\frac{\text { reward-random reward }}{\text { ground truth reward-random reward }}$.}
 \label{Age-task}
 \vspace{-0.2cm}
\end{figure}

For goal 2), we use random and ground truth reward to normalize the results of all tasks in MT20 and arrange them in the order of increasing age. The results is shown in Fig.\ref{Age-task}. It can be seen that in general, as the age of the person corresponding to the task increases, the difficulty for RL also increases, which proves that our task design is designed with rationality and relevance to people.

\vspace{-0.2cm}
\section{Conclusion and Future Work}
\vspace{-0.1cm}

We introduced a benchmark, Bi-DexHands, which consists of well-designed tasks and a large variety of objects for learning bimanual dexterous manipulation. We investigated the motor development process of infants’ dexterity from cognitive science, and carefully designed more than twenty tasks for RL based on the results, hoping that robots can learn dexterity like humans. With the help of the Isaac Gym simulator, it can run thousands of environments in parallel, improving the sample efficiency for RL algorithms. Moreover, the implemented RL/MARL/offline RL algorithms achieve superior performance on tasks with simple manipulation skills required. Meanwhile, complex  manipulations still remain challenging. In particular, when the agent is trained  to master multiple manipulation skills, the results of multi-task/Meta RL are not satisfactory. Interestingly, we found that under the multi-task setting, RL exhibited results associated with the development of human intelligence, that is, the trend of RL performance matches with the development of human ages. So far, in bimanual dexterous robot hand manipulation, the current reinforcement learning can reach the level of 48-months infants. 

However, we think that the limitation of Bi-DexHands is that it does not support the deformable object manipulation tasks. Dexterous hands have unique advantages in manipulating deformable objects, but our tasks currently only cover articulated rigid body object manipulation. We hope to develop in this direction in the future. Another limitation is that our tasks primarily train with policies with a state-based observation space, which is difficult for sim-to-real transfer because such inputs are not available in the real world. Our work will advance the field of robotics, increasing the level of automation in factories or lives to replace human labor. The development of this field will reduce the need to put humans in dangerous situations and improve the quality of human life, but it will also bring about the potential for worker displacement.

We identify four main future directions toward mastering human-level bimanual dexterous manipulation. \textbf{1)} Learning from demonstration: our platform needs some human teaching data to study learning from demonstration.  \textbf{2)} Soft body and deformable objects simulation: we need a better physics engine to support our research on software and task design, to be more specific daily life scenes. \textbf{3)} Current meta/multi-task RL algorithms are unable to perform all tasks in our benchmark successfully, which calls for substantial further development on the algorithmic design end. \textbf{4)} We would like address the sim-to-real gap by transferring the simulation result on real dexterous hands. In particular, we hope our benchmark results can serve as a start point to help researchers transfer RL-learned skills to reality and help real-world robots to learn dexterous manipulation.

% \vspace{-0.2cm}
% \begin{ack}
% Use unnumbered first level headings for the acknowledgments. All acknowledgments
% \end{ack}
% \vspace{-0.2cm}

\normalsize
%\printbibliography
\bibliographystyle{unsrt}
\bibliography{name.bib}

%%%%%%%%%%%%%%%%%%%%%%%%%%%%%%%%%%%%%%%%%%%%%%%%%%%%%%%%%%%%
\newpage
\appendix

\section{Task Specifications}

\subsection{Physical parameters of Shadow Hand}\label{Appendix:A1}
 
The limits of each joint in Shadow Hand are as Table \ref{joint_limit}. The thumb has 5 degrees of freedom with 5 joints, the other fingers are all 3 degrees of freedom and 4 joints, and the joints at the ends of each finger are uncontrollable. The distal joints of the fingers are coupled like that of human fingers, making the angle of the middle joint always bigger or equal to the angle of the distal joint. This allows the middle phalange is curved, while the distal phalange is straight. There is an extra joint (LF5) at the end of the little finger to allow the little finger to rotate in the direction of the thumb. There are two joints at the wrist, which guarantees that the entire hand can rotate 360 degrees. 

\begin{table}[htbp]
    \centering
    \caption{Finger range of motion.}
    \begin{tabular}{cccc}
    \toprule  %添加表格头部粗线
    Joints &  Corresponds to the number of \ref{fig1}& Min & Max\\\hline
    Finger Distal (FF1,MF1,RF1,LF1) & 15, 11, 7, 3 &  0° & 90°\\\hline
    Finger Middle (FF2,MF2,RF2,LF2) & 16, 12, 8, 4 &  0° & 90°\\\hline
    Finger Base Abduction (FF3,MF3,RF3,LF3) & 17, 13, 9, 5 & -15° &90°\\\hline
    Finger Base Lateral (FF4,MF4,RF4,LF4) & 18, 14, 10, 6 & -20° &20°\\\hline
    Little Finger Rotation(LF5) & 19 & 0° &45°\\\hline
    Thumb Distal (TH1) & 20 & -15° &90°\\\hline
    Thumb Middle (TH2)&	21 & -30° &30°\\\hline
    Thumb Base Abduction (TH3)& 22 & -12° &12°\\\hline
    Thumb Base Lateral (TH4)& 23 & 0° &70°\\\hline
    Thumb Base Rotation (TH5)& 24 & -60° &60°\\\hline
    Hand Wrist Abduction (WR1) & 1 & -40° &28°\\\hline
    Hand Wrist Lateral (WR2)& 2 & -28° &8°\\

    \bottomrule %添加表格底部粗线
    \end{tabular}
    \label{joint_limit}
\end{table}

Stiffness, damping, friction, and armature are also important physical parameters in robotics. For each Shadow Hand's joint, we show our DoF properties in Table \ref{dof_properties}. This part can be adjusted in the Isaac Gym simulator. 

\begin{table}[htbp]
    \centering
    \caption{DoF properties of Shadow Hand.}
    \begin{tabular}{ccccc}
    \toprule  %添加表格头部粗线
    Joints& Stifness & Damping & Friction & Armature\\\hline
    WR1 & 100 & 4.78 & 0 & 0\\\hline
    WR2 & 100 & 2.17 & 0 & 0\\\hline
    FF2 & 100 & 3.4e+38 & 0 & 0\\\hline
    FF3 & 100 & 0.9 & 0 & 0\\\hline
    FF4 & 100 & 0.725 & 0 & 0\\\hline
    MF2 & 100 & 3.4e+38 & 0 & 0\\\hline
    MF3 & 100 & 0.9 & 0 & 0\\\hline
    MF4 & 100 & 0.725 & 0 & 0\\\hline
    RF2 & 100 & 3.4e+38 & 0 & 0\\\hline
    RF3 & 100 & 0.9 & 0 & 0\\\hline
    RF4 & 100 & 0.725 & 0 & 0\\\hline
    LF2 & 100 & 3.4e+38 & 0 & 0\\\hline
    LF3 & 100 & 0.9 & 0 & 0\\\hline
    LF4 & 100 & 0.725 & 0 & 0\\\hline
    TH2 & 100 & 3.4e+38 & 0 & 0\\\hline
    TH3 & 100 & 0.99 & 0 & 0\\\hline
    TH4 & 100 & 0.99 & 0 & 0\\\hline
    TH5 & 100 & 0.81 & 0 & 0\\

    \bottomrule %添加表格底部粗线
    \end{tabular}
    \label{dof_properties}
\end{table}

\subsection{Detailed components of tasks }\label{Appendix:A2}

In this section, we detailed the components of tasks in Bi-DexHands. We refer to some designs of existing dexterous hand environments, integrate their advantages, and expand some new environments and unique features for single/multi-agent reinforcement learning. Our environments focus on the application of RL algorithms to dexterous hand control, which is challenging in traditional control algorithms. The difficulty of our environment is not only reflected in the challenging task content but also reflected in the high-dimensional continuous space control. The state space dimension of each environment is up to 400 dimensions in total, and the action space dimension is up to 40 dimensions. A multi-agent feature of our environment is that we use five fingers and palms of each hand as a minimum agent unit. It is mean that you can use each finger and palm as an agent, or combine any number of them as an agent by yourself. All environments are goal-based, and each epoch will randomly reset the object's starting pose and target pose to improve generalization. All objects type can be selected in the config, the basis is egg, block, and pen. We also provide objects type in the YCB dataset as an extension, you can customize the object type they want to use.

The objects in the YCB dataset are used for our object-catching tasks. Because our object-catching environment is only related to the pose of the object, we can arbitrarily replace objects of suitable size in the YCB dataset. Other tasks use items from the Sapien dataset, and can also use other objects from the same category in Sapien dataset. However, because it is related to the shape of the object, some additional operations are required. We have added examples to Github to show how to use objects from YCB and Sapien datasets, see \href{https://github.com/PKU-MARL/DexterousHands/blob/main/docs/customize\%20the\%20environment.md}{here}.

\begin{figure}[htbp]
 \centering  %居中
 \includegraphics[width=\hsize]{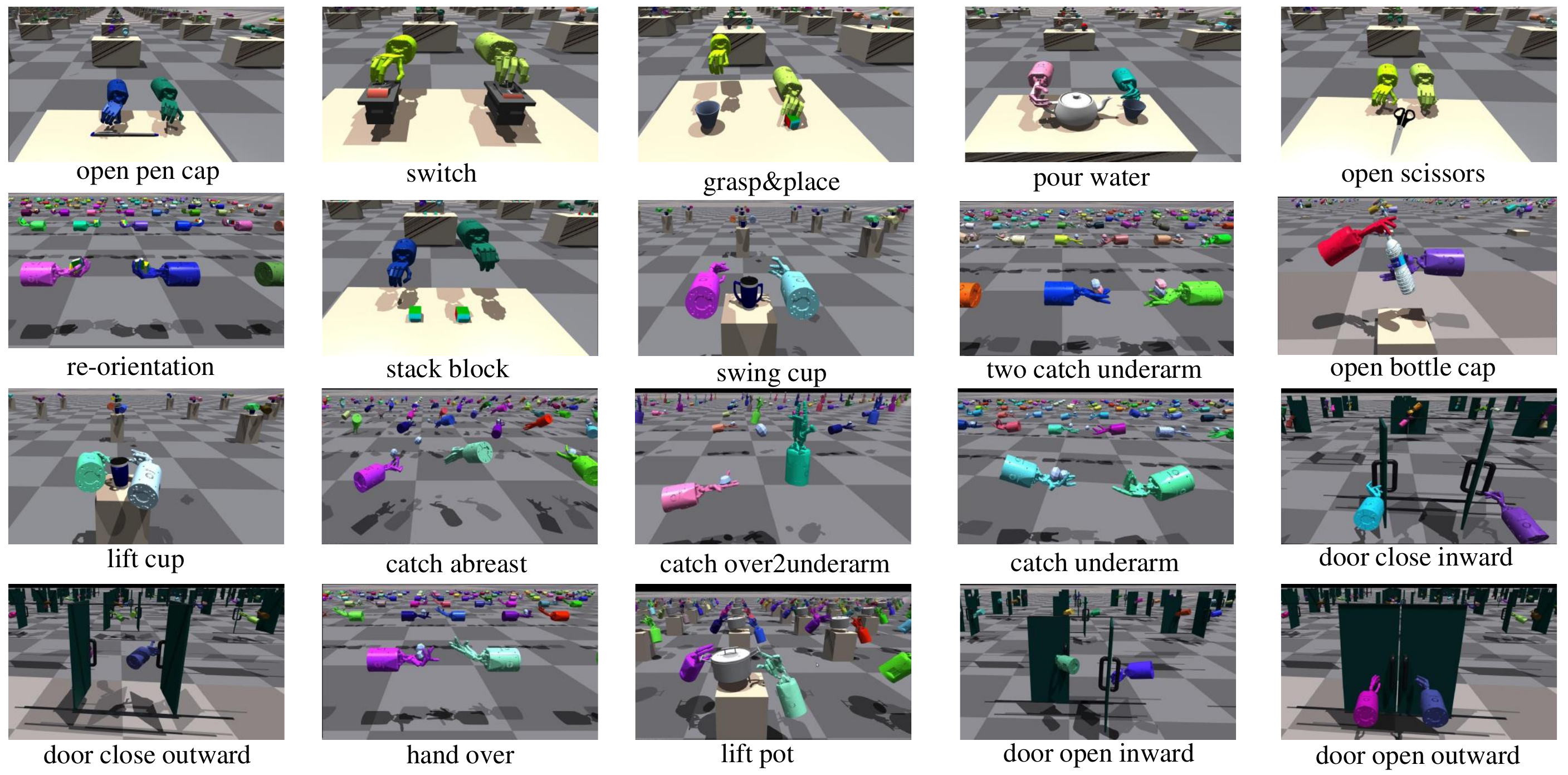} 
 \caption{An overview of all tasks.} 
 \label{fig3}
\end{figure}

An overview of our tasks is shown in Fig.\ref{fig3}. Next, we will introduce the basic description, action space, observation space, and reward function of each task. We only use the Shadow Hand and object state values as observation at present, but we also provide an interface for using point cloud as observation in our Github repository for researchers to study in the future. The observation of all tasks is composed of three parts: the state values of the left and right hands, and the information of objects and target. The state values of the left and right hands were the same for each task, including hand joint and finger positions, velocity, and force information. The state values of the object and goal are different for each task, which we will describe in the following. Table.\ref{observation_dual hands} gives the specific information of the left-hand and right-hand state values. Note that the observation is slightly different in the HandOver task due to the fixed base.

Designing a reward function is very important for an RL task. I would like to introduce the method of our reward design in detail. In general, our reward design is goal-based and follows the same set of logic. For object-catching tasks, our reward is simply related to the difference between the pose of the object and the target. The closer the object is to the target, the greater the reward. For other tasks that require the hand to hold the object, our reward generally consists of three parts: the distance from the left hand to the target point on the object that the left-hand needs to operate, the distance from the right hand to the target point on the object that the right-hand needs to operate, and the distance from the object to the target. 
The principle of our design is to conform to human intuition based on completing the task and to make the reward function structure as unified as possible. This unified reward function structure is also one of the requirements of Meta RL and Multi-task RL environment design. However, because the scenarios of each task are different, the hyperparameters of the reward function will inevitably be different. We have tried our best to avoid manual reward shaping for each task provided that RL can be successfully trained.

\begin{table}[htbp]
    \centering
    \caption{Observation space of dual Shadow Hands.}
    \begin{tabular}{c|c}
    \toprule  %添加表格头部粗线
    Index&Description\\\hline
    0 - 23&	right Shadow Hand dof position\\\hline
    24 - 47&	right Shadow Hand dof velocity\\\hline
    48 - 71	& right Shadow Hand dof force\\\hline
    72 - 136&	right Shadow Hand fingertip pose, linear velocity, angle velocity (5 x 13)\\\hline
    137 - 166&	right Shadow Hand fingertip force, torque (5 x 6)\\\hline
    167 - 169&	right Shadow Hand base position\\\hline
    170 - 172&	right Shadow Hand base rotation\\\hline
    173 - 198&	right Shadow Hand actions\\\hline
    
    199 - 222&	left Shadow Hand dof position\\\hline
    223 - 246&	left Shadow Hand dof velocity\\\hline
    247 - 270	& left Shadow Hand dof force\\\hline
    271 - 335&	left Shadow Hand fingertip pose, linear velocity, angle velocity (5 x 13)\\\hline
    336 - 365&	left Shadow Hand fingertip force, torque (5 x 6)\\\hline
    366 - 368&	left Shadow Hand base position\\\hline
    369 - 371&	left Shadow Hand base rotation\\\hline
    372 - 397&	left Shadow Hand actions\\
    \bottomrule %添加表格底部粗线
    \end{tabular}
    \label{observation_dual hands}
\end{table}
 
Under the multi-agent setting, the partial observation of each agent depends on the observation of the hand it belongs to. For example, if the left distal finger, left thumb, and right distal finger are one agent respectively, the observation of the left distal finger and left thumb are the observation of the entire left hand in the Table \ref{observation_dual hands} plus the object and target information. The obs of the right distal finger is the observation of the entire right hand in the Table \ref{observation_dual hands} plus object and target information. The action of each agent depends on the multi-agent setting (\textit{i.e.}, fingers, hands,...), and the output by each agent is the joint degree of itself. Bi-DexHands is a fully-cooperative game where all agents have the same reward. Therefore, the setting of multi-agent can be completely inferred from the setting of single-agent.

\subsubsection{Hand Over}

This environment consists of two Shadow Hands with palms facing up, opposite each other, and an object that needs to be passed. In the beginning, the object will fall randomly in the area of the Shadow Hand on the right side. Then the hand holds the object and passes the object to the other hand. Note that the base of the hand is fixed. More importantly, the hand which holds the object initially can not directly touch the target, nor can it directly roll the object to the other hand, so the object must be thrown up and stays in the air in the process. There are 398-dimensional observations and 40-dimensional actions in the task. Additionally, the reward function is related to the pose error between the object and the target. When the pose error gets smaller, the reward increases dramatically. Specifically, the observation space of each agent is detailed in the following Table \ref{observation_handover}, and the action space is shown in Table \ref{action_handover}.

\paragraph{Observations}
The 398-dimensional observational space for Hand Over task is shown in Table \ref{observation_handover}. It should be noted that since the base of the dual hands in this task is fixed, the observation of the dual hands is compared to the Table \ref{observation_dual hands} of reduced 24 dimensions.

\begin{table}[htbp]
    \centering
    \caption{Observation space of Hand Over.}
    \begin{tabular}{c|c}
    \toprule  %添加表格头部粗线
    Index&Description\\\hline
    0 - 373&	dual hands observation shown in Table \ref{observation_dual hands}\\\hline
    374 - 380&	object pose\\\hline
    381 - 383&	object linear velocity\\\hline
    384 - 386&	object angle velocity\\\hline
    387 - 393&	goal pose\\\hline
    394 - 397&	goal rot - object rot\\
    \bottomrule %添加表格底部粗线
    \end{tabular}
    \label{observation_handover}
\end{table}

\paragraph{Actions}
The 40-dimensional action space for one hand in Hand Over task is shown in Table \ref{action_handover}.

\begin{table}[htbp]
    \centering
    \caption{Action space of Hand Over.}
    \begin{tabular}{c|c}
    \toprule  %添加表格头部粗线
    Index&Description\\\hline
    0 - 19&	right Shadow Hand actuated joint \\\hline
    20 - 39&	left Shadow Hand actuated joint\\
    \bottomrule %添加表格底部粗线
    \end{tabular}
    \label{action_handover}
\end{table}

\paragraph{Rewards}
Denote the object and goal position as $x_o$ and $x_g$ respectively. Then, the translational position difference between the object and the goal $d_t$ is given by $d_t=\Vert x_o-x_g \Vert_2$. Denote the angular position difference between the object and the goal as $d_a$, then the rotational difference $d_r$ is given by $d_r = 2\arcsin{clamp(\Vert d_a \Vert_2,max = 1.0)}$. Finally, the rewards are given by the following specific formula:
\begin{equation}
    r = exp[-0.2(\alpha d_t + d_r)]
\end{equation}
where $\alpha$ is a constant balancing translational and rotational rewards.

\subsubsection{Catch Underarm}

In this problem, two Shadow Hands with palms facing upwards are controlled to pass an object from one palm to the other. What makes it more difficult than the Handover problem is that the hands' translation and rotation degrees of freedom are no longer frozen but are added into the action space.

\paragraph{Observations}
The 422-dimensional observational space as shown in Table \ref{catch_underarm_obs}.

\begin{table}[htbp]
    \centering
    \caption{Observation space of Catch Underarm.}
    \begin{tabular}{c|c}
    \toprule  %添加表格头部粗线
    Index&Description\\\hline
    0 - 397&	dual hands observation shown in Table \ref{observation_dual hands}\\\hline
    398 - 404&	object pose\\\hline
    405 - 407&	object linear velocity\\\hline
    408 - 410&	object angle velocity\\\hline
    411 - 417&	goal pose\\\hline
    418 - 421&	goal rot - object rot\\

    \bottomrule %添加表格底部粗线
    \end{tabular}
    \label{catch_underarm_obs}
\end{table}

\paragraph{Actions}
The 52-dimensional action space as shown in Table \ref{catch_underarm_action}.

\begin{table}[htbp]
    \centering
    \caption{Action space of Catch Underarm.}
    \begin{tabular}{c|c}
    \toprule  %添加表格头部粗线
    Index&Description\\\hline
    0 - 19&	right Shadow Hand actuated joint\\\hline
    20 - 22&	right Shadow Hand base translation\\\hline
    23 - 25&	right Shadow Hand base rotation\\\hline
    26 - 45&	left Shadow Hand actuated joint\\\hline
    46 - 48&	left Shadow Hand base translation\\\hline
    49 - 51&	left Shadow Hand base rotation\\
    
    \bottomrule %添加表格底部粗线
    \end{tabular}
    \label{catch_underarm_action}
\end{table}

\paragraph{Rewards}
Denote the object and goal position as $x_o$ and $x_g$ respectively. Then, the translational position difference between the object and the goal $d_t$ is given by $d_t=\Vert x_o-x_g \Vert_2$. Denote the angular position difference between the object and the goal as $d_a$, then the rotational difference $d_r$ is given by $d_r = 2\arcsin{clamp(\Vert d_a \Vert_2,max = 1.0)}$. Finally, the rewards are given by the following specific formula:
\begin{equation}
    r = exp[-0.2(\alpha d_t + d_r)]
\end{equation}
where $\alpha$ is a constant balancing translational and rotational rewards.

\subsubsection{Catch Over2Underarm}

This environment is like made up of half Hand Over and Catch Underarm, the object needs to be thrown from the vertical hand to the palm-up hand. 

\paragraph{Observations}
The 422-dimensional observational space as shown in Table \ref{catch_over2underarm_obs}.

\begin{table}[htbp]
    \centering
    \caption{Observation space of Catch Over2Underarm.}
    \begin{tabular}{c|c}
    \toprule  %添加表格头部粗线
    Index&Description\\\hline
    0 - 397&	dual hands observation shown in Table \ref{observation_dual hands}\\\hline
    398 - 404&	object pose\\\hline
    405 - 407&	object linear velocity\\\hline
    408 - 410&	object angle velocity\\\hline
    411 - 417&	goal pose\\\hline
    418 - 421&	goal rot - object rot\\
    \bottomrule %添加表格底部粗线
    \end{tabular}
    \label{catch_over2underarm_obs}
\end{table}

\paragraph{Actions}
The 52-dimensional action space as shown in Table \ref{catch_over2underarm_action}.

\begin{table}[htbp]
    \centering
    \caption{Action space of Catch Over2Underarm.}
    \begin{tabular}{c|c}
    \toprule  %添加表格头部粗线
    Index&Description\\\hline
    0 - 19&	right Shadow Hand actuated joint\\\hline
    20 - 22&	right Shadow Hand base translation\\\hline
    23 - 25&	right Shadow Hand base rotation\\\hline
    26 - 45&	left Shadow Hand actuated joint\\\hline
    46 - 48&	left Shadow Hand base translation\\\hline
    49 - 51&	left Shadow Hand base rotation\\
    \bottomrule %添加表格底部粗线
    \end{tabular}
    \label{catch_over2underarm_action}
\end{table}

\paragraph{Rewards}
Denote the object and goal position as $x_o$ and $x_g$ respectively. Then, the translational position difference between the object and the goal $d_t$ is given by $d_t=\Vert x_o-x_g \Vert_2$. Denote the angular position difference between the object and the goal as $d_a$, then the rotational difference $d_r$ is given by $d_r = 2\arcsin{clamp(\Vert d_a \Vert_2,max = 1.0)}$. Finally, the rewards are given by the following specific formula:
\begin{equation}
    r = exp[-0.2(\alpha d_t + d_r)]
\end{equation}
where $\alpha$ is a constant balancing translational and rotational rewards.

\subsubsection{Two Catch Underarm}

This environment is similar to Catch Underarm, but with an object in each hand and the corresponding goal on the other hand. Therefore, the environment requires two objects to be thrown into the other hand at the same time, which requires a higher manipulation technique than the environment of a single object.

\paragraph{Observations}
The 446-dimensional observational space as shown in Table \ref{two_catch_underarm_obs}.

\begin{table}[htbp]
    \centering
    \caption{Observation space of Two Catch Underarm.}
    \begin{tabular}{c|c}
    \toprule  %添加表格头部粗线
    Index&Description\\\hline
    0 - 397&	dual hands observation shown in Table \ref{observation_dual hands}\\\hline
    398 - 404&	object1 pose\\\hline
    405 - 407&	object1 linear velocity\\\hline
    408 - 410&	object1 angle velocity\\\hline
    411 - 417&	goal1 pose\\\hline
    418 - 421&	goal1 rot - object rot\\
    422 - 428&	object2 pose\\\hline
    429 - 431&	object2 linear velocity\\\hline
    432 - 434&	object2 angle velocity\\\hline
    435 - 441&	goal2 pose\\\hline
    442 - 445&	goal2 rot - object2 rot\\

    \bottomrule %添加表格底部粗线
    \end{tabular}
    \label{two_catch_underarm_obs}
\end{table}

\paragraph{Actions}
The 52-dimensional action space as shown in Table \ref{two_catch_underarm_action}.

\begin{table}[htbp]
    \centering
    \caption{Action space of Two Catch Underarm.}
    \begin{tabular}{c|c}
    \toprule  %添加表格头部粗线
    Index&Description\\\hline
    0 - 19&	right Shadow Hand actuated joint\\\hline
    20 - 22&	right Shadow Hand base translation\\\hline
    23 - 25&	right Shadow Hand base rotation\\\hline
    26 - 45&	left Shadow Hand actuated joint\\\hline
    46 - 48&	left Shadow Hand base translation\\\hline
    49 - 51&	left Shadow Hand base rotation\\
    \bottomrule %添加表格底部粗线
    \end{tabular}
    \label{two_catch_underarm_action}
\end{table}

\paragraph{Rewards}
For the reward part, we use subscripts 1,2 to distinguish the 2 objects.

Denote the object and goal position as $x_{o_1}$,$x_{o_2}$ and $x_{g_1}$,$x_{g_2}$ respectively. Then, the translational position difference between the object and the goal $d_{t_1}$,$d_{t_2}$ is given by $d_{t_i}=\Vert x_{o_i}-x_{g_i} \Vert_2$, where $i = 1,2$. Denote the angular position difference between the object and the goal as $d_{a_1}$,$d_{a_2}$, then the rotational difference $d_{r_1}$,$d_{r_2}$ is given by $d_{r_i} = 2\arcsin{clamp(\Vert d_{a_i} \Vert_2,max = 1.0)}$. Finally, the rewards are given by the following specific formula:
\begin{equation}
    r = exp[-0.2(\alpha d_{t_1} + d_{r_1})]+exp[-0.2(\alpha d_{t_2} + d_{r_2})]
\end{equation}
where $\alpha$ is a constant balancing translational and rotational rewards.

\subsubsection{Catch Abreast}

This environment consists of two Shadow Hands placed side by side in the same direction and an object that needs to be passed. Compared with the previous environment which is more like passing objects between the hands of two people, this environment is designed to simulate the two hands of the same person passing objects, so different catch techniques are also required and require more hand translation and rotation techniques.

\paragraph{Observations}
The 422-dimensional observation space as shown in Table \ref{catch_abreast_obs}.

\begin{table}[htbp]
    \centering
    \caption{Observation space of Catch Abreast.}
    \begin{tabular}{c|c}
    \toprule  %添加表格头部粗线
    Index&Description\\\hline
    0 - 397&	dual hands observation shown in Table \ref{observation_dual hands}\\\hline
    398 - 404&	object pose\\\hline
    405 - 407&	object linear velocity\\\hline
    408 - 410&	object angle velocity\\\hline
    411 - 417&	goal pose\\\hline
    418 - 421&	goal rot - object rot\\
    \bottomrule %添加表格底部粗线
    \end{tabular}
    \label{catch_abreast_obs}
\end{table}

\paragraph{Actions}
The 52-dimensional action space as shown in Table \ref{catch_abreast_action}.

\begin{table}[htbp]
    \centering
    \caption{Action space of Catch Abreast.}
    \begin{tabular}{c|c}
    \toprule  %添加表格头部粗线
    Index&Description\\\hline
    0 - 19&	right Shadow Hand actuated joint\\\hline
    20 - 22&	right Shadow Hand base translation\\\hline
    23 - 25&	right Shadow Hand base rotation\\\hline
    26 - 45&	left Shadow Hand actuated joint\\\hline
    46 - 48&	left Shadow Hand base translation\\\hline
    49 - 51&	left Shadow Hand base rotation\\
    \bottomrule %添加表格底部粗线
    \end{tabular}
    \label{catch_abreast_action}
\end{table}

\paragraph{Rewards}
Denote the object and goal position as $x_o$ and $x_g$ respectively. Then, the translational position difference between the object and the goal $d_t$ is given by $d_t=\Vert x_o-x_g \Vert_2$. Denote the angular position difference between the object and the goal as $d_a$, then the rotational difference $d_r$ is given by $d_r = 2\arcsin{clamp(\Vert d_a \Vert_2,max = 1.0)}$. Finally, the rewards are given by the following specific formula:
\begin{equation}
    r = exp[-0.2(\alpha d_t + d_r)]
\end{equation}
where $\alpha$ is a constant balancing translational and rotational rewards.

\subsubsection{Lift Underarm}

This environment requires grasping the pot handle with two hands and lifting the pot to the designated position. This environment is designed to simulate the scene of lift in daily life and is a  practical skill.

\paragraph{Observations}
The 428-dimensional observation space as shown in Table \ref{lift_underarm_obs}. 

\begin{table}[htbp]
    \centering
    \caption{Observation space of Lift Underarm.}
    \begin{tabular}{c|c}
    \toprule  %添加表格头部粗线
    Index&Description\\\hline
    0 - 397&	dual hands observation shown in Table \ref{observation_dual hands}\\\hline
    398 - 404&	object pose\\\hline
    405 - 407&	object linear velocity\\\hline
    408 - 410&	object angle velocity\\\hline
    411 - 417&	goal pose\\\hline
    418 - 421&	goal rot - object rot\\
    422 - 424&	object right handle position\\\hline
    425 - 427&	object left handle position\\

    \bottomrule %添加表格底部粗线
    \end{tabular}
    \label{lift_underarm_obs}
\end{table}

\paragraph{Actions}
The 40-dimensional action space as shown in Table \ref{lift_underarm_action}. 

\begin{table}[htbp]
    \centering
    \caption{Action space of Lift Underarm.}
    \begin{tabular}{c|c}
    \toprule  %添加表格头部粗线
    Index&Description\\\hline
    0 - 19&	right Shadow Hand actuated joint\\\hline
    20 - 22&	right Shadow Hand base translation\\\hline
    23 - 25&	right Shadow Hand base rotation\\\hline
    26 - 45&	left Shadow Hand actuated joint\\\hline
    46 - 48&	left Shadow Hand base translation\\\hline
    49 - 51&	left Shadow Hand base rotation\\
    \bottomrule %添加表格底部粗线
    \end{tabular}
    \label{lift_underarm_action}
\end{table}

\paragraph{Rewards}
The reward consists of three parts: the distance from the left hand to the left handle, the distance from the right hand to the right handle, and the distance from the object to the target point.
The position difference between the object to the target point $d_{target}$ is given by $d_{target}=\Vert x_{obj}-x_{goal} \Vert_2$. The position difference between the left hand to the left handle $d_{left}$ is given by $d_{left}=\Vert x_{lhand}-x_{lhandle} \Vert_2$.The position difference between the right hand to the right handle $d_{right}$ is given by $d_{right}=\Vert x_{rhand}-x_{rhandle} \Vert_2$. The reward is given by this specific formula:
\begin{equation}
    r = 0.2 - d_{left} - d_{right} + 3 * (0.985 - d_{target})
\end{equation}

\subsubsection{Door Open Outward/Door Close Inward}

These two environments require a closed/opened door to be opened/closed and the door can only be pushed outward or initially open inward. Both these two environments only need to do the push behavior, so it is relatively simple.

\paragraph{Observations}
The 428-dimensional observation space as shown in Table \ref{door_push_obs}. 

\begin{table}[htbp]
    \centering
    \caption{observation space of Door Open Outward/Door Close Inward.}
    \begin{tabular}{c|c}
    \toprule  %添加表格头部粗线
    Index&Description\\\hline
    0 - 397&	dual hands observation shown in Table \ref{observation_dual hands}\\\hline
    398 - 404&	object pose\\\hline
    405 - 407&	object linear velocity\\\hline
    408 - 410&	object angle velocity\\\hline
    411 - 417&	goal pose\\\hline
    418 - 421&	goal rot - object rot\\
    422 - 424&	door right handle position\\\hline
    425 - 427&	door left handle position\\

    \bottomrule %添加表格底部粗线
    \end{tabular}
    \label{door_push_obs}
\end{table}

\paragraph{Actions}
The 52-dimensional action space as shown in Table \ref{door_push_action}. 

\begin{table}[htbp]
    \centering
    \caption{Action space of Door Open Outward/Door Close Inward.}
    \begin{tabular}{c|c}
    \toprule  %添加表格头部粗线
    Index&Description\\\hline
    0 - 19&	right Shadow Hand actuated joint\\\hline
    20 - 22&	right Shadow Hand base translation\\\hline
    23 - 25&	right Shadow Hand base rotation\\\hline
    26 - 45&	left Shadow Hand actuated joint\\\hline
    46 - 48&	left Shadow Hand base translation\\\hline
    49 - 51&	left Shadow Hand base rotation\\
    \bottomrule %添加表格底部粗线
    \end{tabular}
    \label{door_push_action}
\end{table}

\paragraph{Rewards}
The reward consists of three parts: the distance from the left hand to the left handle, the distance from the right hand to the right handle, and the distance between the two handles.
The distance between the two handles $d_{target}$ is given by $d_{target}=\Vert x_{lhandle}-x_{rhandle} \Vert_2$. The position difference between the left hand to the left handle $d_{left}$ is given by $d_{left}=\Vert x_{lhand}-x_{lhandle} \Vert_2$.The position difference between the right hand to the right handle $d_{right}$ is given by $d_{right}=\Vert x_{rhand}-x_{rhandle} \Vert_2$. For DoorOpenOutward, the reward is given by this specific formula:
\begin{equation}
    r = 0.2 - d_{left} - d_{right} + 2 * d_{target}
\end{equation}
For DoorCloseInward, the reward is given by this specific formula:
\begin{equation}
    r = 0.2 - d_{left} - d_{right} + 2 * (1 - d_{target})
\end{equation}

\subsubsection{Door Open Inward/Door Close Outward}

These two environments also require a closed/opened door to be opened/closed and the door can only be pushed inward or initially open outward, but because they can't complete the task by simply pushing, which need to catch the handle by hand and then open or close it, so it is relatively difficult.

\paragraph{Observations}
The 428-dimensional observation space as shown in Table \ref{door_catch_obs}. 

\begin{table}[htbp]
    \centering
    \caption{Observation space of Door Open Inward/Door Close Outward.}
    \begin{tabular}{c|c}
    \toprule  %添加表格头部粗线
    Index&Description\\\hline
    0 - 397&	dual hands observation shown in Table \ref{observation_dual hands}\\\hline
    398 - 404&	object pose\\\hline
    405 - 407&	object linear velocity\\\hline
    408 - 410&	object angle velocity\\\hline
    411 - 417&	goal pose\\\hline
    418 - 421&	goal rot - object rot\\
    422 - 424&	door right handle position\\\hline
    425 - 427&	door left handle position\\

    \bottomrule %添加表格底部粗线
    \end{tabular}
    \label{door_catch_obs}
\end{table}

\paragraph{Actions}
The 52-dimensional action space as shown in Table \ref{door_catch_action}. 

\begin{table}[htbp]
    \centering
    \caption{Action space of Door Open Inward/Door Close Outward.}
    \begin{tabular}{c|c}
    \toprule  %添加表格头部粗线
    Index&Description\\\hline
    0 - 19&	right Shadow Hand actuated joint\\\hline
    20 - 22&	right Shadow Hand base translation\\\hline
    23 - 25&	right Shadow Hand base rotation\\\hline
    26 - 45&	left Shadow Hand actuated joint\\\hline
    46 - 48&	left Shadow Hand base translation\\\hline
    49 - 51&	left Shadow Hand base rotation\\
    \bottomrule %添加表格底部粗线
    \end{tabular}
    \label{door_catch_action}
\end{table}

\paragraph{Rewards}
The reward consists of three parts: the distance from the left hand to the left handle, the distance from the right hand to the right handle, and the distance between the two handles.
The distance between the two handles $d_{target}$ is given by $d_{target}=\Vert x_{lhandle}-x_{rhandle} \Vert_2$. The position difference between the left hand to the left handle $d_{left}$ is given by $d_{left}=\Vert x_{lhand}-x_{lhandle} \Vert_2$.The position difference between the right hand to the right handle $d_{right}$ is given by $d_{right}=\Vert x_{rhand}-x_{rhandle} \Vert_2$. For DoorOpenInward, the reward is given by this specific formula:
\begin{equation}
    r = 0.2 - d_{left} - d_{right} + 2 * d_{target}
\end{equation}
For DoorCloseOutward, the reward is given by this specific formula:
\begin{equation}
    r = 0.2 - d_{left} - d_{right} + 2 * (1 - d_{target})
\end{equation}

\subsubsection{Bottle Cap}

This environment involves two hands and a bottle, we need to hold the bottle with one hand and open the bottle cap with the other hand. This skill requires the cooperation of two hands to ensure that the cap does not fall.

\paragraph{Observations}
The 414-dimensional observation space as shown in Table \ref{bottle_cap_obs}. 

\begin{table}[htbp]
    \centering
    \caption{Observation space of Bottle Cap.}
    \begin{tabular}{c|c}
    \toprule  %添加表格头部粗线
    0 - 397&	dual hands observation shown in Table \ref{observation_dual hands}\\\hline
    398 - 404&	bottle pose\\\hline
    405 - 407&	bottle linear velocity\\\hline
    408 - 410&	bottle angle velocity\\\hline
    411 - 413&	bottle cap position\\

    \bottomrule %添加表格底部粗线
    \end{tabular}
    \label{bottle_cap_obs}
\end{table}

\paragraph{Actions}
The 52-dimensional action space as shown in Table \ref{bottle_cap_action}. 

\begin{table}[htbp]
    \centering
    \caption{Action space of Bottle Cap.}
    \begin{tabular}{c|c}
    \toprule  %添加表格头部粗线
    Index&Description\\\hline
    0 - 19&	right Shadow Hand actuated joint\\\hline
    20 - 22&	right Shadow Hand base translation\\\hline
    23 - 25&	right Shadow Hand base rotation\\\hline
    26 - 45&	left Shadow Hand actuated joint\\\hline
    46 - 48&	left Shadow Hand base translation\\\hline
    49 - 51&	left Shadow Hand base rotation\\
    \bottomrule %添加表格底部粗线
    \end{tabular}
    \label{bottle_cap_action}
\end{table}

\paragraph{Rewards}
The reward also consists of three parts: the distance from the left hand to the bottle cap, the distance from the right hand to the bottle, and the distance between the bottle and bottle cap.
The distance between the bottle and bottle cap $d_{target}$ is given by $d_{target}=\Vert x_{bottle}-x_{bottlecap} \Vert_2$. the distance from the left hand to the bottle cap $d_{left}$ is given by $d_{left}=\Vert x_{lhand}-x_{bottlecap} \Vert_2$. the distance from the right hand to the bottle $d_{right}$ is given by $d_{right}=\Vert x_{rhand}-x_{bottle} \Vert_2$. The reward is given by this specific formula:
\begin{equation}
    r = 0.2 - d_{left} - d_{right} + 30 * d_{target}
\end{equation}

\subsubsection{Push Block}

This environment involves two hands and two blocks, we need to use both hands to reach and push the block to the desired goal separately. This is a relatively simple task.

\paragraph{Observations}
The 417-dimensional observation space as shown in Table \ref{push_block_obs}. 

\begin{table}[htbp]
    \centering
    \caption{Observation space of Push Block.}
    \begin{tabular}{c|c}
    \toprule  %添加表格头部粗线
    Index&Description\\\hline
    0 - 397&	dual hands observation shown in Table \ref{observation_dual hands}\\\hline
    398 - 404&	block1 pose\\\hline
    405 - 407&	block1 linear velocity\\\hline
    408 - 410&	block1 angle velocity\\\hline
    411 - 413&	block1 position\\\hline
    414 - 416&	block2 position\\
    \bottomrule %添加表格底部粗线
    \end{tabular}
    \label{push_block_obs}
\end{table}

\paragraph{Actions}
The 52-dimensional action space as shown in Table \ref{push_block_action}. 

\begin{table}[htbp]
    \centering
    \caption{Action space of Push Block.}
    \begin{tabular}{c|c}
    \toprule  %添加表格头部粗线
    Index&Description\\\hline
    0 - 19&	right Shadow Hand actuated joint\\\hline
    20 - 22&	right Shadow Hand base translation\\\hline
    23 - 25&	right Shadow Hand base rotation\\\hline
    26 - 45&	left Shadow Hand actuated joint\\\hline
    46 - 48&	left Shadow Hand base translation\\\hline
    49 - 51&	left Shadow Hand base rotation\\
    \bottomrule %添加表格底部粗线
    \end{tabular}
    \label{push_block_action}
\end{table}

\paragraph{Rewards}
The reward consists of three parts: the distance from the left hand to block1, the distance from the right hand to block2, and the distance between the block and desired goal.
The distance between the block and desired goal $d_{target}$ is given by $d_{target}=\Vert x_{block1}-x_{block1_goal} \Vert_2 + \Vert x_{block2}-x_{block2_goal} \Vert_2$. the distance from the left hand to the block1 $d_{left}$ is given by $d_{left}=\Vert x_{lhand}-x_{block1} \Vert_2$. the distance from the right hand to the block2 $d_{right}$ is given by $d_{right}=\Vert x_{rhand}-x_{block2} \Vert_2$. The reward is given by this specific formula:
\begin{equation}
    r = 2 - d_{left} - d_{right} + 5 * (0.8 - d_{target})
\end{equation}

\subsubsection{Swing Cup}

This environment involves two hands and a dual handle cup, we need to use two hands to hold and swing the cup together.

\paragraph{Observations}
The 428-dimensional observation space as shown in Table \ref{swing_cup_obs}. 

\begin{table}[htbp]
    \centering
    \caption{Observation space of Swing Cup.}
    \begin{tabular}{c|c}
    \toprule  %添加表格头部粗线
    Index&Description\\\hline
    0 - 397&	dual hands observation shown in Table \ref{observation_dual hands}\\\hline
    398 - 404&	cup pose\\\hline
    405 - 407&	cup linear velocity\\\hline
    408 - 410&	cup angle velocity\\\hline
    411 - 417&	goal pose\\\hline
    418 - 421&	goal rot - object rot\\
    422 - 424&	cup right handle position\\\hline
    425 - 427&	cup left handle position\\

    \bottomrule %添加表格底部粗线
    \end{tabular}
    \label{swing_cup_obs}
\end{table}

\paragraph{Actions}
The 52-dimensional action space as shown in Table \ref{swing_cup_action}. 

\begin{table}[htbp]
    \centering
    \caption{Action space of Swing Cup.}
    \begin{tabular}{c|c}
    \toprule  %添加表格头部粗线
    Index&Description\\\hline
    0 - 19&	right Shadow Hand actuated joint\\\hline
    20 - 22&	right Shadow Hand base translation\\\hline
    23 - 25&	right Shadow Hand base rotation\\\hline
    26 - 45&	left Shadow Hand actuated joint\\\hline
    46 - 48&	left Shadow Hand base translation\\\hline
    49 - 51&	left Shadow Hand base rotation\\
    \bottomrule %添加表格底部粗线
    \end{tabular}
    \label{swing_cup_action}
\end{table}

\paragraph{Rewards}
The reward consists of three parts: the distance from the left hand to the cup's left handle, the distance from the right hand to the cup's right handle, and the rotating distance between the cup and desired goal.
The rotate distance between the cup and desired goal $d_{target}$ is given by $d_{target}=2 * \arcsin{q_{cup} * q_{target}}$. the distance from the left hand to the cup left handle $d_{left}$ is given by $d_{left}=\Vert x_{lhand}-x_{lhandle} \Vert_2$. the distance from the right hand to the cup right handle $d_{right}$ is given by $d_{right}=\Vert x_{rhand}-x_{rhandle} \Vert_2$. The reward is given by this specific formula:
\begin{equation}
    r = - d_{left} - d_{right} + 1 / (abs(d_{target}) + 0.1) * 5 - 1
\end{equation}

\subsubsection{Open Scissors}

This environment involves two hands and scissors, we need to use two hands to open the scissors.

\paragraph{Observations}
The 428-dimensional observation space as shown in Table \ref{scissors_obs}. 

\begin{table}[htbp]
    \centering
    \caption{Observation space of Open Scissors.}
    \begin{tabular}{c|c}
    \toprule  %添加表格头部粗线
    Index&Description\\\hline
    0 - 397&	dual hands observation shown in Table \ref{observation_dual hands}\\\hline
    398 - 404&	scissors pose\\\hline
    405 - 407&	scissors linear velocity\\\hline
    408 - 410&	scissors angle velocity\\\hline
    411 - 417&	goal pose\\\hline
    418 - 421&	goal rot - object rot\\
    422 - 424&	scissors right handle position\\\hline
    425 - 427&	scissors left handle position\\

    \bottomrule %添加表格底部粗线
    \end{tabular}
    \label{scissors_obs}
\end{table}

\paragraph{Actions}
The 52-dimensional action space as shown in Table \ref{scissors_action}. 

\begin{table}[htbp]
    \centering
    \caption{Action space of Open Scissors.}
    \begin{tabular}{c|c}
    \toprule  %添加表格头部粗线
    Index&Description\\\hline
    0 - 19&	right Shadow Hand actuated joint\\\hline
    20 - 22&	right Shadow Hand base translation\\\hline
    23 - 25&	right Shadow Hand base rotation\\\hline
    26 - 45&	left Shadow Hand actuated joint\\\hline
    46 - 48&	left Shadow Hand base translation\\\hline
    49 - 51&	left Shadow Hand base rotation\\
    \bottomrule %添加表格底部粗线
    \end{tabular}
    \label{scissors_action}
\end{table}

\paragraph{Rewards}
The reward consists of three parts: the distance from the left hand to the scissors' left handle, the distance from the right hand to the scissors' right handle, and the target angle at which the scissors need to be opened.
The distance between the scissors dof angle and target dof angle $d_{target}$ is given by $d_{target}=\Vert x_{scissors dof}-x_{target dof} \Vert$. the distance from the left hand to the scissors left handle $d_{left}$ is given by $d_{left}=\Vert x_{lhand}-x_{lhandle} \Vert_2$. the distance from the right hand to the scissors left handle $d_{right}$ is given by $d_{right}=\Vert x_{rhand}-x_{rhandle} \Vert_2$. The reward is given by this specific formula:
\begin{equation}
    r = 2 - d_{left} - d_{right} + (0.59 - d_{target}) * 5
\end{equation}

\subsubsection{Re Orientation}

This environment involves two hands and two objects. Each hand holds an object and we need to reorient the object to the target orientation.

\paragraph{Observations}
The 446-dimensional observation space as shown in Table \ref{re_orientation_obs}. 

\begin{table}[htbp]
    \centering
    \caption{Observation space of Re Orientation.}
    \begin{tabular}{c|c}
    \toprule  %添加表格头部粗线
    Index&Description\\\hline
    0 - 397&	dual hands observation shown in Table \ref{observation_dual hands}\\\hline
    398 - 404&	object1 pose\\\hline
    405 - 407&	object1 linear velocity\\\hline
    408 - 410&	object1 angle velocity\\\hline
    411 - 417&	goal1 pose\\\hline
    418 - 421&	goal1 rot - object rot\\
    422 - 428&	object2 pose\\\hline
    429 - 431&	object2 linear velocity\\\hline
    432 - 434&	object2 angle velocity\\\hline
    435 - 441&	goal2 pose\\\hline
    442 - 445&	goal2 rot - object2 rot\\

    \bottomrule %添加表格底部粗线
    \end{tabular}
    \label{re_orientation_obs}
\end{table}

\paragraph{Actions}
The 52-dimensional action space as shown in Table \ref{re_orientation_action}. 

\begin{table}[htbp]
    \centering
    \caption{Action space of Re Orientation.}
    \begin{tabular}{c|c}
    \toprule  %添加表格头部粗线
    Index&Description\\\hline
    0 - 19&	right Shadow Hand actuated joint\\\hline
    20 - 22&	right Shadow Hand base translation\\\hline
    23 - 25&	right Shadow Hand base rotation\\\hline
    26 - 45&	left Shadow Hand actuated joint\\\hline
    46 - 48&	left Shadow Hand base translation\\\hline
    49 - 51&	left Shadow Hand base rotation\\
    \bottomrule %添加表格底部粗线
    \end{tabular}
    \label{re_orientation_action}
\end{table}

\paragraph{Rewards}
The reward consists of three parts: the distance from the left object to the left object goal, the distance from the right object to the right object goal, and the distance between the object and desired goal.
The distance between the object and desired goal $d_{target}$ is given by $d_{target}=2 * \arcsin{q_{object1} * q_{target}} + 2 * \arcsin{q_{object2} * q_{target}}$. the distance from the left hand to the scissors left handle $d_{left}$ is given by $d_{left}=\Vert x_{lhand}-x_{lhandle} \Vert_2$. the distance from the right hand to the scissors left handle $d_{right}$ is given by $d_{right}=\Vert x_{rhand}-x_{rhandle} \Vert_2$. The reward is given by this specific formula:
\begin{equation}
    r = d_{left} * -10 + d_{right} * -10 + d_{target} * 1.5
\end{equation}

\subsubsection{Open Pen Cap}

This environment involves two hands and a pen, we need to use two hand
to open the pen cap.

\paragraph{Observations}
The 428-dimensional observation space as shown in Table \ref{pen_cap_obs}. 

\begin{table}[htbp]
    \centering
    \caption{Observation space of Open Pen Cap.}
    \begin{tabular}{c|c}
    \toprule  %添加表格头部粗线
    Index&Description\\\hline
    0 - 397&	dual hands observation shown in Table \ref{observation_dual hands}\\\hline
    398 - 404&	pen pose\\\hline
    405 - 407&	pen linear velocity\\\hline
    408 - 410&	pen angle velocity\\\hline
    411 - 417&	goal pose\\\hline
    418 - 421&	goal rot - object rot\\
    422 - 424&	pen body position\\\hline
    425 - 427&	pen cap position\\

    \bottomrule %添加表格底部粗线
    \end{tabular}
    \label{pen_cap_obs}
\end{table}

\paragraph{Actions}
The 52-dimensional action space as shown in Table \ref{pen_cap_action}. 

\begin{table}[htbp]
    \centering
    \caption{Action space of Open Pen Cap.}
    \begin{tabular}{c|c}
    \toprule  %添加表格头部粗线
    Index&Description\\\hline
    0 - 19&	right Shadow Hand actuated joint\\\hline
    20 - 22&	right Shadow Hand base translation\\\hline
    23 - 25&	right Shadow Hand base rotation\\\hline
    26 - 45&	left Shadow Hand actuated joint\\\hline
    46 - 48&	left Shadow Hand base translation\\\hline
    49 - 51&	left Shadow Hand base rotation\\
    \bottomrule %添加表格底部粗线
    \end{tabular}
    \label{pen_cap_action}
\end{table}

\paragraph{Rewards}
The reward consists of three parts: the distance from the left hand to the pen body, the distance from the right hand to the pen cap, and the distance between the pen body and pen cap.
The distance between the pen body and pen cap $d_{target}$ is given by $d_{target}=\Vert x_{pen body}-x_{pen cap} \Vert$. the distance from the left hand to the scissors left handle $d_{left}$ is given by $d_{left}=\Vert x_{lhand}-x_{pen body} \Vert_2$. the distance from the right hand to the scissors left handle $d_{right}$ is given by $d_{right}=\Vert x_{rhand}-x_{pen cap} \Vert_2$. The reward is given by this specific formula:
\begin{equation}
    r = exp(-10 * d_{left}) + exp(-10 * d_{right}) + d_{target} * 5 - 0.8
\end{equation}

\subsubsection{Switch}

This environment involves dual hands and a bottle, we need to use dual hand fingers to press the desired button.

\paragraph{Observations}
The 428-dimensional observation space as shown in Table \ref{switch_obs}. 

\begin{table}[htbp]
    \centering
    \caption{Observation space of Switch.}
    \begin{tabular}{c|c}
    \toprule  %添加表格头部粗线
    Index&Description\\\hline
    0 - 397&	dual hands observation shown in Table \ref{observation_dual hands}\\\hline
    398 - 404&	switch1 pose\\\hline
    405 - 407&	switch1 linear velocity\\\hline
    408 - 410&	switch1 angle velocity\\\hline
    411 - 417&	goal pose\\\hline
    418 - 421&	goal rot - object rot\\
    422 - 424&	switch1 position\\\hline
    425 - 427&	switch2 position\\

    \bottomrule %添加表格底部粗线
    \end{tabular}
    \label{switch_obs}
\end{table}

\paragraph{Actions}
The 52-dimensional action space as shown in Table \ref{switch_action}. 

\begin{table}[htbp]
    \centering
    \caption{Action space of Switch.}
    \begin{tabular}{c|c}
    \toprule  %添加表格头部粗线
    Index&Description\\\hline
    0 - 19&	right Shadow Hand actuated joint\\\hline
    20 - 22&	right Shadow Hand base translation\\\hline
    23 - 25&	right Shadow Hand base rotation\\\hline
    26 - 45&	left Shadow Hand actuated joint\\\hline
    46 - 48&	left Shadow Hand base translation\\\hline
    49 - 51&	left Shadow Hand base rotation\\
    \bottomrule %添加表格底部粗线
    \end{tabular}
    \label{switch_action}
\end{table}

\paragraph{Rewards}
The reward consists of three parts: the distance from the left hand to the left switch, the distance from the right hand to the right switch, and the distance between the button and button's desired goal.
The distance between the button and the button's desired goal $d_{target}$ is given by $d_{target}=\Vert x_{button1}-x_{target1} \Vert_2 + \Vert x_{button2}-x_{target2} \Vert_2$. the distance from the left hand to the scissors left handle $d_{left}$ is given by $d_{left}=\Vert x_{lhand}-x_{switch1} \Vert_2$. the distance from the right hand to the scissors left handle $d_{right}$ is given by $d_{right}=\Vert x_{rhand}-x_{switch2} \Vert_2$. The reward is given by this specific formula:
\begin{equation}
    r = 2 - d_{left} - d_{right} + (1.4 - d_{target}) * 50
\end{equation}

\subsubsection{Stack Block}

This environment involves dual hands and two blocks, and we need to stack the block as a tower.

\paragraph{Observations}
The 428-dimensional observation space as shown in Table \ref{stack_block_obs}. 

\begin{table}[htbp]
    \centering
    \caption{Observation space of Stack Block.}
    \begin{tabular}{c|c}
    \toprule  %添加表格头部粗线
    Index&Description\\\hline
    0 - 397&	dual hands observation shown in Table \ref{observation_dual hands}\\\hline
    398 - 404&	block1 pose\\\hline
    405 - 407&	block1 linear velocity\\\hline
    408 - 410&	block1 angle velocity\\\hline
    411 - 417&	goal pose\\\hline
    418 - 421&	goal rot - object rot\\
    422 - 424&	block1 position\\\hline
    425 - 427&	block2 position\\

    \bottomrule %添加表格底部粗线
    \end{tabular}
    \label{stack_block_obs}
\end{table}

\paragraph{Actions}
The 52-dimensional action space as shown in Table \ref{stack_block_action}. 
\begin{table}[htbp]
    \centering
    \caption{Action space of Stack Block.}
    \begin{tabular}{c|c}
    \toprule  %添加表格头部粗线
    Index&Description\\\hline
    0 - 19&	right Shadow Hand actuated joint\\\hline
    20 - 22&	right Shadow Hand base translation\\\hline
    23 - 25&	right Shadow Hand base rotation\\\hline
    26 - 45&	left Shadow Hand actuated joint\\\hline
    46 - 48&	left Shadow Hand base translation\\\hline
    49 - 51&	left Shadow Hand base rotation\\
    \bottomrule %添加表格底部粗线
    \end{tabular}
    \label{stack_block_action}
\end{table}

\paragraph{Rewards}
The reward consists of three parts: the distance from the left hand to block1, the distance from the right hand to block2, and the distance between the block and desired goal.
The distance between the block and desired goal $d_{target}$ is given by $d_{target}=\Vert x_{block1}-x_{target1} \Vert_2 + \Vert x_{block2}-x_{target2} \Vert_2$. the distance from the left hand to the block1 $d_{left}$ is given by $d_{left}=\Vert x_{lhand}-x_{block1} \Vert_2$. the distance from the right hand to the block2 $d_{right}$ is given by $d_{right}=\Vert x_{rhand}-x_{block2} \Vert_2$. The reward is given by this specific formula:
\begin{equation}
    r = 1.5 - d_{left} - d_{right} + (0.24 - d_{target}) * 2
\end{equation}

\subsubsection{Pour Water}

This environment involves two hands and a bottle, we need to Hold the kettle with one hand and the bucket with the other hand, and pour the water from the kettle into the bucket. In the practice task in Isaac Gym, we use many small balls to simulate the water.

\paragraph{Observations}
The 428-dimensional observation space as shown in Table \ref{pour_water_obs}. 

\begin{table}[htbp]
    \centering
    \caption{Observation space of Pour Water.}
    \begin{tabular}{c|c}
    \toprule  %添加表格头部粗线
    Index&Description\\\hline
    0 - 397&	dual hands observation shown in Table \ref{observation_dual hands}\\\hline
    398 - 404&	kettle pose\\\hline
    405 - 407&	kettle linear velocity\\\hline
    408 - 410&	kettle angle velocity\\\hline
    411 - 417&	goal pose\\\hline
    418 - 421&	goal rot - object rot\\
    422 - 424&	kettle handle position\\\hline
    425 - 427&	bucket position\\
    \bottomrule %添加表格底部粗线
    \end{tabular}
    \label{pour_water_obs}
\end{table}

\paragraph{Actions}
The 52-dimensional action space as shown in Table \ref{pour_water_action}. 
\begin{table}[htbp]
    \centering
    \caption{Action space of Pour Water.}
    \begin{tabular}{c|c}
    \toprule  %添加表格头部粗线
    Index&Description\\\hline
    0 - 19&	right Shadow Hand actuated joint\\\hline
    20 - 22&	right Shadow Hand base translation\\\hline
    23 - 25&	right Shadow Hand base rotation\\\hline
    26 - 45&	left Shadow Hand actuated joint\\\hline
    46 - 48&	left Shadow Hand base translation\\\hline
    49 - 51&	left Shadow Hand base rotation\\
    \bottomrule %添加表格底部粗线
    \end{tabular}
    \label{pour_water_action}
\end{table}

\paragraph{Rewards}
The reward consists of three parts: the distance from the left hand to the bucket, the distance from the right hand to the kettle, and the distance between the kettle spout and desired goal.
The distance between the kettle spout and desired goal $d_{target}$ is given by $d_{target}=\Vert x_{spout}-x_{goal} \Vert_2$. the distance from the left hand to the bucket $d_{left}$ is given by $d_{left}=\Vert x_{lhand}-x_{bucket} \Vert_2$. the distance from the right hand to the kettle $d_{right}$ is given by $d_{right}=\Vert x_{rhand}-x_{kettle} \Vert_2$. The reward is given by this specific formula:
\begin{equation}
    r = 1 - d_{left} - d_{right} + (0.5 - d_{target}) * 2
\end{equation}

\subsection{Offline Data Collection}
\label{app:collection}
We follow the data collection of D4RL\cite{d4rl} mujoco tasks. The medium dataset is generated by first training a policy online using PPO, early-stopping the training, and collecting $10^6$ samples $(s_t, a_t, s_{t+1}, r_t)$ using this medium policy. The
random dataset is collected by a randomly initialized policy and contains $10^6$ samples. The replay dataset consists of $10^6$ experienced samples during training of the medium policy. The medium-expert dataset contains $2 \times 10^6$ samples by mixing equal amounts of samples collected by expert policy and medium policy. To facilitate comparison across tasks, following the setting of D4RL\cite{d4rl}, we normalize scores for each task to the range between $0$ and $100$ , by computing normalized score $=100 * \frac{\text { return-random return }}{\text { expert return-random return }}$. A normalized score of $0$ corresponds to the average return of an agent taking actions uniformly at random across the action space. A score of $100$ corresponds to the average return of an expert policy.

\section{Training details}\label{Appendix:B}

Isaac Gym is different from other simulators in that it can simulate completely on the GPU, so there is no need to exchange data between the GPU and the CPU during the training process. Therefore we reproduced the existing RL algorithm in our Github repository to accommodate this feature. We implemented many different algorithms in the comprehensive RL domain, but only evaluated some of them. We will give a brief introduction to these algorithms below and give the hyperparameters of the algorithms we used in our evaluation.

\subsection{Single-agent algorithms}
\subsubsection{Trust Region Policy Optimization}
TRPO is a basic policy optimization algorithm, with theoretically justified monotonic improvement. Based on the theorem1 in the original paper by John Schulman et. al.
$\eta(\pi_{new})\geq L_{old}(\pi_{new})-\frac{
4\epsilon \gamma}{(1-\gamma)^2}\alpha^2$,where$\epsilon = \mathop{\max}\limits_{s,a}|A_{\pi}(s,a)|$,$\eta$ is the objective function and $L_{old}$ is a surrogate objective: $L_{\pi}(\hat{\pi}) = \eta(\pi)+E_{s \sim \rho_{\pi},a\sim \pi}(A_{\pi}(s,a))$, providing feasible approximation of $\eta$ according to the theorem. To empirically allow for larger update steps, the optimization problem is adjusted to $\pi_{\theta_{new}} = \mathop{\max}\limits_{\theta}L_{\theta_{old}}(\theta)$ subject to $D_{KL}^{max}(\theta_{old},\theta) \leq \delta$. To yield a practical algorithm, TRPO makes a bit of approximation like optimizing with conjugate gradient method followed by a line search. 
% \begin{table}[htbp]
%     \centering
%     \begin{tabular}{c|c|c}
%     \toprule  %添加表格头部粗线
%     Hyperparameters & HandCatch & HandLift\\\hline
%     Num mini-batches & 4 & 4\\
%     Num opt-epochs & 5 & 10\\
%     Num episode-length & 8 & 20\\\hline
%     Hidden size & [1024, 1024, 512] & [1024, 1024, 512]\\
%     Clip range & 0.2 & 0.2\\
%     Max grad norm & 10 & 10\\
%     Learning rate & 1.e-3 & 1.e-3\\
%     Discount ($\gamma$) & 0.99 & 0.99\\
%     Lambda ($\lambda$) & 0.95 & 0.95\\
%     Init noise std & 0.8 & 0.8\\\hline
%     Damping & 0.1 & 0.1\\
%     Cg nsteps & 3 & 3\\
%     Max kl & 0.1 & 0.1\\
%     Max backtrack & 10 & 10\\
%     Accept ratio & 0.01 & 0.01\\
%     Step fraction & 1 & 1\\
%     \bottomrule %添加表格底部粗线
%     \end{tabular}
%     \caption{}
%     \label{trpo_para}
% \end{table}

\subsubsection{Proximal Policy Optimization}
PPO is a policy optimization algorithm enjoying simpler implementation, more general application and better sample complexity over TRPO. Based on the surrogate objective in TRPO: $L^{CPI}(\theta) = \hat{E}_t[r_t(\theta)\hat{A}_t]$, PPO proposed a new approximate surrogate function $L^{CLIP}(\theta) = \hat{E}_t[min(r_t(\theta)\hat{A}_t,clip(r_t(\theta),1-\epsilon,1+\epsilon)\hat{A}_t)]$, which restricts policy optimization step by removing the incentive for $r_t$ to move outside of the interval $[1 - \epsilon,1 + \epsilon]$. Another alternative surrogate objective is given by incorporating a penalty on KL divergence, and adapting the penalty coefficient. During traning, PPO uses a combined objective, consisting of surrogate objective for the policy, value function loss for the critic and a bonus entropy term: $L^{CLIP+VF+S}(\theta) = \hat{E}_t[L_t^{CLIP}(\theta)-c_1 L_t^{VF}(\theta)+c_2S[\pi_{\theta}](s_t)]$.
\begin{table}[htbp]
    \centering
    \caption{Hyperparameters of PPO.}
    \begin{tabular}{c|c|c|c}
    \toprule  %添加表格头部粗线
    Hyperparameters & Other Tasks & Lift Underarm & Stack Block\\\hline
    Num mini-batches & 4 & 4 & 8\\
    Num opt-epochs & 5 & 10 & 2\\
    Num episode-length & 8 & 20 & 8\\\hline
    Hidden size & [1024, 1024, 512] & [1024, 1024, 512] & [1024, 1024, 512]\\
    Clip range & 0.2 & 0.2 & 0.2\\
    Max grad norm & 1 & 1 & 1\\
    Learning rate & 3.e-4 & 3.e-4 & 3.e-4\\
    Discount ($\gamma$) & 0.96 & 0.96 & 0.9\\
    GAE lambda ($\lambda$) & 0.95 & 0.95& 0.95\\
    Init noise std & 0.8 & 0.8& 0.8\\\hline
    Desired kl & 0.016 & 0.016 & 0.016\\
    Ent-coef & 0 & 0 & 0\\
    \bottomrule %添加表格底部粗线
    \end{tabular}
    \label{ppo_para}
\end{table}

\subsubsection{Deep Deterministic Policy Gradient}
DDPG, based on the DPG algorithm, is a model-free, off-policy actor-critic algorithm using deep function approximators that can learn policies in high-dimensional, continuous action spaces. It uses a copy of the actor and critic networks $Q^{\prime}(s,a|\theta^{Q^{\prime}})$ and  $\mu^{\prime}(s|\theta^{\mu^{\prime}})$ to calculate the target values, and use "soft" target updates to update the target networks more stably by having them slowly track the learned networks: $\theta^{\prime} \leftarrow \tau\theta+(1-\tau)\theta^{\prime}$ with $\tau\ll 1$. It follows an exploration policy  $\mu^{\prime}$ by adding noise sampled from a noise process $\mathbf{N}$: $\mu^{\prime}(S_t) = \mu(s_t|\theta_t^{\mu}) + \mathbf{N}_t$. 
The critic is updated by minimizing the loss: $L(\phi)=\frac{1}{N}\sum_i(y_i -Q(s_i,a_i|\theta^{Q}))^2$ where $y_i=r_i+\gamma Q^{\prime}(s_{i+1},\mu^{\prime}(s_{i+1}|\theta^{\mu^{\prime}})|\theta^{Q^{\prime}})$, and the actor is updated using sampled policy gradient: 
$\nabla_{\theta}J\approx\frac{1}{N}\sum_i\nabla_{\alpha}Q(s,a|\theta^{Q})|_{s=s_i,a=\mu(s_i)}\nabla_{\theta\mu}\mu(s|\theta^{\mu})|_{s_i}$.

% \begin{table}[htbp]
%     \centering
%     \begin{tabular}{c|c|c}
%     \toprule  %添加表格头部粗线
%     Hyperparameters & HandCatch & HandLift\\\hline
%     Num mini-batches & 4 & 4\\
%     Num opt-epochs & 5 & 10\\
%     Num batch-size & 64 & 64\\
%     Num episode-length & 8 & 20\\\hline
%     Hidden size & [1024, 1024, 512] & [1024, 1024, 512]\\
%     Replay Size & 10000 & 10000\\
%     Polyak & 0.995 & 0.995\\
%     Reward scale & 1 & 1\\
%     Clip range & 0.2 & 0.2\\
%     Max grad norm & 1 & 1\\
%     Learning rate & 1.e-4 & 1.e-4\\
%     Discount ($\gamma$) & 0.96 & 0.96\\
%     GAE lambda ($\lambda$) & 0.95 & 0.95\\
%     Init noise std & 1 & 1\\\hline
%     Act noise & 0.1 & 0.1\\
%     Target noise & 0.2 & 0.2\\
%     Noise Clip & 0.5 & 0.5\\
%     \bottomrule %添加表格底部粗线
%     \end{tabular}
%     \caption{}
%     \label{ddpg_para}
% \end{table}

\subsubsection{Twin Delayed Deep Deterministic policy gradient}
TD3 is an actor-critic algorithm which applies its modifications to the state of the art actor-critic method for continuous control, DDPG. It focused on two outcomes that occur as the result of estimation error, overestimation bias and a high variance build-up. It uses Clipped Double Q-learning method to reduce overestimation bias: $y \leftarrow r+\gamma min_{i=1,2}Q_{\theta_i^{\prime}}(s^{\prime},\tilde{a})$, where $\tilde{a} \leftarrow \pi_{\phi^{\prime}}(s)+\epsilon,\epsilon \sim clip(\mathbf{N}(0, \tilde{\sigma}),-c,c)$, which uses target policy smoothing regularization to avoid overfitting and enforce the value similarity between similar actions, It uses delayed policy and target network updates to ensure small value error.
% \begin{table}[htbp]
%     \centering
%     \begin{tabular}{c|c|c}
%     \toprule  %添加表格头部粗线
%     Hyperparameters & HandCatch & HandLift\\\hline
%     Num mini-batches & 4 & 4\\
%     Num opt-epochs & 5 & 10\\
%     Num batch-size & 100 & 100\\
%     Num episode-length & 8 & 20\\\hline
%     Hidden size & [1024, 1024, 512] & [1024, 1024, 512]\\
%     Replay Size & 10000 & 10000\\
%     Polyak & 0.995 & 0.995\\
%     Reward scale & 1 & 1\\
%     Clip range & 0.2 & 0.2\\
%     Max grad norm & 1 & 1\\
%     Learning rate & 1.e-4 & 1.e-4\\
%     Discount ($\gamma$) & 0.96 & 0.96\\
%     GAE lambda ($\lambda$) & 0.95 & 0.95\\
%     Init noise std & 1 & 1\\\hline
%     Policy delay & 2 & 2\\
%     Act noise & 0.1 & 0.1\\
%     Target noise & 0.2 & 0.2\\
%     Noise Clip & 0.5 & 0.5\\
%     \bottomrule %添加表格底部粗线
%     \end{tabular}
%     \caption{}
%     \label{td3_para}
% \end{table}

\subsubsection{Soft Actor-Critic}
SAC is an off-policy maximum entropy actor-critic algorithm. It considers a more general maximum entropy objective: $J(\pi)=\sum_{t=0}^{T}\mathbf{E}_{(s_t, a_t)\sim \mathbf{D}_{\pi}}[r(s_t, a_t)+\alpha \mathbf{H}(\pi(\cdot|s_t))]$, in which the temperature parameter $\alpha$ determines the relative importance of the entropy term. The soft value function $V_{\psi}(s_t)$ is trained to minimize the squared residual error: $L_v(\psi) = \mathbf{E}_{s_t \sim \rho}[\frac{1}{2} (V_{\psi}(s_t)-\mathbf{E}_{a_t\sim \pi_{\phi}}[Q_{\theta}(s_t,a_t)-\log \pi_{\phi}(a_t|s_t)])^2]$. The soft Q-function parameters can be trained to minimize the soft Bellman residual: $L_Q(\theta)=\mathbf{E}_{(s_t, a_t)\sim \mathbf{D}}[\frac{1}{2}(Q_\theta(s_t, a_t)-\hat{Q}(s_t, a_t))^2]$, in which $\hat{Q}(s_t, a_t)=r(s_t, a_t)+\gamma\mathbf{E}_{s_{t+1}\sim p}[V_{\bar{\phi}}(s_{t+1})]$ and $V_{\bar{\phi}}$ is the target value network. The policy parameters can be learned by directly minimizing the expected KL-divergence: $KL_{\pi}(\phi)=\mathbf{E}_{s_t \sim \rho}[D_{KL}(\pi_{\phi}(\cdot|s_t)||\frac{exp(Q_{\theta}(s_t,\cdot))}{Z_{\theta}(s_t)})]$, in which $Z_{\theta}(s_t)$ normalizes the distribution.

\begin{table}[htbp]
    \centering
    \caption{Hyperparameters of SAC.}
    \begin{tabular}{c|c|c|c}
    \toprule  %添加表格头部粗线
    Hyperparameters&Other Tasks&Lift Underarm & Stack Block\\\hline
    Num opt-epochs & 1 & 1 & 1\\
    Num mini-batches&4 & 4 &4\\\hline
    Hidden size & [1024, 1024, 1024] & [1024, 1024, 1024] & [1024, 1024, 1024]\\
    Learning rate & 3.e-4 & 3.e-4 & 3.e-4 \\
    ReplayBuffer size& 5000 & 5000 & 5000\\
    Discount ($\gamma$)& 0.96 & 0.96 & 0.96\\
    Polyak ($1-\tau$)& 0.99 & 0.99 & 0.99\\
    Entropy coef & 0.2 & 0.2 & 0.2\\
    Reward scale & 1 & 1 & 1 \\
    Max grad norm & 1 & 1 & 1\\
    Batch size & 32 & 32 & 32\\
    
    \bottomrule %添加表格底部粗线
    \end{tabular}
    \label{hp of sac}
\end{table}

\subsection{Multi-agent algorithms }
\subsubsection{Independent Proximal Policy Optimization}
IPPO (Independent PPO) is a multi-agent variant of proximal policy optimization(PPO). It uses PPO to learn decentralized policies $\pi^{i}$ for agents with indicidual policy clippng based on the objective: $\mathbf{L}^{i}(\theta)=\mathbf{E}_{s_t^i,a_t^i}[min(\frac{\pi_{\theta}(a_t^i|s_t^i)}{\pi_{\theta_{old}}(a_t^i|s_t^i)}A_t^i,clip(\frac{\pi_{\theta}(a_t^i|s_t^i)}{\pi_{\theta_{old}}(a_t^i|s_t^i)},1-\epsilon,1+\epsilon)A_t^i)]$, and the advantage function is based on independent learning, where each agent a learns a local observation based critic $V_{\phi}(z_t^i)$ parameterised by $\phi$ using GAE. Additionally, it uses value clipping to restrict the update of critic function for each agent $i$:
% 这里的a替换一下preliminary里面的定义
$\mathbf{L}^i(\phi)=\mathbf{E}_{z_t^i}[min\{(V_{\phi}(z_t^i)-\hat{V_t^i})^2, (V_{\phi_{old}}(z_t^i)+clip(V_{\phi}(z_t^i)-V_{\phi_{old}}(z_t^i),-\epsilon,+\epsilon)-\hat{V_t^i})^2\}]$. The overall learning loss additionally adds an entropy regularization term of policy $\pi ^i$.

% \begin{table}[htbp]
%     \centering
%     \begin{tabular}{c|c|c}
%     \toprule  %添加表格头部粗线
%     Hyperparameters&HandOver&CatchUnderarm\\\hline
%     Num mini-batch&&\\
%     Num epoch&&\\
%     len. data chunk&&\\\hline
%     Value loss coeff. $c_1$&&\\
%     Entropy coeff. $c_2$&&\\
%     Huber loss delta ($\delta$)&&\\
%     Discount ($\gamma$)&&\\
%     Clip parameter ($\epsilon$)&&\\
    
%     \bottomrule %添加表格底部粗线
%     \end{tabular}
%     \caption{}
%     \label{table1}
% \end{table}

\subsubsection{Heterogenous-Agent Trust Region Policy Optimization}
HATRPO is a multi-agent algorithm developed from TRPO. With the advantage decomposition lemma, the algorithm is proposed to implement a multi-agent policy iteration procedure with monotonic improvement guarantee. It requires no homogeneity of agents, nor any restrictive assumptions on the decomposibility of joint Q-functions. At each iteration k+1, given a random permutation of agents $i_{1:n}$, agent $i_m$ sequentially optimizes its own policy parameter $\theta_{k+1}^{i_m}$ by maximizing the objective: $\theta_{k+1}^{i_m}=arg max_{\theta^{i_m}}\mathbf{E}_{s\sim \rho_{\theta_k},a^{i_{1:m-1}}\sim \pi_{\theta_{k+1}^{i_{1:m-1}}}^{i_{1:m-1}},a^{i_m}\sim \pi_{\theta^{i_m}}^{i_m}}[A_{\pi_{\theta_k}}^{i_m}(s,a^{i_{1:m-1}},a^{i_m})]$, subject to $\mathbf{E}_{s\sim \rho_{\theta_k}}[D_{KL}(\pi_{\theta_{k}^{i_m}}^{i_m}(\cdot|s),\pi_{\theta^{i_m}}^{i_m}(\cdot|s))]\le \delta$. Apply a linear approximation to the objective function and a quadratic approximation to the KL constraint: $\theta_{k+1}^{i_m}=\theta_{k}^{i_m}+\alpha^j\sqrt{\frac{2\delta}{g_k^{i_m}(H_k^{i_m})^{-1}g_k^{i_m}}}(H_k^{i_m})^{-1}g_k^{i_m}$, in which $H_k^{i_m}$ is the Hessian of the expected KL-divergence, $g_k^{i_m}$ is the gradient of the objective function, and $\alpha^j < 1$ is a positive coefficient. Estimate the advantage function $\mathbf{E}[A_{\pi_{\theta_k}}^{i_m}(s,a^{i_{1:m-1}},a_{i_m})]$ with $(\frac{\pi_{\theta}^{i_m}(a_{i_m}|s)}{\pi_{\theta_k}^{i_m}(a_{i_m}|s)}-1)M^{i_{1:m}}(s,a)$, where $M^{i_{1:m}}=\frac{\bar{\pi}^{i_{1:m-1}}}{\pi_{i_{1:m-1}}}\hat{A}(s,a)$ and $\bar{\pi}^{i_{1:m-1}}=\prod_{j=1}^{m-1}\bar{\pi}^{i_j}$ is the policies of agents $i_{1:m-1}$ just updated in the same iteration k+1.

% \begin{table}[htbp]
%     \centering
%     \begin{tabular}{c|c|c}
%     \toprule  %添加表格头部粗线
%     Hyperparameters & HandCatch & HandLift\\\hline
%     Num mini-batches&&\\
%     Num opt-epochs&&\\
%     Num episode-length&&\\\hline
%     Hidden size&&\\
%     Use popart&&\\
%     Use value norm&&\\
%     Use proper time limits&&\\
%     Use huber loss&&\\
%     KL threshold&&\\
%     Huber delta&&\\
%     Clip range&&\\
%     Max grad norm&&\\
%     Learning rate&&\\
%     Opt-eps&&\\
%     Discount ($\gamma$)&&\\
%     GAE lambda ($\lambda$)&&\\
%     Value loss coeff.&&\\
%     Entropy coeff.&&\\
%     \bottomrule %添加表格底部粗线
%     \end{tabular}
%     \caption{}
%     \label{mappo_para}
% \end{table}

\subsubsection{Heterogeneous-Agent Proximal Policy Optimisation}
HAPPO is a multi-agent policy optimization algorithm that follows the centralized training decentralized execution (CTDE) paradigm. HAPPO doesn't assume homogeneous agents and doesn't require decomposibility of the joint value function. The theoretical core of extending PPO to multi-agent settings is the advantage decomposition lemma(Lemma 1 in the original paper). As a result of it, similar to single agent PPO, we have a theoretical monotonic improvement guarantee for the multi-agent setting: $J(\overline{\pi}) \geq J(\pi) + \sigma_{m=1}^{n}[L_{\pi}^{i_{1:m}}(\overline{\pi}^{i_{1:m-1}},\overline{\pi}^{i_m})-CD^{max}_{KL}(\pi^{i_m},\overline{\pi}^{i_m})]$(Lemma 2 in the original paper). This lemma yields a similar policy optimization iteration: $\pi^{i_m}_{k+1} = arg \mathop{\max}\limits_{\pi^{i_m}}[L_{\pi}^{i_{1:m}}({\pi}^{i_{1:m-1}},{\pi}^{i_m})-CD^{max}_{KL}(\pi^{i_m},{\pi}^{i_m})]$. To avoid maintaining value functions for each single agent, the following proposition is used: $E
[A^{i_m}_{\pi}(s,a^{i_{1:m-1}},a^{i_m})]=E[(\dfrac{\hat{\pi}^{i_m}(a^{i_m}|s)}{\pi^{i_m}(a^{i_m}|s)}-1)\dfrac{\overline{\pi}^{i_{1:m-1}}(a^{i_{1:m-1}}|s)}{{\pi}^{i_{1:m-1}}(a^{i_{1:m-1}}|s)}A_{\pi}(s,a)])$, so that it only need to keep one value function $A_{\pi}(s,a)$ for all agents.Finally, it uses the clipping trick similar to single agent PPO, obtaining the final practical algorithm, for details, please refer to (11) in the original paper.

\begin{table}[htbp]
    \centering
    \caption{Hyperparameters of HAPPO.}
    \begin{tabular}{c|c|c|c}
    \toprule  %添加表格头部粗线
    Hyperparameters & Other Tasks & Lift Underarm & Stack Block\\\hline
    Num mini-batches & 1 & 1 & 1\\
    Num opt-epochs & 5 & 10 & 5\\
    Num episode-length & 8 & 20 & 8\\\hline
    Hidden size & [1024, 1024, 512] & [1024, 1024, 512] & [1024, 1024, 512]\\
    Use popart & True & True & True\\
    Use value norm & True & True & True\\
    Use proper time limits & False & False & False\\
    Use huber loss & True & True & True\\
    Huber delta & 10 & 10 & 10\\
    Replay Size & 10000 & 10000 & 10000\\
    Polyak & 0.995 & 0.995 & 0.995\\
    Reward scale & 1 & 1 & 1\\
    Clip range & 0.2 & 0.2 & 0.2\\
    Max grad norm & 1 & 1 & 1\\
    Learning rate & 1.e-4 & 1.e-4 & 1.e-4\\
    Discount ($\gamma$) & 0.96 & 0.96 & 0.96\\
    GAE lambda ($\lambda$) & 0.95 & 0.95 & 0.95\\
    Init noise std & 1 & 1 & 1\\\hline
    Ent-coef & 0 & 0 & 0\\

    \bottomrule %添加表格底部粗线
    \end{tabular}
    \label{happo_para}
\end{table}

\subsubsection{Multi-Agent Proximal Policy Optimization}
MAPPO (Multi-Agent PPO) is an application of the actor-critic single-agent PPO algorithm to multi-agent tasks. It follows the CTDE structure. Each agent i follows a shared policy $\pi_\theta(a_i|o_i)$ based on local observation $o_i=O(s;i)$ at global state $s$, takes its action $a_i$ and optimizes its reward $J(\theta)=E_{a^t,s^t}[\sum_t \gamma^t R(s^t, a^t)]$. The actor network maximizes: $L(\theta)=[\frac{1}{Bn}\sum_{i=1}^{B}\sum_{k=1}^{n} min(r_{\theta,i}^{(k)}A_i^{(k)}, clip(r_{\theta,i}^{(k)}, 1-\epsilon, 1+\epsilon)A_i^{(k)})]+\sigma\frac{1}{Bn}\sum_{i=1}^{B}\sum_{k=1}^n S[\pi_{\theta}(o_i^{(k)})]$, where $n$ refers to the agent number, $A_I^{(k)}$ is computed using GAE method, $S$ is policy entropy and $\sigma$ is entropy coefficient hyper-parameter. The critic network minimizes: $L(\phi)=\frac{1}{Bn}\sum_{i=1}^{B}\sum_{k=1}^{n}(max[(V_{\phi}(s_i^{(k)})-\hat{R_i})^2, (clip(V_{\phi}(s_i^{(k)}),V_{\phi_{old}}(s_i^{(k)})-\epsilon, V_{\phi_{old}}(s_i^{(k)})+\epsilon)-\hat{R_i})^2])$, where $\hat{R_i}$ is reward-to-go.

\begin{table}[htbp]
    \centering
    \caption{Hyperparameters of MAPPO.}
    \begin{tabular}{c|c|c|c}
    \toprule  %添加表格头部粗线
    Hyperparameters & HandCatch & HandLift & Stack Block\\\hline
    Num mini-batches & 1 & 1 & 1\\
    Num opt-epochs & 5 & 10 & 5\\
    Num episode-length & 8 & 20 & 8\\\hline
    Hidden size & [1024, 1024, 512] & [1024, 1024, 512] & [1024, 1024, 512]\\
    Use popart & True & True & True\\
    Use value norm & True & True & True\\
    Use proper time limits & False & False & False\\
    Use huber loss & True & True & True\\
    Huber delta & 10 & 10 & 10\\
    Clip range & 0.2 & 0.2 & 0.2\\
    Max grad norm & 10 & 10 & 10\\
    Learning rate & 5.e-4 & 5.e-4 & 5.e-4\\
    Opt-eps & 5.e-4 & 5.e-4 & 5.e-4\\
    Discount ($\gamma$) & 0.96 & 0.96 & 0.96\\
    GAE lambda ($\lambda$) & 0.95 & 0.95 & 0.95\\
    Std x coef & 1 & 1 & 1\\
    Std y coef & 0.5 & 0.5 & 0.5\\\hline
    Ent-coef & 0 & 0 & 0\\
    \bottomrule %添加表格底部粗线
    \end{tabular}
    \label{mappo_para}
\end{table}

\subsubsection{Multi-Agent Deep Deterministic Policy Gradient}
MADDPG (Multi-Agent DDPG) is an actor-critic deep policy gradient algorithm solving multi-agent tasks. Based on DDPG, it uses CTDE structure, in which the critic uses global information to optimize Q-function while training and the actor uses local observation to take actions while testing. For each agent i, update the critic by minimizing the loss function: $L^i(\phi)=\frac{1}{S}\sum_j(y^{j}-Q_i^{\mu}(\mathbf{x}^j,a_1^{j},\dots,a_N^{j}))^2$, where $y^j=r_i^j+\gamma Q_i^{\mu^{\prime}}(\mathbf{x}^{\prime j},a_1^{\prime},\dots,a_N^{\prime})|_{a_k^{\prime}=\mu_k^{\prime}(\sigma_k^j)}$, and update actor using the sampled policy gradient: $\nabla_{\theta_i}J\approx\frac{1}{S}\sum_j\nabla_{\theta_i}\mu_i(\sigma_i^j)\nabla_{a_i}Q_i^{\mu}(\mathbf{x}^{j},a_{i}^{j},\dots,a_{i},\dots,a_{N}^{j})|_{a_i=\mu_i(\sigma_i^j)}$, where $S$ is the size of the mini-batch. 

% \begin{table}[htbp]
%     \centering
%     \begin{tabular}{c|c|c}
%     \toprule  %添加表格头部粗线
%     Hyperparameters & HandCatch & HandLift\\\hline
%     Num mini-batches & 1 & 1\\
%     Num opt-epochs & 5 & 10\\
%     Num episode-length & 8 & 20\\\hline
%     Hidden size & [1024, 1024, 512] & [1024, 1024, 512]\\
%     Use popart & True & True\\
%     Use value norm & True & True\\
%     Use proper time limits & False & False\\
%     Use huber loss & True & True\\
%     Huber delta & 10 & 10\\
%     Clip range & 0.2 & 0.2\\
%     Max grad norm & 10 & 10\\
%     Learning rate & 5.e-4 & 5.e-4\\
%     Opt-eps & 5.e-4 & 5.e-4\\
%     Discount ($\gamma$) & 0.96 & 0.96\\
%     GAE lambda ($\lambda$) & 0.95 & 0.95\\
%     Std x coef & 1 & 1\\
%     Std y coef & 0.5 & 0.5\\\hline
%     Act noise & 0.1 & 0.1\\
%     Target noise & 0.2 & 0.2\\
%     Noise Clip & 0.5 & 0.5\\
%     \bottomrule %添加表格底部粗线
%     \end{tabular}
%     \caption{}
%     \label{maddpg_para}
% \end{table}

\subsection{Offline algorithms}

\subsubsection{BCQ}

BCQ constrains the selected actions to be in the action distribution of the dataset. It trains a Q-network $Q$, a perturbation network $\xi$, and a conditional VAE $G=\{E\left(\mu, \sigma | s, a\right), D\left(a | s, z \sim\left(\mu, \sigma\right)\right)\}$. The agent generates $n$ actions by $G$, adds small perturbations $\in [-\Phi ,\Phi ]$ on the actions using $\xi$, and then selects the action with the highest value in $Q$. The policy can be written as
    \[\begin{split}
	\pi(s)=\underset{a^{j}+\xi\left(s, a^{j}\right)}{\operatorname{argmax}} Q\left(s, a^{j}+\xi\left(s, a^{j}\right)\right), \quad
	\text{where } \left\{a^{j} \sim G(s)\right\}_{j=1}^{n}.
	\end{split}\]
$Q$ is updated by minimizing 
$\mathbb{E}_{(s, a, s') \sim \mathcal{D}}|Q(s,a)-y|^2, \quad \text{where } y=r + \gamma \hat{Q}(s',\hat{\pi}(s')).$
$y$ is calculated by the target networks $\hat{Q}$ and $\hat{\xi}$, where $\hat{\pi}$ is correspondingly the policy induced by $\hat{Q}$ and $\hat{\xi}$. $\xi_i$ is updated by maximizing $\mathbb{E}_{(s, a) \sim \mathcal{D}}Q\left(s, a+\xi\left(s, a\right)\right)$.

\subsubsection{TD3+BC}
TD3+BC simply adds the behavior clone term into the objective of policy optimization in TD3 to constrain the learned policy to be close to the behavior policy. Specifically, $$\pi=\underset{\pi}{\operatorname{argmax}} \mathbb{E}_{(s, a) \sim \mathcal{D}}\left[\lambda Q(s, \pi(s))-(\pi(s)-a)^{2}\right].$$

\begin{table}[t]
    \centering
    \caption{Hyperparameters of offline algorithms.}
    \begin{tabular}{c|c|c|c}
    \toprule 
    Hyperparameters& BCQ & TD3+BC & IQL\\\hline
    Hidden size & [400,300] & [256,256] & [256,256]\\
    Learning rate & 1.e-3 & 3.e-4 & 3.e-4 \\
    Discount ($\gamma$)& 0.99 & 0.99 & 0.99\\
    Polyak ($1-\tau$)& 0.995 & 0.995 & 0.995\\
    Batch size & 100 & 256 & 256\\
    $\Phi$ & 0.05 & - & -\\
    generated actions & 10 & - & -\\
    $\alpha$ & - & 0.2 & -\\
    $\beta$ & - & - &3.0\\
    $\tau$ (IQL) & - & - &0.7\\
    
    \bottomrule 
    \end{tabular}
    \label{hp of offline}
\end{table}

\subsubsection{IQL}

IQL avoids to query the values of any out-of-distribution actions without explicit constraints. It approximates an upper expectile of the value distribution by simply modifying the loss function in a SARSA-style TD backup, without ever using out-of-distribution actions in the target value. The V values are updated by minimizing $$\mathbb{E}_{(s, a) \sim \mathcal{D}}\left[L_{2}^{\tau}\left(Q(s, a)-V(s)\right)\right],$$ where $L_{2}^{\tau}(u)=|\tau-\mathbb{1}(u<0)| u^{2}$. And Q values are updated by minimizing $$\mathbb{E}_{\left(s, a, s^{\prime}\right) \sim \mathcal{D}}\left[\left(r(s, a)+\gamma V\left(s^{\prime}\right)-Q(s, a)\right)^{2}\right].$$
After the Q values have converged, the policy are updated by advantage-weighted behavioral cloning: $$\mathbb{E}_{(s, a) \sim \mathcal{D}}\left[\exp \left(\beta\left(Q(s, a)-V(s)\right)\right) \log \pi(a \mid s)\right].$$

Most of parameters of offline algorithms follow the official settings. We find that a small $\alpha$ for TD3+BC would achieve better performance and we choose $0.2$ rather than $2.5$ (official setting). BC is TD3+BC with $\alpha = 0$.

% \begin{table}[htbp]
%     \centering
%     \caption{The definition of the symbol.}
%     \begin{tabular}{c|c}
%     \toprule  %添加表格头部粗线
%     variable&notation\\\hline
%     space of states&$\rho$\\
%     space of state-action pairs&
%     $\mathbf{D}$\\
%     mini-batch&
%     $\mathbf{S}$\\
%     policy&$\pi$\\
%     policy of agent i&$\pi^i$\\
%     recently updated policy&$\bar{\pi}$\\
%     objective(total reward)&$J(\pi)$\\
%     reward to go&$\hat{R_i}$\\
%     expected KL-divergence&$KL_{\pi}(\phi)$\\
%     soft value function&$V_{\psi}$\\
%     value function loss&$L_{v}(\psi)$\\
%     soft Q-function loss$L_Q(\theta)$\\
%     loss function for critic&$L(\phi)$\\
%     loss function for agent i's critic&$L^i(\phi)$\\
%     loss function for actor&$L(\theta)$\\
%     loss function for agent i's actor&$L^i(\theta)$\\
%     advantage function for agent i&$A^i$\\
%     actor policy gradient&$\nabla_{\theta}J$\\
%     actor olicy gradient of agent i&$\nabla_{\theta_i}J$\\
    
%     \bottomrule %添加表格底部粗线
%     \end{tabular}
%     \label{table1}
% \end{table}

\subsection{Multi-task RL algorithms}

\subsubsection{Multi-task PPO/SAC/TRPO}

Multi-task PPO, Multi-task SAC, and Multi-task TRPO are basically the same as the original PPO, SAC, and TRPO, except for a small change called "disentangled alphas" in the Multi-task SAC algorithm. Alpha is the entropy coefficient used to control policy exploration. Disentangled alpha means that the learning of each task has a separate alpha coefficient for better exploration between different tasks.

\begin{table}[htbp]
    \centering
    \caption{Hyperparameters of Multi-task PPO.}
    \begin{tabular}{c|c}
    \toprule  %添加表格头部粗线
    Hyperparameters & MT1, MT4, and MT20\\\hline
    Num mini-batches & 4 \\
    Num opt-epochs & 5\\
    Num episode-length & 8\\\hline
    Hidden size & [2048, 1024, 512]\\
    Clip range & 0.2 \\
    Max grad norm & 1\\
    Learning rate & 3.e-4 \\
    Discount ($\gamma$) & 0.96\\
    GAE lambda ($\lambda$) & 0.95\\
    Init noise std & 0.8\\\hline
    Desired kl & 0.016\\
    Ent-coef & 0\\
    \bottomrule %添加表格底部粗线
    \end{tabular}
    \label{mtppo_para}
\end{table}

\subsection{Meta RL algorithms}
\subsubsection{MAML}
MAML is a model-agnostic algorithm for meta learning, it can be used for both supervised learning and reinforcement learning. In reinforcement learning, the goal of meta-learning is to allow the agent to quickly acquire policy for new tasks through only a small amount of experience samples in the testing phase. A task is an MDP, and any aspect of the MDP may change across tasks in the task distribution $p(\mathcal{T})$. At this time, the $f_\theta$ represents the agent's policy (a mapping from state $\mathbf{x}_t$ to action $\mathbf{a}_t$), and the loss function of each task $\mathcal{T}_i$ is: 
$$
\mathcal{L}_{\mathcal{T}_i}(f_\phi) = -\mathbb{E}_{\mathbf{x}_t, \mathbf{a}_t\sim f_\phi, p(\mathcal{T}_i})[\sum_{t=1}^H R_i(\mathbf{x}_t, \mathbf{a}_t)]
$$
where H is the horizon of MDP. In a $K$ shot reinforcement learning, $K$ rollouts $(\mathbf{x}_1, \mathbf{a}_1,...,\mathbf{x}_H)$ generated from $f_\theta$, task $\mathcal{T}_i$, and their corresponding rewards $R(\mathbf{x}_t, \mathbf{a}_t)$ will be used to adapt to the new task $\mathcal{T}_i$.

\subsubsection{ProMP}

ProMP (Proximal Meta-Policy search) proposes a novel meta-learning algorithm based on the MAML. It combines the PPO algorithm with the idea of MAML and improves the efficiency and stability of the meta-learning training process by controlling the statistical distance of both pre-adaptation and adapted policies. In general, ProMP optimizes
$$
\mathcal{L}_{\mathcal{T}}^{ProMP}(\theta)=\mathcal{L}_{\mathcal{T}}^{CLIP}(\theta') - \eta\mathcal{D_{KL}(\pi_{\theta_o}, \pi_{\theta})} \ s.t.  \ \theta'=\theta+\alpha\nabla_{\theta}\mathcal{L}_{\mathcal{T}}^{LR}(\theta), \ \mathcal{T} \sim p(\mathcal{T})
$$

where $\mathcal{L}_{\mathcal{T}}^{CLIP}(\theta')$ is the same as PPO which allows it to safely use a single trajectory for multiple gradient update steps, and $\mathcal{L}_{\mathcal{T}}^{LR}(\theta)$ results in the following objective:
$$
\mathcal{L}_{\mathcal{T}}^{LR}(\theta) = \mathbb{E}_{\mathbb{\tau} \sim P_\mathcal{T}(\mathbb{\tau}, \theta_o)}[\sum_{t=1}^{H-1}\frac{\pi_\theta(\mathbf{a}_t|\mathbf{s}_t)}{\pi_{\theta_o}(\mathbf{a}_t|\mathbf{s}_t)}A_{\pi_{\theta_o}}(\mathbf{a}_t, \mathbf{s}_t)]
$$

\begin{table}[htbp]
    \centering
    \caption{Hyperparameters of ProMP.}
    \begin{tabular}{c|c}
    \toprule  %添加表格头部粗线
    Hyperparameters & ML1, ML4, and ML20\\\hline
    Num mini-batches& 1 \\
    Inner loop opt-epochs & 1 \\
    Outer loop opt-epochs & 3\\
    Num episode-length& 8 \\\hline
    Hidden size & [2048, 1024, 512]\\
    Clip range & 0.2 \\
    Max grad norm & 1\\
    Outer loop learning rate & 3.e-4 \\
    Inner loop learning rate & 3.e-4 \\
    Discount ($\gamma$) & 0.9\\
    GAE lambda ($\lambda$) & 0.95\\
    Init noise std & 1\\\hline
    Desired kl & 0.016\\
    Ent-coef & 0\\
    \bottomrule %添加表格底部粗线
    \end{tabular}
    \label{promp_para}
\end{table}

\section{Performance discussion of PPO and SAC}\label{Appendix:C}

\begin{figure}[h]
 \centering  %居中
 \subfigure[SAC]{\includegraphics[width=0.45\hsize]{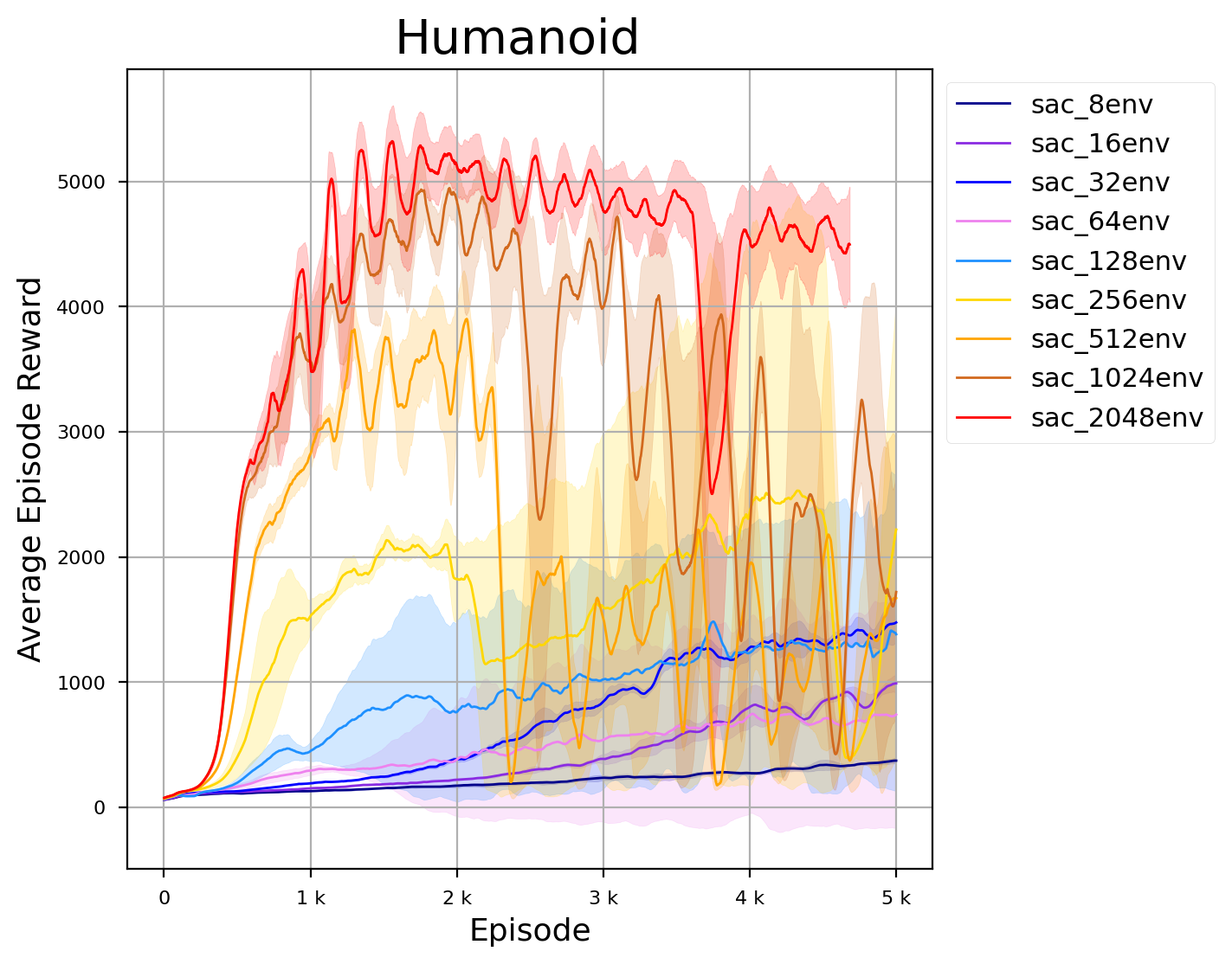}}
  \subfigure[PPO]{\includegraphics[width=0.45\hsize]{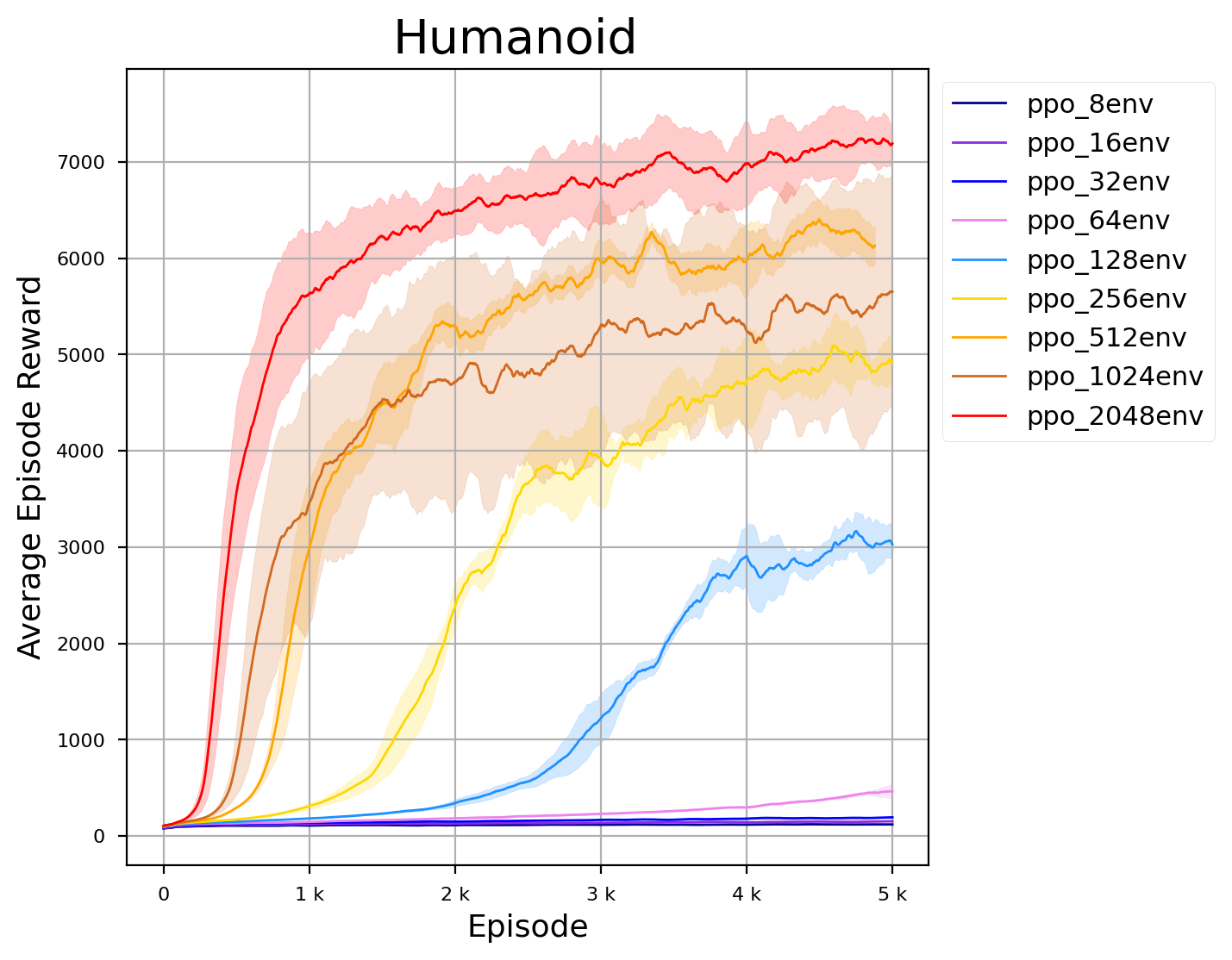}}
 \caption{Performance of SAC and PPO algorithms on humanoid with different numbers of environments}
 \label{Humanoid}
\end{figure}

In our RL/MARL experiments, we found that SAC does not work on almost all tasks, which is an anomalous phenomenon. Firstly, bimanual dexterous manipulation is a  challenging task, and previous studies have shown that simple model-free RL is basically unable to complete the task. So why do we get such good performance with PPO, and SAC almost all fail? We speculate that it is because the success of PPO relies on the huge improvement in sampling efficiency brought by 2048 parallel environments. Empirically, the gain of on-policy RL due to the improvement of sampling efficiency is larger than that of off-policy RL, so SAC can not be improved to the extent that it can complete the task of bimanual dexterous manipulation. In other words, it is normal that SAC can not complete our task, and PPO can complete it because of the high sampling efficiency brought by Isaac Gym. To verify our conjecture, we tested the SAC and PPO algorithm in different environments number (8, 16, 32, 64, 128, 256, 512, 1024, 2048) in the humanoid environment officially implemented by NVIDIA\,\cite{makoviychuk2021isaac}. The results are shown in the Figure.\ref{Humanoid}. It can be seen that the performance of the SAC algorithm is better than that of the PPO below 128 environments, indicating that the implementation of our SAC algorithm is good and meets our expectations. After more than 128 environments, the performance improvement of PPO by the increase of the number of environments is apparent, while the training of the SAC algorithm is unstable, and the performance is obviously inferior to the PPO. This proves our previous conjecture and explains why SAC performs so poorly on Bi-DexHands. In addition, because the action dimension of the Bi-DexHands has 50+ dimensions, the policy entropy method used by the SAC algorithm is easy causes instability during training. This instability appears to be exacerbated in the case of high sampling efficiency, and may also be a reason for the poor performance of SAC. In general, RL algorithms with high sampling efficiency will show some different characteristics. We also hope that Bi-DexHands can help researchers to study how to design RL algorithms with high sampling efficiency.

\begin{minipage}{\textwidth}
 \begin{minipage}[ht]{0.45\textwidth}
  \centering
     \makeatletter\def\@captype{table}\makeatother\caption{Hyperparameters of SAC.}
      \begin{tabular}{c|c} 
    \toprule  %添加表格头部粗线
    Hyperparameters & Humanoid\\\hline
    Num opt-epochs & 2 \\
    Num mini-batches& 1 \\
    Num episode-length & 32 \\\hline
    Hidden size & [1024, 1024, 1024]\\
    ReplayBuffer size& 40000 \\
    Learning rate & 3.e-4 \\
    Discount ($\gamma$)& 0.99 \\
    Polyak ($1-\tau$)& 0.995 \\
    Ent-coef & 0.2 \\
    Reward scale & 1 \\
    Max grad norm & 1 \\
    Batch size & 64 \\
    \bottomrule %添加表格底部粗线
    \end{tabular}
  \end{minipage}
      \label{sac_para_humanoid}
  \begin{minipage}[ht]{0.45\textwidth}
  \centering
        \makeatletter\def\@captype{table}\makeatother\caption{Hyperparameters of PPO.}
         \begin{tabular}{c|c}        
    \toprule  %添加表格头部粗线
    Hyperparameters & Humanoid \\\hline
    Num opt-epochs & 5 \\
    Num mini-batches & 4 \\
    Num episode-length & 32 \\\hline
    Hidden size & [1024, 1024, 1024]\\
    Clip range & 0.1\\
    Learning rate & 3.e-4 \\
    Discount ($\gamma$) & 0.99 \\
    GAE lambda ($\lambda$) & 0.95\\
    Init noise std & 1.0 \\
    Desired kl & 0.01 \\
    Max grad norm & 1 \\
    Ent-coef & 0\\
    \bottomrule %添加表格底部粗线
\end{tabular}
    \label{ppo_para_numanoid}
  \end{minipage}
\end{minipage}

\section{Details of multi-task/Meta RL training}\label{Appendix:D}

In order to better take advantage of Isaac Gym's large-scale parallel simulation, the design of our multi-task/Meta RL pipeline is different from all existing benchmarks. The largest difference is that we do not need to only sample part of all tasks for training, all tasks are trained at the same time. I will introduce our pipeline and the detail of the multi-task/Meta RL categories respectively below.

\subsection{High performance multi-task/meta RL pipeline using Isaac Gym}

Isaac Gym is a recent promising simulator for reinforcement learning. Different from previous simulators that can only use CPU to simulate, it can put all simulation calculations in GPU. Benefiting from the powerful parallel computing capability of GPU and avoiding switching data between CPU and GPU, Isaac Gym is able to create a large number of simulation environments in parallel without consuming many resources. This improvement in sampling efficiency is helpful for reinforcement learning, especially in on-policy RL and multi-task/meta RL. It also has a problem that Isaac Gym only allows one single environment instance to be created on a single GPU, so we can not create multiple gym-like environments at the same time as other simulators. So we designed a pipeline that runs through the entire training pipeline of one single environment instance, to make the multi-task/meta RL algorithm better leverage Isaac Gym's advantages. We directly load all tasks into an environment instance when initializing the environment, and use all tasks for data sampling and policy update at the same time, which is equivalent to that we sample all the environments every time in other simulators. In this way, each task can be trained synchronously, and the FPS is not significantly lower than one single task in parallel environments. To the best of our knowledge, our benchmark is the first to use Isaac Gym as a simulator for multi-task/meta RL. The sampling efficiency is greatly improved compared to previous simulators that rely on python parallel programs, which is helpful for multi-task/meta RL training. We hope that this will facilitate the research of multi-task/meta RL.

\subsection{Detail implementation of MT1, ML1, MT4, ML4, MT20, and ML20}

\begin{figure}[t]
 \centering  %居中
 \includegraphics[width=\hsize]{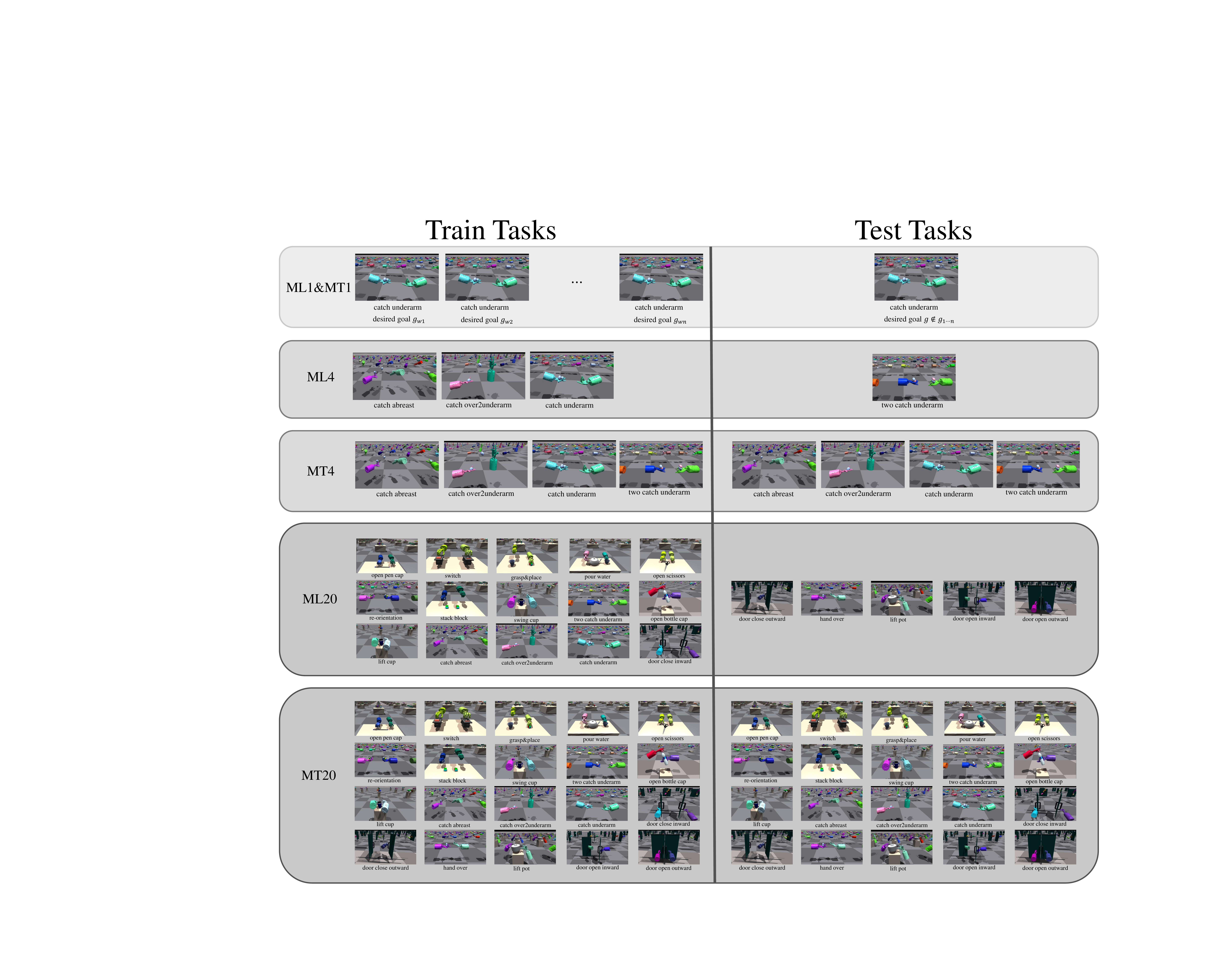} 
 \caption{Detail implementations of multi-task/meta settings.}
 \label{meta-world}
 \vspace{-0.2cm}
\end{figure}

Our multi-task/meta RL categories are formed by our carefully designed combinations of individual tasks detailed above. According to what we said above, the ML category is that all tasks are trained and tested at the same time. Therefore, MT1 and ML1, MT4 and ML4, MT20 and ML20 are all the same in terms of category settings. The difference is 1) ML categories only use a part of tasks as meta-train sets, and the other part is used for meta-test sets, while the MT categories are all trained together. 2) From the perspective of observation, multi-task adds a one-hot vector to represent task ID, while meta masks the observation related to the goal, which requires the Meta RL algorithm to learn by itself. Figure.\ref{fig2} visualizes the detailed design of our multi-task and meta categories. Let’s introduce their settings in detail separately:

\textbf{MT1\&ML1:} These two categories are only trained and tested in one type of task, only the pose of the goal is different between different tasks. We use Catch Underarm as the basic category, and translate the goal pose to the left, right, and back by 0.03cm, plus the goal of the original pose to form the task of MT1\&ML1. ML1 train on left, right translation, and in-position tasks, and have to quickly adapt to backward translation tasks.

\textbf{MT4\&ML4:} These two categories consist of 4 tasks, namely Catch Underarm, Hand Over, Catch Abreast, and Two Catch Underarm. The main reason for choosing these four tasks is that they are all object throwing and catching tasks, and the skills required are relatively similar, which is conducive to multi-task and Meta RL. It should be noted that to maintain the consistency of the environment, we no longer fix the base of the handover task. ML4 train on Catch Underarm, handover, catch abreast tasks and have to adapt to two Catch Underarm tasks.

\textbf{MT20\&ML20:} These two categories are composed of all of the 20 designed tasks. Due to the large span between different tasks, they are undoubtedly the most challenging tasks in Bi-DexHands. But it is also the most meaningful task because it covers the development of human dexterity and provides a good environment for us to master human-level dexterity. Note that there are some orders of magnitude differences in rewards between tasks. To make their rewards as close as possible, we scale the rewards in Grasp\&Place, Door Open Outward, Door Open Inward, Bottle Cap, Block Stack, Door Close Inward, Door close Outward, Lift Underarm, Re Orientation, Scissors, and Swing Cup tasks by 0.1 factor to ensure the order of magnitude consistency between their rewards. ML20 needs to adapt quickly in Door Close Outward, Hand Over, Lift Pot, Open Scissors, and Two Catch Underarm tasks. All the remaining environments are given for training.

\section{Visual observation about the Bi-DexHands}

Currently, the observations of Bi-DexHands are state-based. This is good for a beginning research, but not a realistic setting. In the real world, the agent always needs to estimate the states of other objects by visual observations, so using visual input RL is very important for sim2real transfer. Isaac Gym can use RGBD cameras to provide us with visual information, which can be directly used as image input or processed into a point cloud. We have tried point cloud RL in Bi-DexHands, but it still has some problems. Below I will detail why we are not using visual input in Bi-DexHands.

\begin{figure}[t]
 \centering  %居中
 \includegraphics[width=\hsize]{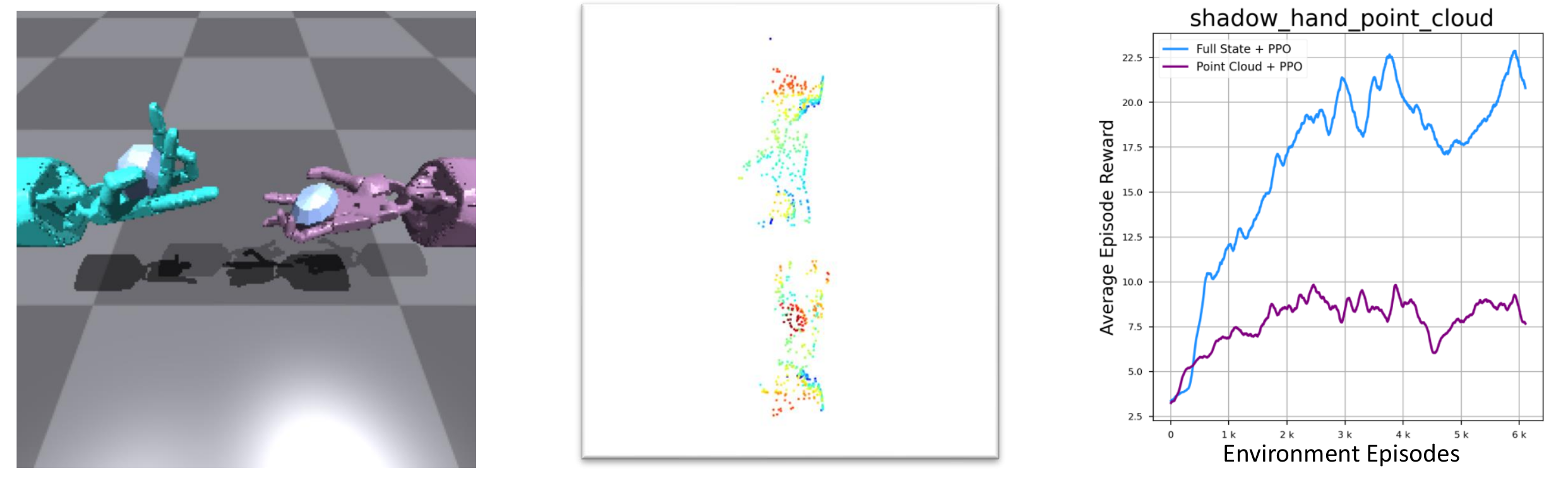} 
 \caption{Simple point cloud RL experiment. \textbf{Left image} is an RGB image taken with the RGBD camera that comes with isaacgym and is used to convert it into a point cloud. \textbf{Middle image} is a point cloud image converted from an RGB image. \textbf{Right image} is the experimental result of using PPO to run 6000 episodes in 256 parallel ShadowHandOver environments. The purple line is the result of RL training using point cloud input, and the blue line is the result of RL training using state input. Other parameters and settings are the same as baseline.}
 \label{point_cloud}
 \vspace{-0.2cm}
\end{figure}

The problem is that the parallelism of Isaac Gym's cameras is not very good. It can only obtain images one by one env serially, which will greatly slow down the running speed. At the same time, the training of the dexterous hand is very difficult and greatly depends on the high sampling efficiency. we do a simple experiment and result in Figure.\ref{point_cloud}. We replace the object information with point clouds in the case of a small number of environments, and use PointNet to extract point cloud features. It can be seen that under the same episode and same number of environments, the performance of point cloud input is not as good as full state input, but it can also achieve some performance. But also using an RTX 3090 GPU, the point cloud RL has only 200+ fps, and the full state can reach 30000+. In fact, we can only open up to 256 environments when using point clouds. This was a problem with Isaac Gym's poor parallel support for cameras, so we didn't use point clouds or other visual inputs as our baselines, but they certainly could.

\end{document}